\RequirePackage{rotating} 
\documentclass[nonacm,screen,acmsmall,authorversion]{acmart} %

\AtBeginDocument{%
  \providecommand\BibTeX{{%
    \normalfont B\kern-0.5em{\scshape i\kern-0.25em b}\kern-0.8em\TeX}}}

\usepackage{bbding}
\usepackage{rotating}
\usepackage[most]{tcolorbox}
\usepackage{caption}
\usepackage{amsmath}

\usepackage{newpxmath}
\usepackage{pifont}

\usepackage[shortlabels]{enumitem}
\usepackage{multirow}

\usepackage[para]{threeparttable}
\usepackage{tikz}
\usepackage{graphicx}
\usetikzlibrary{shapes, arrows.meta, positioning}
\usepackage{placeins}
\usetikzlibrary{arrows,shapes,positioning}
\usetikzlibrary{calc,decorations.markings}
\usepackage{color}

\usepackage{cleveref}

\newtheorem{definition}{\bf Definition} 
 
\newtheorem{example}{\bf Example}

\newcommand\RTI{\textsc{ti}$^\textsc{r}$}
\newcommand\RSA{\textsc{sa}$^\textsc{r}$}
\newcommand\REA{\textsc{ea}$^\textsc{r}$}
\newcommand\RCC{\textsc{cc}$^\textsc{r}$}
\newcommand\GTS{\textsc{ts}$^\textsc{g}$}
\newcommand\GFC{\textsc{fc}$^\textsc{g}$}
\newcommand\GRG{\textsc{rg}$^\textsc{g}$}
\newcommand\GCI{\textsc{ci}$^\textsc{g}$}

\newcommand\repoUrl{\url{https://github.com/SuDIS-ZJU/awesome-tabular-data-augmentation}}

\newcommand{\remarkbox}[1]{
\vspace{-\topsep}
\begin{tcolorbox}[colback=gray!10, colframe=gray!50, coltitle=black, width=\textwidth, boxrule=0.5pt, arc=1mm, boxsep=0mm, left=1.5mm, right=1.5mm, top=1.5mm, bottom=1.5mm, breakable]
\textbf{Remarks}.{#1}
\end{tcolorbox}
\vspace{-\topsep}
}

\begin{document}

\title{Tabular Data Augmentation for Machine Learning: Progress and Prospects of Embracing Generative AI}
\subtitle{An Extended Version}
\author{Lingxi Cui}
\email{cuilingxi.cs@zju.edu.cn}
\author{Huan Li}
\email{lihuan.cs@zju.edu.cn}
\author{Ke Chen}
\email{chenk@zju.edu.cn}
\author{Lidan Shou}
\email{should@zju.edu.cn}
\author{Gang Chen}
\email{cg@zju.edu.cn}
\affiliation{%
  \institution{The State Key Laboratory of Blockchain and Data Security, Zhejiang University}
  \city{Hangzhou}
  \country{China}
  \postcode{310027}
}

\begin{abstract}
Machine learning (ML) on tabular data is ubiquitous, yet obtaining abundant high-quality tabular data for model training remains a significant obstacle. 
Numerous works have focused on tabular data augmentation (TDA) to enhance the original table with additional data, thereby improving downstream ML tasks.
Recently, there has been a growing interest in leveraging the capabilities of generative AI for TDA. Therefore, we believe it is time to provide a comprehensive review of the progress and future prospects of TDA, with a particular emphasis on the trending generative AI.
Specifically, we present an architectural view of the TDA pipeline, comprising three main procedures: pre-augmentation, augmentation, and post-augmentation.
Pre-augmentation encompasses preparation tasks that facilitate subsequent TDA, including error handling, table annotation, table simplification, table representation, table indexing, table navigation, schema matching, and entity matching.
Augmentation systematically analyzes current TDA methods, categorized into retrieval-based methods, which retrieve external data, and generation-based methods, which generate synthetic data. We further subdivide these methods based on the granularity of the augmentation process at the row, column, cell, and table levels.
Post-augmentation focuses on the datasets, evaluation and optimization aspects of TDA.
We also summarize current trends and future directions for TDA, highlighting promising opportunities in the era of generative AI.
In addition, the accompanying papers and related resources are continuously updated and maintained in the GitHub repository at \repoUrl~to reflect ongoing advancements in the field.
\end{abstract}

\maketitle

\section{Introduction}
\label{section:intro}

Tabular data, such as relational tables, Web tables and CSV files, is among the most primitive and essential forms of data~\cite{DBLP:conf/iclr/BorisovSLPK23} in machine learning (ML), characterized by excellent structural properties, readability, and interpretability.
A testament to its significance, more than 65\% of datasets available on the Google Dataset Search platform are tabular files~\cite{DBLP:conf/semweb/BenjellounCN20}. This prevalence underscores its critical role across a myriad of fields, such as finance~\cite{rundo_machine_2019}, healthcare~\cite{hernandez_synthetic_2022}, education~\cite{luan_review_2021}.
The growing availability of repositories containing structured or semi-structured data offers new opportunities for tabular data research and applications built upon it, particularly in the fields of ML and artificial intelligence (AI).

\begin{figure}[!htbp]
  \centering
  \includegraphics[width=\linewidth]{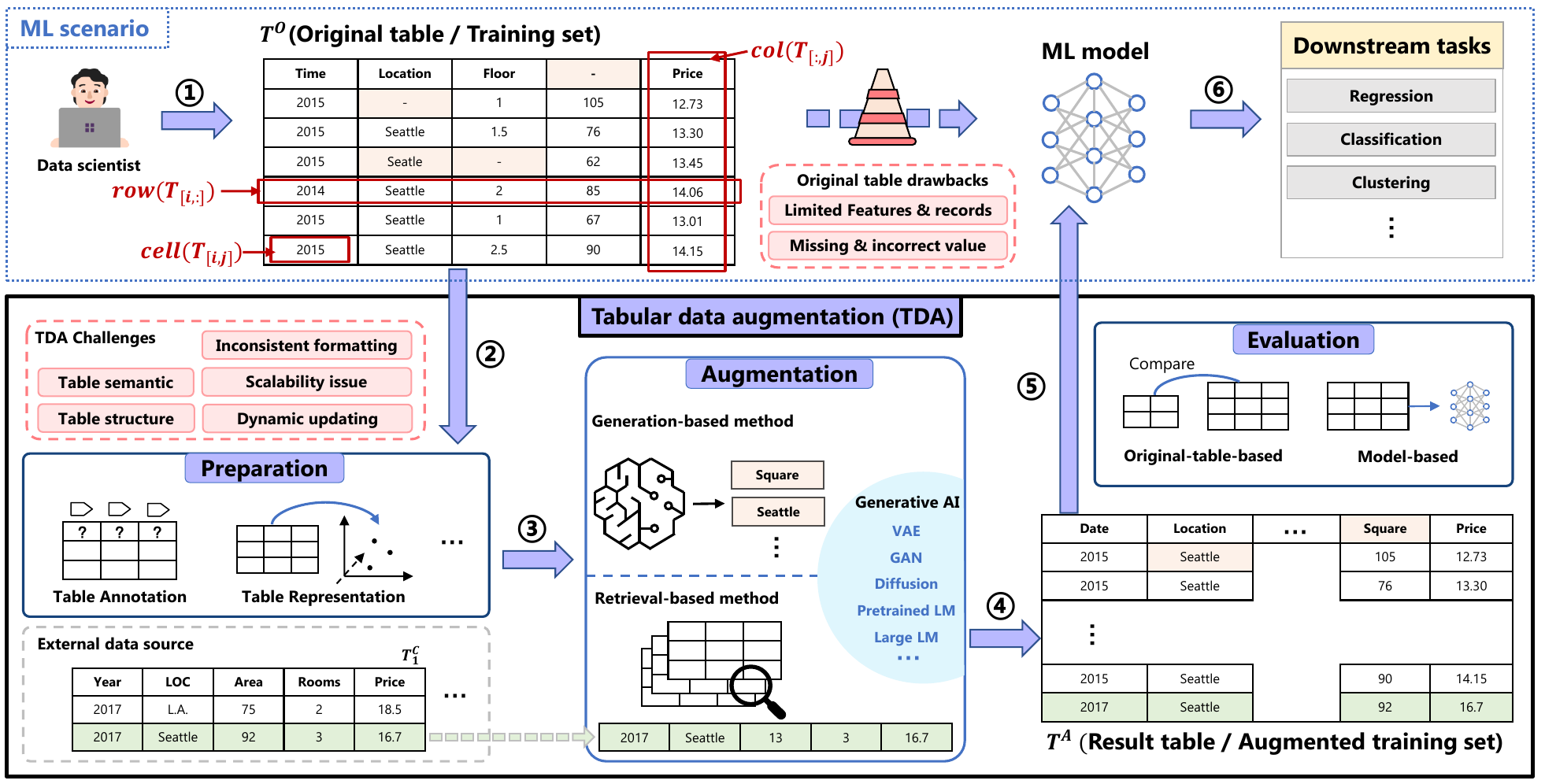}
  \caption{Example of TDA for ML:
  \ding{172} A data scientist aims to predict house prices based on factors like location and floor using an original training set ($T^O$) with limited features and records. The initial ML model yields sub-optimal results due to insufficient data and numerous missing or incorrect values. To improve performance, the scientist uses TDA to augment the original dataset with additional attributes (columns), records (rows), and corrected values (cells). \ding{173} Before augmentation, preparation steps, such as table annotation (e.g., recovering the missing column type of the 4th column in $T^O$), enhance the TDA process's effectiveness. \ding{174} Augmentation can be achieved through retrieval-based methods (e.g., integrating the 2nd row in $T^C_1$ from external data source) or generation-based methods that synthesize new data. \ding{175} The augmented table ($T^A$) combines the original and new data. \ding{176} After augmentation, evaluation steps evaluate the effectiveness of TDA process. \ding{177} Finally, the result-TDA table enable the scientist to train a more accurate price prediction model.
  }
  \label{f1_TDA_example}
  \Description{}
\end{figure}

However, acquiring substantial amounts of high-quality tabular data for ML model training remains a persistent challenge~\cite{chai_data_2022,liu_feature_2022}.
This is especially demanding because each individual table is modest in size and self-contained, making the overall data collection process resource-intensive and time-consuming.
According to the oft-cited~\cite{paton_dataset_2024} statistics, data scientists spend over 80\% of their time on ML data preparation tasks, including data discovery and augmentation.
The complexity and uneven quality of massive tabular datasets from various domains further complicate the acquisition of high-quality tabular data~\cite{castro_fernandez_aurum_2018,fan_table_2023}. 
Furthermore, in the era of large language models (LLMs), tabular data is one of the preferred data formats that LLMs consume, and existing high-quality tabular datasets may soon be exhausted~\cite{zhou_survey_2024}. 
Additionally, in the industrial sector where tabular data is most commonly used, the availability of data is often limited due to privacy concerns~\cite{li_chatdoctor_nodate}.
All of these factors have led to significant efforts being devoted to developing techniques that support \textbf{tabular data augmentation (TDA)}.
Through our extensive investigation, we have collected a total of 70 highly relevant studies from 2010 to 2024.
In this study, we define TDA as the process of augmenting the original dataset (table) to enhance the performance of downstream ML models.
An illustration of TDA and its role in ML scenarios is provided in Fig.~\ref{f1_TDA_example}. 
TDA techniques aim to enrich the ML tasks by incorporating additional data, either from external data sources\footnote{In this study, we also refer to external data sources as \emph{table pools} as they can contain various table sources, including databases, Web tables, and so on.} or synthesized by generative methods\footnote{In addition to the recently much-discussed language models~\cite{tran_differentially_2024}, generative methods also include statistical approaches such as MICE~\cite{buuren_mice_2011}, deep generative models like diffusion models~\cite{liu_controllable_2024}, and so on.}.

On the business side, the market for tabular data persists in its growth. For example, by year 2022, the US open data market (\texttt{data.gov}) has amassed a total of 335,000 datasets, contributing to a staggering \$3 trillion to the US economy~\cite{fan_table_2023}.
Moreover, the Augmented Analytics Market is expected to attain \$35.6 billion expanding at a 22.70\% CAGR (Compound Annual Growth Rate) report by Market Research Future\footnote{\url{https://www.marketresearchfuture.com/reports/augmented-analytics-market-7464}.}.
Overall, TDA has garnered extensive attention from the research community and generated significant demand from the business sector.

However, the TDA task can be particularly challenging due to the unique characteristics of tabular data and the potential scale of table pools. 
Unlike homogeneous data such as images or text, tabular data is heterogeneous, typically containing both dense numerical and sparse categorical attributes~\cite{borisov_deep_2022}. 
Additionally, it has complicated structure, such as row and column permutation invariance and hierarchical organization, where cells belong to rows and rows belong to tables. 
Moreover, many TDA tasks involve large-scale table pools, sometimes encompassing millions of tables.
These tables often have inconsistent attribute naming and value formatting, and table pools  themselves are dynamic, changing over time.
Various TDA approaches have been proposed to address these challenges in diverse ways.
As mentioned and illustrated in Fig.~\ref{f1_TDA_example}, these methods can be broadly categorized into retrieval-based approaches, which involve retrieving data from table pools, and generation-based approaches, which involve generating new content.
With the rise of generative AI, many recent TDA works including both retrieval- and generation-based approaches (16 from 2023 to 2024 in our review) have embraced these models from different perspectives, and this trend continues to grow.
\textbf{Generative AI}, designed for generating new content based on patterns learned from training data, broadly includes various models such as pre-trained language models (e.g., BERT~\cite{fan_semantics-aware_2023} and T5~\cite{fang_large_2024}), large language models (e.g., ChatGPT~\cite{li_table-gpt_2023} and LLaMA~\cite{tang_struc-bench_nodate}), variational autoencoders~\cite{nazabal_handling_2020}, generative adversarial networks~\cite{engelmann_conditional_2021}, and diffusion models~\cite{liu_controllable_2024}.
Such models have swept across various fields from computer vision (CV)~\cite{li_data_2021,zhou_survey_2024,wang_comprehensive_2024} to natural language processing (NLP)~\cite{DBLP:conf/acl/FengGWCVMH21,DBLP:journals/tacl/ChenTRBY23,li_data_2021,zhou_survey_2024,wang_comprehensive_2024}, and are now beginning to permeate the field of tabular data analysis.
In this context, we observe that there has yet to be a systematic review, synthesis, and categorization of existing TDA methods, let alone a discussion on the integration of current trending generative AI methods. Therefore, we believe it is both timely and essential to conduct such a survey to help readers and practitioners grasp the advancements in this critical field and pinpoint significant research opportunities in the era of generative AI.

\begin{table}[!htbp]
\caption{Overview of prior related surveys. Some related concepts to TDA: \textbf{Data Integration} is the task of combining information from different relational data sources without an original table, while TDA aims to augment the original table to boost downstream ML tasks; \textbf{Table Discovery} identifies datasets from external data sources that may contain useful information, often serving as an intermediate step in TDA; \textbf{Table Representation} aims to transform tabular data into meaningful vector representation for further processing, which can be considered as one of preparation tasks for TDA.}
\label{T1_related_survey}
\resizebox{\linewidth}{!}{
\begin{threeparttable}
\begin{tabular}{llll}
\toprule 
Reference                     & Field                                & Data Type                                & Task                                              \\ \midrule

Zhang and Balog~\cite{zhang_web_2020}         & Data Management                     & Web Tables                               & Web Table Extraction, Retrieval, and Augmentation \\

Li et al.~\cite{li_data_2021}           & NLP, CV                              & Relational Tables, Text and Images                            & Data Augmentation, Data Preparation and Integration                                \\
Zhou et al.~\cite{zhou_survey_2024}       & NLP, CV, and Multimedia              & Text, Image, Audio Signal              & Data Augmentation                                 \\
Wang et al.~\cite{wang_comprehensive_2024}                       & NLP, CV, and Multimedia              & Image, Text, Graph, Table, and Time-series & Data Augmentation                                 \\
Fonseca and Bacao~\cite{fonseca_tabular_2023}   & Data Management                     & Relational and Web Tables~$^\clubsuit$           & Tabular Data Generation                               \\
Hulsebos et al.~\cite{hulsebos_models_2023}   & Data Management                     & Relational and Web Tables           & Table Representation                               \\
Chapman et al.~\cite{chapman_dataset_2020}   & Data Management                     & Relational and Web Tables           & Table Discovery                                 \\
Fan et al.~\cite{fan_table_2023}         & Data Management                     & Relational and Web Tables           & Table Discovery                                 \\ 
Paton et al.~\cite{paton_dataset_2024}     & Data Management                     & Relational and Web Tables           & Table Discovery and Exploration                 \\
\midrule

\textbf{Ours}                 & Data Management                     & Relational and Web Tables  & Data Augmentation                       \\ 

\bottomrule                 
\end{tabular}
\begin{tablenotes}
    \item[$\clubsuit$] To the best of our knowledge, the TDA studies we primarily review focus on relational tables and Web tables, whereas other data formats, such as CSV and JSON files, can be handled with minimal conversion.
\end{tablenotes}
\end{threeparttable}
}
\end{table}

Previous efforts, as summarized in Table~\ref{T1_related_survey}, have focused on other research fields or only touched on aspects of TDA. 
For example, Zhang et al.~\cite{zhang_web_2020} mention tabular data augmentation only briefly in their research on Web tables.
Another tutorial~\cite{li_data_2021} focuses on data augmentation for data preparation and data integration, but does not specifically discuss its support for ML training; moreover, this tutorial does not place a particular emphasis on tabular data.
A more recent survey~\cite{zhou_survey_2024} on data augmentation in the era of large models mainly focuses on NLP and CV tasks, rather than tabular data.
Similarly, Wang et al.~\cite{wang_comprehensive_2024} cover data augmentation techniques across various data modalities (i.e., image, text, graph, table, and time-series), but TDA is only a small part of their broader scope. 
Their modality-independent taxonomy may not be the most suitable for an in-depth exploration of TDA specifically.
For example, one of the key methods covered in our survey, retrieval-based TDA, is not addressed in their work.
Other literature reviews on tabular data focus on table discovery~\cite{chapman_dataset_2020,fan_table_2023,paton_dataset_2024}, table representation~\cite{hulsebos_models_2023} or tabular data generation~\cite{fonseca_tabular_2023}, rather than augmentation.

Our survey stands out from previous reviews and tutorials by providing a comprehensive examination of TDA methods tailored for ML scenarios, with a special emphasis on the recent advancements in incorporating generative AI techniques.
We have meticulously selected 70 significant works from the fields of data management and artificial intelligence, dating back to 2010, to offer a diverse range of perspectives. Based on these works, we thoroughly investigate TDA-related research.
Specifically, we introduce taxonomies that categorize these works from both task and table content granularity perspectives. This allows us to clearly compare retrieval-based and generation-based methods, highlighting their respective strengths and weaknesses. Furthermore, we present a complete TDA pipeline that covers the entire process from preparation to augmentation to evaluation. Finally, we summarize future trends and highlight new opportunities in the TDA field, particularly in the context of generative AI.

The survey is structured as follows. Section~\ref{section:preliminaries} provides preliminaries on TDA for ML and outlines an overall pipeline to characterize and classify approaches.
Section~\ref{section:preaug} offers a comprehensive overview of eight classes of pre-augmentation techniques. 
Section~\ref{section:aug} delves into the details of specific TDA techniques at different levels, categorized into retrieval-based and generation-based methods.
Section~\ref{section:postaug} explores post-augmentation techniques mainly for evaluation and optimization after TDA.
Section~\ref{section:future} discusses challenges and future directions, with an emphasize on the trending technologies like generative AI. 
Section~\ref{section:conclusion} concludes the survey.

\section{Preliminaries}
\label{section:preliminaries}

In this section, we will start by introducing the notation related to TDA and outlining the level-based taxonomy that defines the various levels of TDA methods (i.e., row, column, cell, and table) in Section~\ref{subsection:preliminaries-notation}. 
Subsequently, we will present the TDA pipeline and offer a taxonomy of methods from a task-oriented perspective in Section~\ref{subsection:preliminaries-pipeline}. 
In the following sections, \emph{tasks} will serve as the primary basis for categorization, with \emph{levels} providing a more granular categorization criterion.

\subsection{Notation in TDA and Level-based Taxonomy}
\label{subsection:preliminaries-notation}

First, we provide a formalization of \emph{tables}, a prevalent data structure essential for the organization and presentation of data as follows.

\begin{definition}[Table]
A \textbf{table} $T$ is an arrangement that organizes data into rows and columns, forming a grid of cells for systematic information representation. Each \textbf{cell}, denoted as $T[i,j]$, is at the intersection of row $i$ and column $j$, serving as the basic unit for data storage. 
The \textbf{rows} ($T[i, :]$) run horizontally and group data entries, while \textbf{columns} ($T[:, j]$) extend vertically, with each focusing on a specific data attribute.
Additionally, \textbf{metadata}, such as table captions, provides contextual textual information around the table.
\end{definition}

An intuitive example of a table and its primary components --- columns, rows, and cells --- is depicted in Fig.~\ref{f1_TDA_example}.
Given a table $T$, we use $T.\mathcal{R}$ and $T.\mathcal{A}$ to denote the set of its rows and columns (attributes), respectively. 
Notably, our analysis is restricted to tables that solely manage numerical and textual data structured in rows and columns.
This explicitly excludes tables that incorporate nested tables, lists, forms, images, or any other non-textual and non-numerical values within their cells.
We now provide the formal definition of tabular data augmentation as follows.

\begin{definition}[Tabular Data Augmentation, TDA]
Given an original table $T^O$ and a specific ML model $f(\Theta)$ parameterized by $\Theta$, the task of \emph{Tabular Data Augmentation} aims to expand $T^O$ into an augmented table $T^A$ that includes additional data values in its rows and/or columns.
The goal is for the ML model $f(\Theta)$, trained with $T^A$, to achieve superior model performance compared to the version trained with $T^O$.
Formally,
\begin{equation}
    \begin{aligned}
    & T^A \gets \operatorname{TDA}(T^O, \texttt{level}, [\mathbb{T}], [\mathbb{G}]), \\
    & \text{s.t.} \,\,\,\, \mathbb{E}(f(\Theta^A)) < \mathbb{E}(f(\Theta^O)),
    \end{aligned}
\end{equation}
where $\texttt{level} = \{  \texttt{row}, \texttt{column}, \texttt{cell}, \texttt{table}\}$ refers to the granularity at which the TDA operates;
$\mathbb{T}$, an optional input for retrieval-based TDA,
represents the pool of tables (simply called a \emph{table pool}) for information enrichment use\footnote{The pool of tables, $\mathbb{T}$, is often necessary for retrieval-based TDA methods. However, for generation-based TDA methods, the original table as input is usually sufficient. This is because generative models have typically been pre-trained on large amounts of external data, allowing them to retain and leverage a wealth of background information.};
$\mathbb{G}$, an optional input for generation-based TDA, implies the use of a specific generative method.
and $\mathbb{E}(f(\Theta^A))$ and $\mathbb{E}(f(\Theta^O))$ refer to the empirical errors of the ML models trained on the datasets $T^A$ and $T^O$, respectively\footnote{Notably, the model training process can involve advanced feature engineering techniques such as coreset~\cite{wang_coresets_2022} and feature selection~\cite{chepurko_arda_2020} to refine the training data from the input table. Here, we assume that these techniques will be identically applied despite the difference in the input tables.}. 
\end{definition}

\begin{figure}[!htbp]
  \centering
  \includegraphics[width=\linewidth]{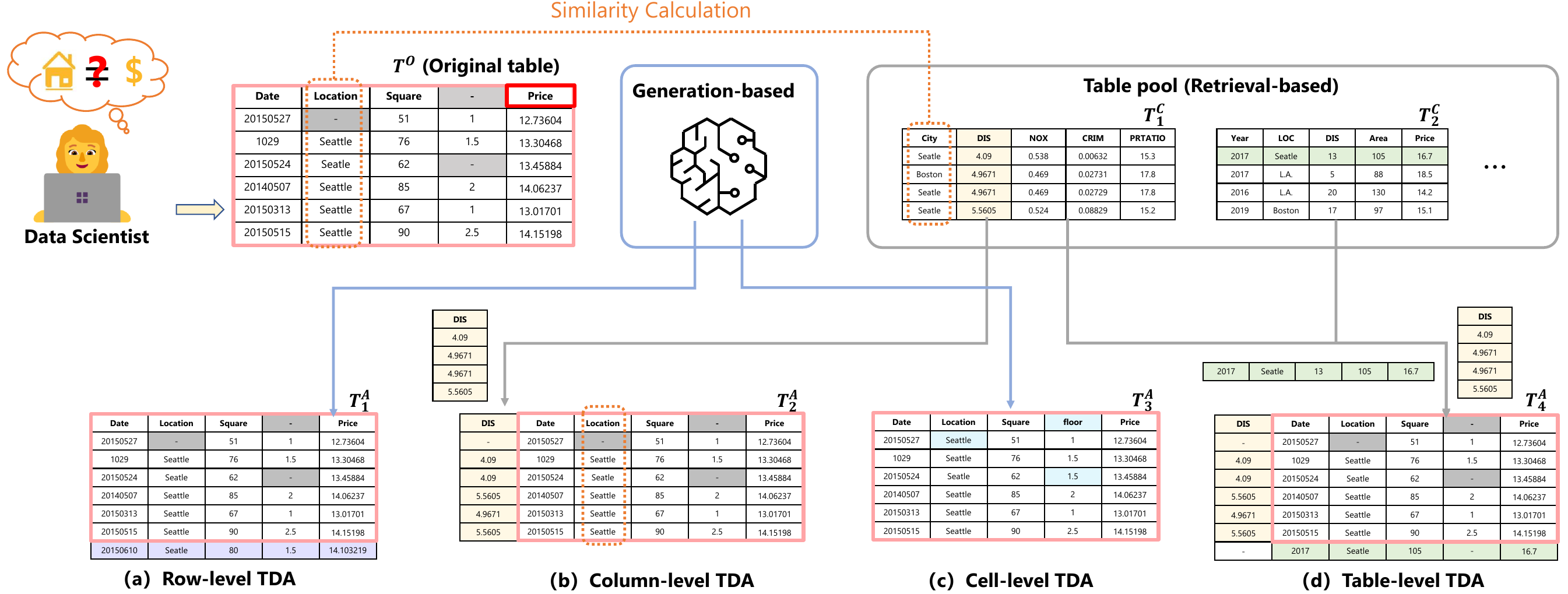}
  \caption{Instantiation of TDA tasks at different levels: \textbf{(a) row-level TDA}, which adds new rows to the original table, and \textbf{(b) column-level TDA}, which introduces new columns, \textbf{(c) cell-level TDA} addresses missing values, while \textbf{(d) table-level TDA} enhances the table by adding both rows and columns. 
  From another viewpoint, \textbf{retrieval-based TDA methods} (grey arrows) augment the original table with data sourced from the table pool, while \textbf{generation-based TDA methods} (blue arrows) generate new data directly based on the original table. Both retrieval- and generation-based TDA methods can be implemented at various levels, including row, column, cell, and table. The illustration shows only \emph{some} of these possibilities for clarity. We use the housing dataset~\cite{DVN/DHZ9VY_2018} for the toy example.}
  \label{f2_TDA_levels}
  \Description{}
\end{figure}

\begin{example}
    \label{TDA_level_ex2}
    Fig.~\ref{f2_TDA_levels} depicts a typical TDA scenario:
    A data scientist aims to build a house price prediction model, but the available training data (the original table $T^O$ containing locations and prices) is limited in features and records, and contains numerous missing or incorrect metadata and cell values (shown in gray). 
    A model trained on this low-quality data is likely to produce subpar results. 
    To address this issue, the data scientist needs to augment the original training set with more comprehensive data. This can be done a) by retrieving additional data from
    table pools for retrieval-based TDA, or b) by generating new data using existing generative methods for generation-based TDA. 
    The augmented data can include additional attributes, records, and/or cell values, reflected as enriched features and samples in the training set.
    The primary goal of this TDA process is to improve the overall quality and performance of the downstream house price prediction model.
\end{example}

Referring to Fig.~\ref{f2_TDA_levels}, these TDA sub-tasks are instantiated based on the \texttt{level} that the TDA procedure is acting on. We formally define these tasks one by one as follows.

\begin{definition}[Row-level TDA]
One significant challenge in ML is the scarcity and often uneven distribution of samples, such as in long-tail data.
To address this, \emph{row-level TDA} involves adding additional rows to a given table $T^O$.
This procedure aims to obtain more samples for training, potentially increasing the variety of sample categories and altering the sample distribution to some extent.
For row-level TDA, only additional rows are considered, and the relationship between $T^A$ and $T^O$ satisfies:
$$
(T^O.\mathcal{A} = T^A.\mathcal{A}) \wedge (T^O.\mathcal{R} \subset T^A.\mathcal{R}).
$$
\end{definition}

\begin{example}
    \label{TDA_level_ex3}
    Continuing from the scenario in Example~\ref{TDA_level_ex2}, the data scientist aims to expand the size and diversity of samples in the training set. 
    Fig.~\ref{f2_TDA_levels} (a) demonstrates a generation-based TDA that only takes the original table $T^O$ as input without external data. Generative methods typically learn the structure and the pattern of the original table $T^O$ and then generate new synthetic record (shown in purple), resulting in the augmented table $T^A_1$.
    These missing values can be addressed during post-processing steps, such as filtering and imputation.
\end{example}

\begin{definition}[Column-level TDA]
Adequate features are crucial for training high-performing ML models, but good features alone are not always enough. Therefore, \emph{column-level TDA} involves extending the original table with additional columns, enriching the feature set.
Retrieval-based TDA at the column level often involves joining related tables. This procedure may necessitate feature engineering to remove less informative records and features, which can result in the augmented table having fewer rows than the original ($|T^O.\mathcal{R}| > |T^A.\mathcal{R}|$).
Generation-based column-level TDA, on the other hand, retains the number of rows ($|T^O.\mathcal{R}| = |T^A.\mathcal{R}|$).
All in all, it satisfies:
$$
(T^O.\mathcal{A} \subset T^A.\mathcal{A}) \wedge (|T^O.\mathcal{R}| \geq |T^A.\mathcal{R}|). 
$$
\end{definition}

\begin{example}
    \label{TDA_level_ex4}
    Continuing from Example~\ref{TDA_level_ex2}, the data scientist decides to expand the number of features in $T^O$. Still, we take the retrieval-based TDA as an example, as shown in Fig.~\ref{f2_TDA_levels} (b). 
    This procedure involves computing column similarity. For example, if the column ``Location'' (the 2nd column in $T^O$) and the column ``City'' (the 1st column in table $T^C_1$) have largely similar values, $T^O$ can be augmented with the column ``DIS (distance)'' (the 2nd column from $T^C_1$), resulting in the augmented table $T^A_2$. 
\end{example}

\begin{definition}[Cell-level TDA]
Empty table cells are common, and generating data representations with $\texttt{null}$ values can lead to suboptimal results in downstream tasks.
\emph{Cell-level TDA} involves filling these empty table cells to improve the quality of training data for ML tasks. 
For cell-level TDA, the relationship between $T^A$ and $T^O$ satisfies:
$$
(T^O.\mathcal{A} = T^A.\mathcal{A}) \wedge (|T^O.\mathcal{R}| = |T^A.\mathcal{R}|) \wedge (\forall i,j: T^A[i,j] \neq \texttt{null}).
$$
\end{definition}

\begin{example}
    \label{TDA_level_ex5}
    Continuing from Example~\ref{TDA_level_ex2}, the data scientist decides to fill the empty cells using a generation-based TDA, as shown in Fig.~\ref{f2_TDA_levels} (c). 
    Generative methods leverage the context (e.g., statistical distribution) from the original table $T^O$ to fill in missing metadata and data values (shown in blue), resulting in the augmented table $T^A_3$.
\end{example}

\begin{definition}[Table-level TDA]
\emph{Table-level TDA} involves enriching the original table with both additional rows and columns. This aims to acquire more features and samples for ML purposes. For table-level TDA, the relationship between $T^A$ and $T^O$ satisfies:
$$
(T^O.\mathcal{A} \subset T^A.\mathcal{A}) \wedge (|T^O.\mathcal{R}| < |T^A.\mathcal{R}|).
$$
\end{definition}

\begin{example}
    Continuing from Example~\ref{TDA_level_ex2}, the data scientist now seeks a more balanced approach that can augment the table along both the row and column dimensions. 
    Using a retrieval-based TDA, as shown in Fig.~\ref{f2_TDA_levels} (d), the approach first retrieves the top-related tables from the table pool (e.g., $T^C_1$ and $T^C_2$). These tables are then integrated with $T^O$, resulting in the augmented table $T^A_4$, which includes additional rows (e.g., the 1st row from $T^C_2$) and columns (e.g., the 2nd column from $T^C_1$).
\end{example}

The aforementioned TDA methods of four different levels are summarized in Table~\ref{preliminaries_tda_tasks}. 
While the approaches differ, they all aim to enhance the quality of the original training dataset, thereby improving the performance of the resulting trained model.

\begin{table}[!htbp]
\caption{The level-based taxonomy for TDA methods.}
\label{preliminaries_tda_tasks}
\footnotesize
    \begin{tabular}{lp{60mm}p{48mm}}
    \toprule
    \texttt{level}  & \textbf{Description}                                     & \textbf{Relationship between $T^A$ and $T^O$}                                                                                     \\ \midrule
    \texttt{row}    & enrich $T^O$ with additional rows           & $(T^O.\mathcal{A} = T^A.\mathcal{A}) \wedge (T^O.\mathcal{R} \subset T^A.\mathcal{R})$                                               \\
    \texttt{column} & enrich $T^O$ with additional columns        & $(T^O.\mathcal{A} \subset T^A.\mathcal{A}) \wedge (|T^O.\mathcal{R}| \geq |T^A.\mathcal{R}|)$                                        \\
    \texttt{cell}   & fill in the missing values within the cells of $T^O$ & $(T^O.\mathcal{A} = T^A.\mathcal{A}) \wedge (|T^O.\mathcal{R}| = |T^A.\mathcal{R}|) \wedge (\forall i,j: T^A[i,j] \neq \texttt{null})$ \\
    \texttt{table}  & enrich $T^O$ with both additional rows and columns  & $(T^O.\mathcal{A} \subset T^A.\mathcal{A}) \wedge (|T^O.\mathcal{R}| < |T^A.\mathcal{R}|)$                                           \\ \bottomrule
    \end{tabular}
\end{table}

\subsection{Pipeline of TDA and Task-based Taxonomy}
\label{subsection:preliminaries-pipeline}

\begin{figure}[!htbp]
  \centering
  \includegraphics[width=\linewidth]{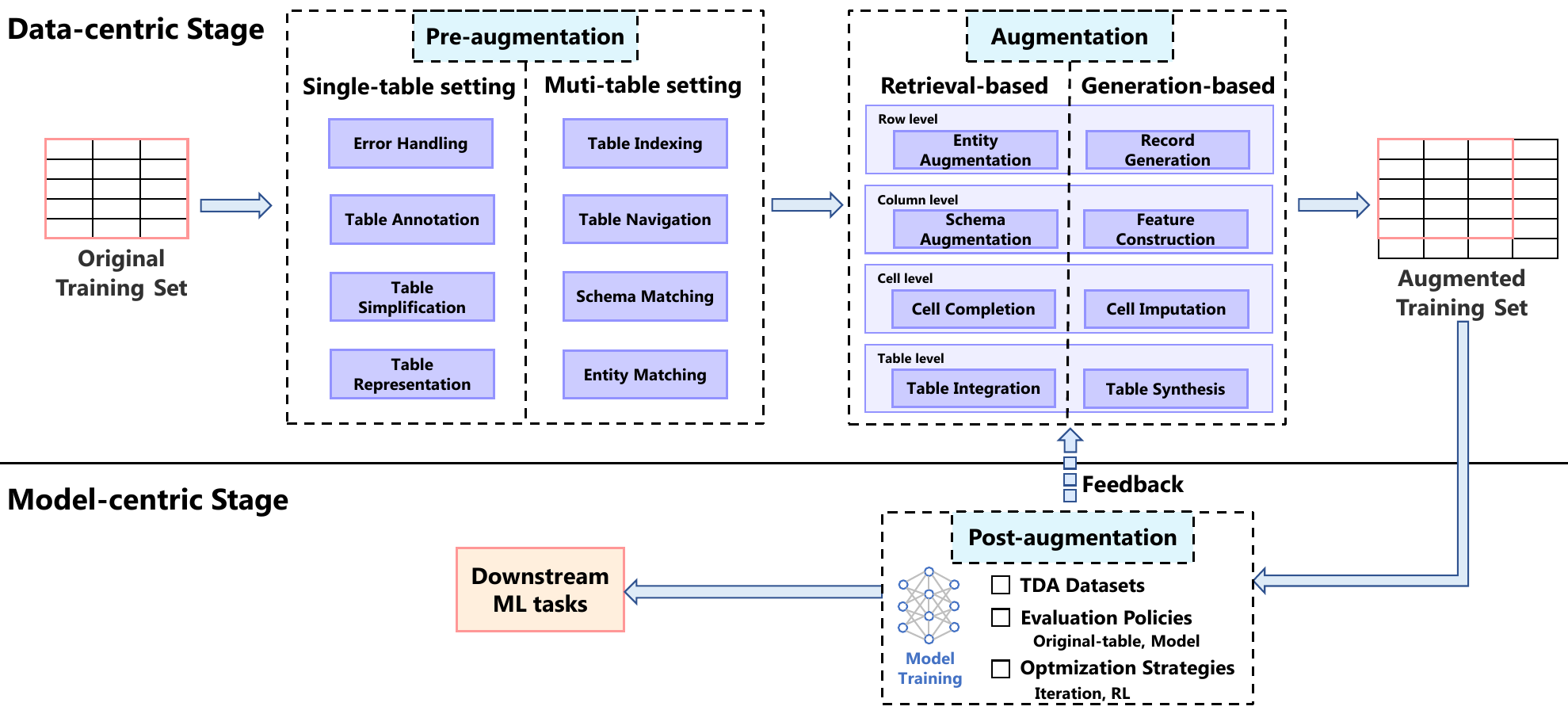}
  \caption{The overview of TDA pipeline and the task-based taxonomy for TDA approaches. The input and output of the TDA pipeline are the original table $T^O$ and the augmented table $T^A$, respectively. The TDA pipeline comprises three main procedures: \textbf{pre-augmentation}, \textbf{augmentation}, and \textbf{post-augmentation}.}
  \label{f3_TDA_pipeline}
  \Description{}
\end{figure}

In this section, we provide an overview of the key topics covered in the survey, structured around the TDA pipeline from a task-oriented perspective, as shown in Fig.~\ref{f3_TDA_pipeline}.
We categorize tasks by levels for finer classification.
The pipeline highlights critical stages and procedures from the original training dataset $T^O$ to the augmented training dataset $T^A$. We first overview the entire TDA pipeline, followed by a brief introduction to each pivotal procedure within it.

The TDA pipeline has two main stages:
\begin{itemize}[leftmargin=*]
\item \textbf{Data-centric stage} includes pre-augmentation and augmentation procedures to transform the original data into augmented data. Pre-augmentation involves preparation to enhance augmentation, while augmentation details existing TDA methods, divided into retrieval-based and generation-based methods.
\item \textbf{Model-centric stage} focuses on post-augmentation, primarily involving ML model training and evaluation to assess and optimize the augmented dataset. If the augmented dataset is not satisfactory, it is necessary to return to the data-centric stage for further augmentation.
\end{itemize}

\paragraph{\bf (1) Pre-Augmentation}

In the TDA pipeline, pre-augmentation encompasses preparation tasks to facilitate effective augmentation. For tabular data, issues like missing or incorrect cell values and unreliable metadata are common due to inappropriate data sharing and incompatible naming conventions~\cite{fan_table_2023}.
For the table pool, with tables reaching millions or more, data preparation before TDA is crucial. 
Pre-augmentation aims to a) improve the quality of both the original table and the tables from the pool (in the case of retrieval-based methods) and b) better organize the table pool for improved acceleration and scalability. 
This can involve a range of tasks, applicable to both single-table and multi-table settings, as listed in Table~\ref{preliminaries-pipeline-preaug}.
The table's right part presents their support for various different TDA tasks. The following observations can be drawn:
\begin{enumerate}[leftmargin=*]
\item The tasks in single-table setting are applicable to all target TDA tasks, as all target TDA tasks require preprocessing of the original table. 
\item The multi-table setting tasks are specifically for handling tables from the pool and may not be suitable for generation-based TDA tasks. 
\item Entity matching, a pre-augmentation task that focuses on the relationships between rows, may not be much beneficial for the schema augmentation (\RSA{}), a TDA task at the column level. 
\end{enumerate}
Section~\ref{section:preaug} will provide a detailed introduction to relevant techniques for pre-augmentation tasks.

\begin{table}[!htbp]
\caption{Overview of the pre-augmentation tasks and their target TDA tasks (see Table~\ref{preliminaries-pipeline-aug} for the definitions): \REA{} (Entity Augmentation), \RSA{} (Schema Augmentation), \RCC{} (Cell Completion), \RTI{} (Table Integration), \GRG{} (Record Generation), \GFC{} (Feature Construction), \GCI{} (Cell Imputation), and \GTS{} (Table Synthesis).}%
\label{preliminaries-pipeline-preaug}
\centering
\resizebox{\linewidth}{!}{
\begin{tabular}{ccl|cccccccc}
\toprule
\multirow{2}{*}{Setting}      & \multirow{2}{*}{Task Name} & \multirow{2}{*}{Description}                        & \multicolumn{4}{c}{Retrieval-based TDA}           & \multicolumn{4}{c}{Generation-based TDA}          \\ \cmidrule(l){4-11} 
                              &                            &                                                     &\REA{}    & \RSA{}       & \RCC{}       & \RTI{}       & \GRG{}       & \GFC{}       & \GCI{}       & \GTS{}       \\ \midrule
\multirow{4}{*}{\begin{tabular}[c]{@{}c@{}}Single-\\ table\end{tabular}} & {Error Handling}            & process dirty data in tables                        & \Checkmark & \Checkmark & \Checkmark & \Checkmark & \Checkmark & \Checkmark & \Checkmark & \Checkmark \\
                              & {Table Annotation}           & infer table metadata information                    & \Checkmark & \Checkmark & \Checkmark & \Checkmark & \Checkmark & \Checkmark & \Checkmark & \Checkmark \\
                              & {Table Simplification}              & streamline a table down to its essential elements                   & \Checkmark & \Checkmark & \Checkmark & \Checkmark & \Checkmark & \Checkmark & \Checkmark & \Checkmark \\
                              & {Table Representation}       & encode table elements to a latent vector space      & \Checkmark & \Checkmark & \Checkmark & \Checkmark & \Checkmark & \Checkmark & \Checkmark & \Checkmark \\ \midrule
\multirow{4}{*}{\begin{tabular}[c]{@{}c@{}}Multi-\\ table\end{tabular}}  & {Table Indexing}             & assign a unique identifier to table elements        & \Checkmark & \Checkmark & \Checkmark & \Checkmark &            &            &            &            \\
                              & {Table Navigation}           & organize the table pool by connecting similar tables     & \Checkmark & \Checkmark & \Checkmark & \Checkmark &            &            &            &            \\
                              & {Schema Matching}            & find matching pairs of columns in different tables  & \Checkmark & \Checkmark & \Checkmark & \Checkmark &            &            &            &            \\
                              & {Entity Matching}            & find matching pairs of entities in different tables & \Checkmark &            & \Checkmark & \Checkmark &            &            &            &            \\ \bottomrule
\end{tabular}
}
\end{table}

\paragraph{\bf (2) Augmentation}

Augmentation is the core procedure of the TDA pipeline,
enhancing the original table with more data to improve downstream ML tasks.
The techniques for TDA can be broadly divided into retrieval-based and generation-based methods. 
Retrieval-based methods are considered a data-driven TDA task based on the original table $T^O$ (called \emph{query table}), with table pools as additional input. 
The key to this type of methods lies in properly modeling the similarity between the query table and the tables from the pool. 
Generation-based methods can effectively leverage pre-existing knowledge acquired from pre-training to augment the input tables, without requiring additional data sources. 
Both methods can be further subdivided by the \texttt{level} of the augmentation, as listed in Table~\ref{preliminaries-pipeline-aug}.
Section~\ref{section:aug} will cover the typical techniques for these TDA tasks.

\begin{table}[!htbp]
\caption{Overview of the TDA tasks. 
The superscripts \textsc{r} and \textsc{g} indicate the TDA task is whether retrieval-based and generation-based, respectively.}%
\label{preliminaries-pipeline-aug}
\centering
\footnotesize
\begin{tabular}{@{}cccl@{}}
\toprule
Method                                                                       & \texttt{level}  & TDA Task                      & Description                                                   \\ \midrule
\multirow{4}{*}{\begin{tabular}[c]{@{}c@{}}Retrieval-\\ based\end{tabular}}  & \texttt{row}    & Entity Augmentation, \REA{}  & add rows retrieved from the table pool               \\
                                                                             & \texttt{column} & Schema Augmentation, \RSA{}  & add columns retrieved from the table pool                     \\
                                                                             & \texttt{cell}   & Cell Completion, \RCC{}      & complete empty cells with values retrieved from the table pool \\
                                                                             & \texttt{table}  & Table Integration, \RTI{}    & integrate retrieved tables from the pool with the original table  \\ \midrule
\multirow{4}{*}{\begin{tabular}[c]{@{}c@{}}Generation-\\ based\end{tabular}} & \texttt{row}    & Record Generation, \GRG{}    & generate new records (rows) by generative models              \\
                                                                             & \texttt{column} & Feature Construction, \GFC{} & construct new features (columns) by generative models         \\
                                                                             & \texttt{cell}   & Cell Imputation, \GCI{}      & impute empty cells by generative models                       \\
                                                                             & \texttt{table}  & Table Synthesis, \GTS{}      & synthesize rows and columns by generative models              \\ \bottomrule
\end{tabular}
\end{table}

\paragraph{\bf (3) Post-Augmentation}
Finally, the TDA pipeline includes a post-augmentation procedure that occurs primarily after ML model training. This involves evaluating and optimizing the augmented TDA results to enhance the performance of the downstream ML task.
We focus on three key aspects in post-augmentation: \emph{TDA datasets}, \emph{evaluation policies}, and \emph{optimization strategies}.
We first elaborate on the commonly used datasets associated with TDA and these datasets' characteristics.
We then analyze evaluation policies from two perspective: original-table-based evaluation, which compares the augmented table with the original one (e.g, comparing their statistical distributions), and model-based evaluation, which compares the performance of ML models trained on augmented datasets versus baseline datasets.
We finally delve into the optimization strategies. The downstream ML model can iteratively optimize augmented results by retaining data that improves performance until a target accuracy is reached. More complex strategies, like reinforcement learning (RL) based frameworks, can also guide the data optimization process.
Section~\ref{section:postaug} will detail post-augmentation techniques from these three aspects.

\section{Techniques in Pre-Augmentation}
\label{section:preaug}

In this section, we review the techniques used in the pre-augmentation procedure, as introduced in Section~\ref{subsection:preliminaries-pipeline}.
As shown in Table~\ref{preaug-main-table}, we have selected a collection of representative TDA works and summarized the pre-augmentation tasks they involve. 
Most of the selected works are oriented to TDA and have been published in well-known conferences or journals with high citation counts, reflecting their significance within the field. 
We have also included several target tasks other than TDA (see the rightmost part of Table~\ref{preaug-main-table}), namely table search~\cite{castro_fernandez_aurum_2018} and semantics detection~\cite{zhang_sato_2020}, as these often serve as intermediate steps in TDA.

Our task-oriented approach, illustrated in Table~\ref{preliminaries-pipeline-preaug}, examines four pre-augmentation tasks for the single-table setting (Sections~\ref{section:preaug-error-handling} to~\ref{section:preaug-table-representation}) and four for the multi-table setting (Sections~\ref{section:preaug-table-indexing} to~\ref{section:preaug-entity-matching}).
Pre-augmentation is essential for most TDA works, and the pre-augmentation tasks in Table~\ref{preliminaries-pipeline-preaug} are not mutually exclusive. A TDA work may involve one or more of these eight tasks. 
For example, \texttt{Infogather}~\cite{yakout_infogather_2012} (No.6 work in Table~\ref{preaug-main-table}) employs multiple pre-augmentation tasks (table representation, table annotations, etc.) to complete its entire TDA process.

\begin{sidewaystable}[!htbp]
\caption{Overview of pre-augmentation tasks. The table contains a non-exhaustive list of representative TDA-related works (arranged in chronological order) and the corresponding pre-augmentation tasks they involve. Detailed categorization of each task is provided in the respective sections.}%
\label{preaug-main-table}
\centering
\renewcommand{\arraystretch}{0.05}
\resizebox{\linewidth}{!}{
\begin{tabular}{c|l|c|cccccccccccccccc|cccccc}
\toprule
No. & \multicolumn{1}{c|}{Reference}                           & Pub.\,Year & \multicolumn{16}{c|}{Pre-augmentation Methods}                                                                                                                                                                                                                                                                                                                 & \multicolumn{6}{c}{Target Tasks}                                                                                                        \\ \midrule
    & \multicolumn{1}{c|}{}                                    &          &                      & Error Handling &                      & Table Annotation    &                      & Table Simplification &                      & Table Representation     &                      & Table Indexing &                      & Table Navigation       &                      & Schema Matching   &                      & Entity Matching &                      & Augmentation         &                      & Table Search         &                      & Semantics Detection  \\ \midrule
1   & Limaye et al.~\cite{limaye_annotating_2010} &2010  &                      & $\smallsetminus$  &                      & Ontology            &                      & $\smallsetminus$        &                      & Content                  &                      & $\smallsetminus$  &                      & $\smallsetminus$          &                      & $\smallsetminus$     &                      & $\smallsetminus$   &                      &                      &                      & \Checkmark            &                      & \Checkmark            \\
2   & Venetis et al.~\cite{venetis_recovering_2011} &2011 &                      & $\smallsetminus$  &                      & Ontology            &                      & $\smallsetminus$        &                      & Content                  &                      & $\smallsetminus$  &                      & $\smallsetminus$          &                      & Textual           &                      & $\smallsetminus$   &                      &                      &                      & \Checkmark            &                      & \Checkmark            \\
3   & \texttt{MICE}~\cite{buuren_mice_2011} &2011    &  & $\smallsetminus$  &                      & $\smallsetminus$       &                      & $\smallsetminus$        &  & Content                  &  & $\smallsetminus$  &                      & $\smallsetminus$          &                      & $\smallsetminus$     &  & $\smallsetminus$   &  & \Checkmark            &  &  &  &  \\
4   & \texttt{MissForest}~\cite{stekhoven_missforestnon-parametric_2012} &2012   &  & $\smallsetminus$  &                      & $\smallsetminus$       &                      & $\smallsetminus$        &  & Content                  &  & $\smallsetminus$  &                      & $\smallsetminus$          &                      & $\smallsetminus$     &  & $\smallsetminus$   &  & \Checkmark            &  &  &  &  \\
5   & Sarma et al.~\cite{das_sarma_finding_2012}  &2012  &                      & $\smallsetminus$  &                      & $\smallsetminus$       &                      & $\smallsetminus$        &                      & Content+Metadata         &                      & $\smallsetminus$  &                      & $\smallsetminus$          &                      & Textual+Metadata  &                      & KB              &                      &                      &                      & \Checkmark            &                      &                      \\
6   & \texttt{InfoGather}~\cite{yakout_infogather_2012} &2012    &                      & $\smallsetminus$  &                      & $\smallsetminus$       &                      & Summarization        &                      & Content+Metadata         &                      & Inverted index &                      & $\smallsetminus$          &                      & Textual+Metadata  &                      &        &                      & \Checkmark            &                      &                      &                      &                      \\
7   & Ahmadov et al.~\cite{ahmadov_towards_2015} &2015  &  & $\smallsetminus$  &                      & $\smallsetminus$       &  & $\smallsetminus$        &  & Content+Metadata         &  & Inverted index &  & $\smallsetminus$          &  & $\smallsetminus$     &  & $\smallsetminus$   &  & \Checkmark            &  &  &  &  \\
8   & \texttt{TabEL}~\cite{arenas_tabel_2015} &2015 &                      & $\smallsetminus$  &                      & $\smallsetminus$       &                      & $\smallsetminus$        &                      & Content                  &                      & $\smallsetminus$  &                      & $\smallsetminus$          &                      & $\smallsetminus$     &                      & DB              &                      & \Checkmark            &                      & \Checkmark            &                      &                      \\
9   & Kanter and Veeramachaneni~\cite{kanter_deep_2015} &2015  &  & $\smallsetminus$  &                      & $\smallsetminus$       &                      & $\smallsetminus$        &  & Content                  &  & $\smallsetminus$  &                      & $\smallsetminus$          &                      & $\smallsetminus$     &  & $\smallsetminus$   &  & \Checkmark            &  &  &  &  \\
10  & Christophides et al.~\cite{christophides_entity_2015} &2015 &                      & $\smallsetminus$  &                      & $\smallsetminus$       &                      & $\smallsetminus$        &                      & Content                  &                      & $\smallsetminus$  &                      & $\smallsetminus$          &                      & $\smallsetminus$     &                      & KB              &                      & \Checkmark            &                      & \Checkmark            &                      &                      \\
11  & \texttt{ExploreKit}~\cite{katz_explorekit_2016} &2016 &  & $\smallsetminus$  &                      & $\smallsetminus$       &                      & $\smallsetminus$        &  & Content                  &  & $\smallsetminus$  &                      & $\smallsetminus$          &                      & $\smallsetminus$     &  & $\smallsetminus$   &  & \Checkmark            &  &  &  &  \\
12  & \texttt{LSH Ensemble}~\cite{zhu_lsh_2016} &2016 & &$\smallsetminus$  &                      & $\smallsetminus$       &                      & $\smallsetminus$        &                      & Content                  &                      & LSH            &                      & $\smallsetminus$          &                      & $\smallsetminus$     &                      & $\smallsetminus$   &                      & \Checkmark            &                      & \Checkmark            &                      &                      \\
13  & \texttt{EntiTables}~\cite{zhang_entitables_2017} &2017 &                      & $\smallsetminus$  &                      & $\smallsetminus$       &                      & $\smallsetminus$        &                      & Content                  &                      & Inverted index &                      & $\smallsetminus$          &                      & $\smallsetminus$     &                      & KB+DB           &                      & \Checkmark            &                      &                      &                      &                      \\
14  & \texttt{TUS}~\cite{nargesian_table_2018}  &2018 &                      & Implicit       &                      & Supervised-learning &                      & $\smallsetminus$        &                      & Content                  &                      & LSH            &                      & $\smallsetminus$          &                      & Textual           &                      & $\smallsetminus$   &                      & \Checkmark            &                      & \Checkmark            &                      &                      \\
15  & \texttt{Aurum}~\cite{castro_fernandez_aurum_2018} &2018   &                      & $\smallsetminus$  &                      & $\smallsetminus$       &                      & $\smallsetminus$        &                      & Content                  &                      & $\smallsetminus$  &                      & Linkage graph          &                      & Numerical         &                      & $\smallsetminus$   &                      &                      &                      & \Checkmark            &                      &                      \\
16  & \texttt{table-GAN}~\cite{park_data_2018} &2018      &  & $\smallsetminus$  &                      & $\smallsetminus$       &                      & $\smallsetminus$        &                      & Content                  &                      & $\smallsetminus$  &                      & $\smallsetminus$          &                      & $\smallsetminus$     &  & $\smallsetminus$   &  & \Checkmark            &  &  &  &  \\
17  & \texttt{GAIN}~\cite{yoon_gain_2018} & 2018     &  & $\smallsetminus$  &                      & $\smallsetminus$       &                      & $\smallsetminus$        &                      & Content                  &                      & $\smallsetminus$  &                      & $\smallsetminus$          &                      & $\smallsetminus$     &  & $\smallsetminus$   &  & \Checkmark            &  &  &  &  \\
18  & \texttt{Table2Vec}~\cite{zhang_table2vec_2019} &2019  &                      & $\smallsetminus$  &                      & $\smallsetminus$       &                      & $\smallsetminus$        &                      & Content                  &                      & $\smallsetminus$  &                      & $\smallsetminus$          &                      & $\smallsetminus$     &                      & KB+DB           &                      & \Checkmark            &                      &                      &                      &                      \\
19  & \texttt{JOSIE}~\cite{zhu_josie_2019} &2019 &                      & $\smallsetminus$  &                      & $\smallsetminus$       &                      & $\smallsetminus$        &                      & Content                  &                      & Inverted index &                      & $\smallsetminus$          &                      & $\smallsetminus$     &                      & $\smallsetminus$   &                      & \Checkmark            &                      & \Checkmark            &                      &                      \\
20  & \texttt{Sherlock}~\cite{hulsebos_sherlock_2019} &2019           &                      & $\smallsetminus$  &                      & Supervised-learning &                      & $\smallsetminus$        &                      & Content                  &                      & $\smallsetminus$  &                      & $\smallsetminus$          &                      & $\smallsetminus$     &                      & $\smallsetminus$   &                      & \Checkmark            &                      &                      &                      &                      \\
21  & \texttt{CellAutoComplete}~\cite{zhang_auto-completion_2019} &2019  &  & $\smallsetminus$  &  & $\smallsetminus$       &  & Summarization        &  & Content+Metadata         &  & $\smallsetminus$  &                      & $\smallsetminus$          &                      & $\smallsetminus$     &  & KB+DB           &  & \Checkmark            &  &  &  &  \\
22  & \texttt{MIWAE}~\cite{mattei_miwae_2019} &2019    &  & $\smallsetminus$  &                      & $\smallsetminus$       &                      & $\smallsetminus$        &  & Content                  &  & $\smallsetminus$  &                      & $\smallsetminus$          &                      & $\smallsetminus$     &  & $\smallsetminus$   &  & \Checkmark            &  &  &  &  \\
23  & \texttt{ITS-GAN}~\cite{chen_faketables_2019}&2019  &  & $\smallsetminus$  &                      & $\smallsetminus$       &                      & $\smallsetminus$        &                      & Content                  &                      & $\smallsetminus$  &                      & $\smallsetminus$          &                      & $\smallsetminus$     &  & $\smallsetminus$   &  & \Checkmark            &  &  &  &  \\
24  & \texttt{PATE-GAN}~\cite{jordon_pate-gan_2019} &2019  &  & $\smallsetminus$  &                      & $\smallsetminus$       &                      & $\smallsetminus$        &                      & Content                  &                      & $\smallsetminus$  &                      & $\smallsetminus$          &                      & $\smallsetminus$     &  & $\smallsetminus$   &  & \Checkmark            &  &  &  &  \\
25  & \texttt{CTGAN}\cite{xu_modeling_2019} &2019    &  & $\smallsetminus$  &                      & $\smallsetminus$       &                      & $\smallsetminus$        &                      & Content                  &                      & $\smallsetminus$  &                      & $\smallsetminus$          &                      & $\smallsetminus$     &  & $\smallsetminus$   &  & \Checkmark            &  &  &  &  \\
26  & \texttt{MisGAN}~\cite{li_misgan_2019} &2019    &  & $\smallsetminus$  &                      & $\smallsetminus$       &                      & $\smallsetminus$        &                      & Content                  &                      & $\smallsetminus$  &                      & $\smallsetminus$          &                      & $\smallsetminus$     &  & $\smallsetminus$   &  & \Checkmark            &  &  &  &  \\
27  & \texttt{HI-VAE}~\cite{nazabal_handling_2020} &2020  &  & $\smallsetminus$  &                      & $\smallsetminus$       &                      & $\smallsetminus$        &  & Content                  &  & $\smallsetminus$  &                      & $\smallsetminus$          &                      & $\smallsetminus$     &  & $\smallsetminus$   &  & \Checkmark            &  &  &  &  \\
28  & \texttt{D$^3$L}~\cite{bogatu_dataset_2020} &2020  &                      & Implicit       &                      & Supervised-learning &                      & $\smallsetminus$        &                      & Content                  &                      & LSH            &                      & $\smallsetminus$          &                      & Textual+Numerical &                      & $\smallsetminus$   &                      & \Checkmark            &                      & \Checkmark            &                      &                      \\
29  & \texttt{ARDA}~\cite{chepurko_arda_2020} & 2020 &  & Explicit  &                      & $\smallsetminus$       &                      & Sampling        &  & Content           &                      & $\smallsetminus$  &  & $\smallsetminus$             &  & $\smallsetminus$     &                      & $\smallsetminus$   &  & \Checkmark            &  &            &  &  \\
30  & \texttt{EmbDI}~\cite{cappuzzo_creating_2020} &2020 &                      & Explicit       &                      & $\smallsetminus$       &                      & $\smallsetminus$        &                      & Content+Context          &                      & $\smallsetminus$  &                      & $\smallsetminus$          &                      & $\smallsetminus$     &                      & $\smallsetminus$   &                      & \Checkmark            &                      &                      &                      &                      \\
31  & \texttt{Sato}~\cite{zhang_sato_2020} &2020  &                      & Implicit       &                      & Supervised-learning &                      & Summarization        &                      & Content+Context          &                      & $\smallsetminus$  &                      & $\smallsetminus$          &                      & $\smallsetminus$     &                      & $\smallsetminus$   &                      &                      &                      &                      &                      & \Checkmark            \\
32  & Nargesian et al.~\cite{nargesian_organizing_2020} &2020 &                      & $\smallsetminus$  &                      & $\smallsetminus$       &                      & $\smallsetminus$        &                      & Content                  &                      & $\smallsetminus$  &                      & Linkage graph          &                      & $\smallsetminus$     &                      & $\smallsetminus$   &                      &                      &                      & \Checkmark            &                      &                      \\
33  & \texttt{DLN}~\cite{bharadwaj_discovering_2021} &2021     &  & $\smallsetminus$  &                      & $\smallsetminus$       &  & Sampling             &  & Content+Metadata         &  & $\smallsetminus$  &  & $\smallsetminus$          &  & $\smallsetminus$     &                      & $\smallsetminus$   &  &  &  & \Checkmark            &  &  \\
34  & \texttt{PEXESO}~\cite{dong_efficient_2021} &2021     &                      & Implicit       &                      & $\smallsetminus$       &                      & $\smallsetminus$        &                      & Content                  &                      & Inverted index &                      & $\smallsetminus$          &                      & Textual           &                      & $\smallsetminus$   &                      & \Checkmark            &                      & \Checkmark            &                      &                      \\
35  & Liu et al.~\cite{liu_automatic_2021} &2021     &  & $\smallsetminus$  &                      & $\smallsetminus$       &                      & $\smallsetminus$        &  & Content           &                      & $\smallsetminus$  &  & $\smallsetminus$             &  & $\smallsetminus$     &                      & $\smallsetminus$   &  & \Checkmark            &  &            &  &  \\
36  & \texttt{RONIN}~\cite{ouellette_ronin_2021} &2021    &                      & $\smallsetminus$  &                      & $\smallsetminus$       &                      & $\smallsetminus$        &                      & Content                  &                      & $\smallsetminus$  &                      & Hierarchical structure &                      & $\smallsetminus$     &                      & $\smallsetminus$   &                      &                      &                      & \Checkmark            &                      &                      \\
37  & \texttt{cWGAN}~\cite{engelmann_conditional_2021} &2021     &  & $\smallsetminus$  &                      & $\smallsetminus$       &                      & $\smallsetminus$        &                      & Content                  &                      & $\smallsetminus$  &                      & $\smallsetminus$          &                      & $\smallsetminus$     &  & $\smallsetminus$   &  & \Checkmark            &  &  &  &  \\
38  & \texttt{SIGRNN}~\cite{al-bahrani_sigrnn_2021} &2021      &  & $\smallsetminus$  &                      & $\smallsetminus$       &                      & $\smallsetminus$        &                      & Content                  &                      & $\smallsetminus$  &                      & $\smallsetminus$          &                      & $\smallsetminus$     &  & $\smallsetminus$   &  & \Checkmark            &  &  &  &  \\
39  & \texttt{MIGAN}~\cite{dai_multiple_2021} &2021    &  & $\smallsetminus$  &  & $\smallsetminus$       &                      & $\smallsetminus$        &                      & Content                  &                      & $\smallsetminus$  &                      & $\smallsetminus$          &                      & $\smallsetminus$     &  & $\smallsetminus$   &  & \Checkmark            &  &  &  &  \\
40  & \texttt{Leva}~\cite{zhao_leva_2022} &2022    &                      & Explicit       &                      & $\smallsetminus$       &                      & $\smallsetminus$        &                      & Content                  &                      & $\smallsetminus$  &                      & $\smallsetminus$          &                      & $\smallsetminus$     &                      & $\smallsetminus$   &                      & \Checkmark            &                      &                      &                      &                      \\
41  & \texttt{MATE}~\cite{esmailoghli_mate_2022} &2022     &                      & $\smallsetminus$  &                      & $\smallsetminus$       &                      & $\smallsetminus$        &                      & Content                  &                      & Inverted index &                      & $\smallsetminus$          &                      & $\smallsetminus$     &                      & $\smallsetminus$   &                      & \Checkmark            &                      &                      &                      &                      \\
42  & \texttt{ALITE}~\cite{khatiwada_integrating_2022}  &2022     &  & $\smallsetminus$  &  & PLMs                &  & $\smallsetminus$        &  & Content                  &  & $\smallsetminus$  &  & Clustering             &  & $\smallsetminus$     &                      & $\smallsetminus$   &  & \Checkmark            &  &  &  &  \\
43  & \texttt{AutoFeature}~\cite{liu_feature_2022}&2022    &  & $\smallsetminus$  &                      & $\smallsetminus$       &  & Sampling             &  & Content            &                      & $\smallsetminus$  &  & $\smallsetminus$          &                      & $\smallsetminus$     &  & $\smallsetminus$   &  & \Checkmark            &  &  &  &  \\
44  & Chai et al.~\cite{chai_selective_2022} &2022          &  & $\smallsetminus$  &                      & $\smallsetminus$       &                      & $\smallsetminus$        &  & Content           &                      & $\smallsetminus$  &  & Clustering             &  & $\smallsetminus$     &                      & $\smallsetminus$   &  & \Checkmark            &  & \Checkmark            &  &  \\
45  & Santos et al.~\cite{santos_sketch-based_2022} &2022    &                      & $\smallsetminus$  &                      & $\smallsetminus$       &                      & $\smallsetminus$        &                      & Content                  &                      & $\smallsetminus$  &                      & $\smallsetminus$          &                      & Numerical     &                      & $\smallsetminus$   &                      & \Checkmark            &                      & \Checkmark            &                      &                      \\
46  & \texttt{StruBERT}~\cite{trabelsi_strubert_2022} & 2022              &                      & Implicit       &                      & $\smallsetminus$       &                      & Summarization        &                      & Content+Context+Metadata &                      & $\smallsetminus$  &                      & $\smallsetminus$          &                      & $\smallsetminus$     &                      & $\smallsetminus$   &                      &                      &                      & \Checkmark            &                      &                      \\
47  & \texttt{TURL}~\cite{deng_turl_2022} &2022 &                      & Implicit       &                      & $\smallsetminus$       &                      & Summarization        &                      & Content+Context+Metadata &                      & $\smallsetminus$  &                      & $\smallsetminus$          &                      & $\smallsetminus$     &                      & $\smallsetminus$   &                      & \Checkmark            &                      &                      &                      & \Checkmark            \\
48  & Nargesian et al.~\cite{nargesian_data_2022} &2022    &                      & $\smallsetminus$  &                      & $\smallsetminus$       &                      & $\smallsetminus$        &                      & Content                  &                      & $\smallsetminus$  &                      & Linkage graph          &                      & $\smallsetminus$     &                      & $\smallsetminus$   &                      &                      &                      & \Checkmark            &                      &                      \\
49  & \texttt{Doduo}~\cite{suhara_annotating_2022}  &2022    &                      & Implicit       &                      & PLMs                &                      & $\smallsetminus$        &                      & Content+Context          &                      & $\smallsetminus$  &                      & $\smallsetminus$          &                      & $\smallsetminus$     &                      & $\smallsetminus$   &                      &                      &                      &                      &                      & \Checkmark            \\
50  & \texttt{TransTab}~\cite{wang_transtab_2022} &2022   &  & $\smallsetminus$  &                      & $\smallsetminus$       &                      & $\smallsetminus$        &                      & Content                  &                      & $\smallsetminus$  &                      & Linkage graph          &                      & $\smallsetminus$     &                      & $\smallsetminus$   &  & \Checkmark            &  &  &  &  \\
51  & \texttt{SOS}~\cite{kim_sos_2022} &2022   &  & $\smallsetminus$  &                      & $\smallsetminus$       &                      & $\smallsetminus$        &                      & Content                  &                      & $\smallsetminus$  &                      & $\smallsetminus$           &                      & $\smallsetminus$     &                      & $\smallsetminus$   &  & \Checkmark            &  &  &  &  \\
52  & \texttt{Watchog}~\cite{miao_watchog_2023} &2023    &                      & Implicit       &                      & PLMs                &                      & Summarization        &                      & Content+Context+Metadata &                      & $\smallsetminus$  &                      & $\smallsetminus$          &                      & $\smallsetminus$     &                      & $\smallsetminus$   &                      &                      &                      &                      &                      & \Checkmark            \\
53  & \texttt{DeepJoin}~\cite{dong_deepjoin_2023}  &2023 &                      & Implicit       &                      & $\smallsetminus$       &                      & Sampling             &                      & Content                  &                      & HNSW           &                      & $\smallsetminus$          &                      & Textual           &                      & $\smallsetminus$   &                      & \Checkmark            &                      & \Checkmark            &                      &                      \\
54  & \texttt{SANTOS}~\cite{khatiwada_santos_2023}  &2023  &                      & Implicit       &                      & Supervised-learning &                      & $\smallsetminus$        &                      & Content+Context          &                      & Inverted index &                      & $\smallsetminus$          &                      & Textual           &                      & $\smallsetminus$   &                      & \Checkmark            &                      & \Checkmark            &                      &                      \\
55  & \texttt{Starmie}~\cite{fan_semantics-aware_2023}  &2023 &                      & Implicit       &                      & $\smallsetminus$       &                      & Sampling             &                      & Content+Context          &                      & LSH+HNSW       &                      & $\smallsetminus$          &                      & Textual           &                      & $\smallsetminus$   &                      & \Checkmark            &                      & \Checkmark            &                      &                      \\
56  & \texttt{AUTOTUS}~\cite{hu_automatic_2023} &2023&                      & Implicit       &                      & $\smallsetminus$       &                      & Sampling             &                      & Content+Context          &                      & $\smallsetminus$  &                      & $\smallsetminus$          &                      & $\smallsetminus$     &                      & $\smallsetminus$   &                      & \Checkmark            &                      & \Checkmark            &                      &                      \\
57  & \texttt{RADA}~\cite{glass_retrieval-based_2023} &2023  &  & Implicit       &  & $\smallsetminus$       &  & $\smallsetminus$        &  & Content+Context+Metadata &  & $\smallsetminus$  &  & $\smallsetminus$          &  & $\smallsetminus$     &  & $\smallsetminus$   &  & \Checkmark            &  &                      &  &                      \\
58  & \texttt{HYTREL}~\cite{chen_hytrel_2023}  &2023 &  & Implicit       &  & $\smallsetminus$       &  & $\smallsetminus$        &  & Content+Context          &  & $\smallsetminus$  &  & $\smallsetminus$          &  & $\smallsetminus$     &  & $\smallsetminus$   &  &  &  & \Checkmark            &  & \Checkmark            \\
59  & \texttt{GOGGLE}~\cite{liu_goggle_2023} &2023 &  & $\smallsetminus$  &  & $\smallsetminus$       &  & $\smallsetminus$        &  & Content+Context          &  & $\smallsetminus$  &  & $\smallsetminus$          &  & $\smallsetminus$     &  & $\smallsetminus$   &  & \Checkmark            &  &  &  &  \\
60  & \texttt{GANBLR}~\cite{zhang_interpretable_2023} &2023 &  & $\smallsetminus$  &                      & $\smallsetminus$       &                      & $\smallsetminus$        &                      & Content                  &                      & $\smallsetminus$  &                      & $\smallsetminus$          &                      & $\smallsetminus$     &                      & $\smallsetminus$   &  & \Checkmark            &  &  &  &  \\
61  & \texttt{STaSy}~\cite{kim_stasy_2023} &2023 &  & $\smallsetminus$  &                      & $\smallsetminus$       &                      & $\smallsetminus$        &                      & Content                  &                      & $\smallsetminus$  &                      & $\smallsetminus$          &                      & $\smallsetminus$     &                      & $\smallsetminus$   &  & \Checkmark            &  &  &  &  \\
62  & \texttt{CoDi}~\cite{lee_codi_2023} &2023 &  & $\smallsetminus$  &                      & $\smallsetminus$       &                      & $\smallsetminus$        &                      & Content                  &                      & $\smallsetminus$  &                      & $\smallsetminus$          &                      & $\smallsetminus$     &                      & $\smallsetminus$   &  & \Checkmark            &  &  &  &  \\
63  & \texttt{TabCSDI}~\cite{zheng_diffusion_2023} &2023 &  & $\smallsetminus$  &                      & $\smallsetminus$       &                      & $\smallsetminus$        &                      & Content                  &                      & $\smallsetminus$  &                      & $\smallsetminus$          &                      & $\smallsetminus$     &                      & $\smallsetminus$   &  & \Checkmark            &  &  &  &  \\
64  & \texttt{GAINS}~\cite{xiao_beyond_2023}  &2023 &  & $\smallsetminus$  &                      & $\smallsetminus$       &                      & $\smallsetminus$        &                      & Content                  &                      & $\smallsetminus$  &                      & $\smallsetminus$          &                      & $\smallsetminus$     &                      & $\smallsetminus$   &  & \Checkmark            &  &  &  &  \\
65  & \texttt{MOAT}~\cite{wang_reinforcement-enhanced_2023} &2023 &  & $\smallsetminus$  &                      & $\smallsetminus$       &                      & $\smallsetminus$        &                      & Content                  &                      & $\smallsetminus$  &                      & $\smallsetminus$          &                      & $\smallsetminus$     &                      & $\smallsetminus$   &  & \Checkmark            &  &  &  &  \\
66  & \texttt{OmniMatch}~\cite{koutras_omnimatch_2024} &2024 &  & Implicit       &  & $\smallsetminus$       &                      & $\smallsetminus$        &  & Content                  &  & $\smallsetminus$  &  & Linkage graph          &  & $\smallsetminus$     &                      & $\smallsetminus$   &  &  &  & \Checkmark            &  &  \\
67  & \texttt{FeatNavigator}~\cite{liang_featnavigator_2024} &2024 &  & Implicit       &  & $\smallsetminus$       &                      & $\smallsetminus$        &  & Content                  &  & $\smallsetminus$  &  & $\smallsetminus$          &  & $\smallsetminus$     &                      & $\smallsetminus$   &  &  &  & \Checkmark            &  &  \\
68  & \texttt{RelDDPM}~\cite{liu_controllable_2024}  &2024 &  & $\smallsetminus$  &                      & $\smallsetminus$       &                      & $\smallsetminus$        &  & Content                  &                      & $\smallsetminus$  &  & $\smallsetminus$          &  & $\smallsetminus$     &  & $\smallsetminus$   &  & \Checkmark            &  &                      &  &  \\
69  & \texttt{SMARTFEAT}~\cite{lin_smartfeat_2024}  &2024 &  & Implicit       &  & $\smallsetminus$       &                      & $\smallsetminus$        &  & Content+Context          &  & $\smallsetminus$  &                      & $\smallsetminus$          &  & $\smallsetminus$     &                      & $\smallsetminus$   &  & \Checkmark            &  &  &  &  \\
70  & \texttt{DP-LLMTGen}~\cite{tran_differentially_2024}  &2024 &  & Implicit       &  & $\smallsetminus$       &                      & $\smallsetminus$        &  & Content         &  & $\smallsetminus$  &                      & $\smallsetminus$          &  & $\smallsetminus$     &                      & $\smallsetminus$   &  & \Checkmark            &  &  &  &  \\
\bottomrule
\end{tabular}
}
\end{sidewaystable}

\subsection{Error Handling}
\label{section:preaug-error-handling}

Error handling in pre-augmentation refers to the preprocessing of the dirty data in tables. 
Real-world tabular data often contain errors, such as mistakenly substituted proximal characters and unnecessary repetition of tokens in cell values.
Generally, there are three types of errors that considered in pre-augmentation, missing values~\cite{cappuzzo_creating_2020-1,zhao_leva_2022}, misspellings~\cite{hu_automatic_2023,dong_deepjoin_2023}, and numerical outliers~\cite{chepurko_arda_2020}.
When generating table representations based on token embedding with such errors, incorrect forecasts are inevitable~\cite{hu_automatic_2023}. Therefore, various methods have been developed to address tabular data errors, which can be categorized into explicit and implicit approaches.

\textbf{Explicit error handling} involves detecting and correcting errors directly. 
For example, \texttt{ARDA}~\cite{chepurko_arda_2020} handles missing data with a simple approach that uses uniform random sampling for categorical columns and the median value for numerical columns.
Cappuzzo et al.~\cite{cappuzzo_creating_2020-1} handle missing values using classical database techniques like \emph{Skolemization}~\cite{artho_skolem_2016}, which introduces a new constant or function symbol (called a Skolem function) to represent the unknown or missing value. 
Zhao and Fernandez~\cite{zhao_leva_2022} detect missing data through a voting mechanism. 
They observe that missing values are usually covered by multiple attributes while common values typically only appear under very few. 
Thus, they identify value nodes with votes exceeding a predefined threshold for different attributes as missing data.

\textbf{Implicit error handling} does not directly detect errors but instead enhances the model's robustness to them. 
For example, Hu et al.~\cite{hu_automatic_2023} propose a table noise generator in their automatic augmentation framework.
The generator introduces artificial table noise to the training data to enhance the model's ability to handle potential errors. 
They generate cell value noise in three ways: (1) replacing/inserting characters with proximal characters, (2) deleting/repeating characters, and (3) changing the numeral display format such as using scientific notation. 
Furthermore, multiple studies~\cite{nargesian_table_2018,bogatu_dataset_2020,zhang_sato_2020,dong_efficient_2021,dong_deepjoin_2023,trabelsi_strubert_2022,deng_turl_2022,suhara_annotating_2022,miao_watchog_2023,khatiwada_santos_2023,fan_semantics-aware_2023,hu_automatic_2023,glass_retrieval-based_2023,chen_hytrel_2023,koutras_omnimatch_2024,lin_smartfeat_2024,liang_featnavigator_2024,tran_differentially_2024} adopt word embedding that can tolerate misspellings to some extent. 
For example, Dong et al.~\cite{dong_deepjoin_2023} support semantic joins by using word table representation learning to join cells with similar meanings, thus handling data with misspellings and discrepancies in formats or terminologies.

\remarkbox{
Real-world tables often contain various types of errors, making it essential for table augmentation models to be robust in combating such issues. 
Explicit methods require an established procedure of error detection and correction, adding additional steps and potentially leading to error propagation. 
On the other hand, implicit methods, while more streamlined, lack the interpretability of explicit methods. 
Moreover, both methods typically ignore addressing errors in numerical columns, such as out-of-bound values, which are often harder to detect and correct than textual errors.
It is worth mentioning a recent trend of exploring the use of pre-trained language models (PLMs) for table representation. These approaches~\cite{fan_semantics-aware_2023,hu_automatic_2023} show certain resistance to spelling errors and offer a promising direction for handling noisy data. 
}

\subsection{Table Annotation}
\label{section:preaug-table-annotation}

Table annotation involves inferring metadata information about a table, such as column types and the relationship between columns~\cite{suhara_annotating_2022}. 
This task helps recover the semantic information within a table and is particularly useful for table augmentation by assessing the similarity between tables. 
Typically, metadata for tables is unreliable or incomplete due to inappropriate data sharing methods~\cite{fan_table_2023}. 
Even when metadata is available, tables from a wide range of sources can have incompatible metadata with different naming conventions and terminologies. 
Consequently, table annotation is crucial for retrieving syntactically and semantically relevant tables to augment the original table.
These approaches are divided into the three following categories.

\begin{figure}[!htbp]
  \centering
  \includegraphics[width=\linewidth]{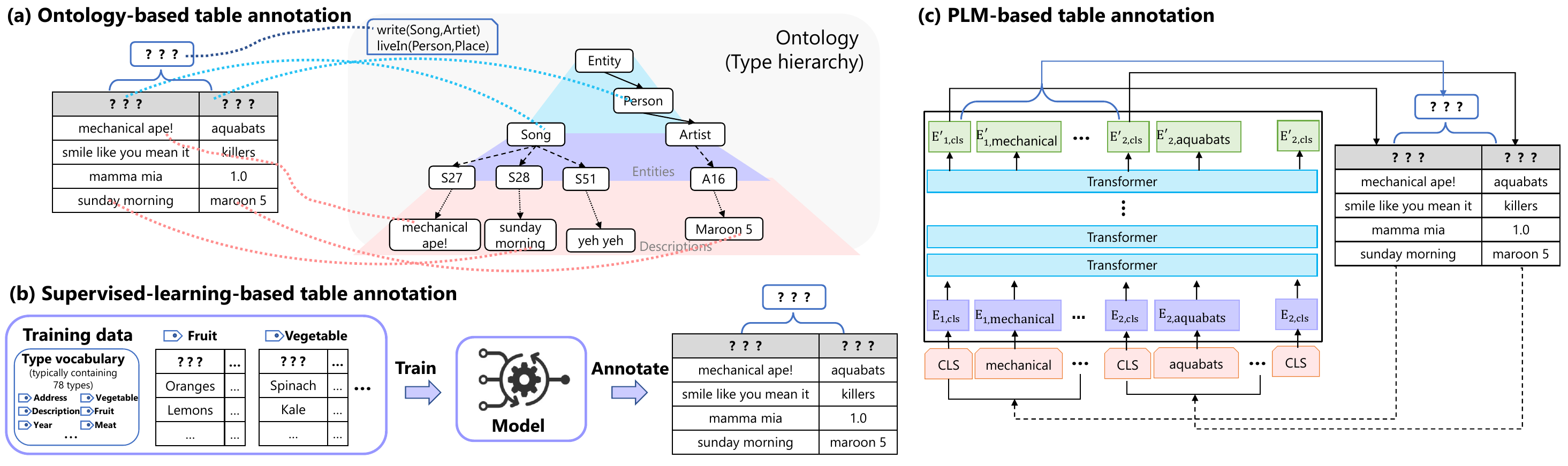}
  \caption{The illustration of table annotation, including (a) ontology-based table annotation, (b) supervised-learning-based table annotation, and (c) PLM-based table annotation.}
  \label{f4_preaug_table_annotation}
  \Description{}
\end{figure}

\textbf{Ontology-based table annotation}. 
In earlier research, ontologies were frequently used for annotating table elements.
For example, as illustrated in Fig.~\ref{f4_preaug_table_annotation} (a), Limaye et al.~\cite{limaye_annotating_2010} develop a classic table discovery system that utilizes an ontology for table annotation at multiple levels.
This includes cell-level annotation with ontology entities (e.g., labeling the cell ``mechanical ape!'' as a song name), column-level annotation with ontology types (e.g., identifying the column header ``Title'' as ``Song''), and pairwise column annotation with ontology relationships (e.g., annotating the relationship between ``Artist'' and ``Song'' as ``written-by'').
Similarly, Venetis et al.~\cite{venetis_recovering_2011} annotate columns using class labels from a ``isA'' database for column labels and binary relationships automatically extracted from the Web for relationship labels.

\textbf{Supervised-learning-based table annotation}. 
Recent studies have leveraged labeled data and supervised learning techniques for table annotation. 
Specifically, researchers~\cite{hulsebos_sherlock_2019,zhang_sato_2020} train ML models on tables labeled with a fixed set of 78 types (address, description, year, etc.) and then use the trained models to annotate unknown tables, as illustrated in Fig.~\ref{f4_preaug_table_annotation} (b). 
On the other hand, feature-based approaches~\cite{nargesian_table_2018,bogatu_dataset_2020,khatiwada_santos_2023} that compute several similarity signals (e.g., value overlap) to learn column representations have been used to detect semantic types.
\texttt{SATO}~\cite{zhang_sato_2020} further enhances semantic type detection at the column level by incorporating the topic of table (a.k.a. \emph{global context}\footnote{In the context of learning from tabular data, global context refers to the values from the entire table, whereas local context pertains to values from neighboring table elements. For example, for a specific row in the original table, the rows immediately above and below it represent the local context, while the entire original table represents the global context.}) as a new similarity signal for column representation.

\textbf{PLM-based table annotation}.
More recently, PLMs have been employed for table annotation due to their superior performance~\cite{khatiwada_integrating_2022,suhara_annotating_2022,miao_watchog_2023}.
For example, \texttt{ALITE}~\cite{khatiwada_integrating_2022} first leverages PLMs to encode columns, and then these column embeddings are used for clustering to determine the column IDs.
Suhara et al.~\cite{suhara_annotating_2022} propose \texttt{Doduo}, a multi-task learning framework, as illustrated in Fig.~\ref{f4_preaug_table_annotation} (c). Specifically, \texttt{Doduo} uses a single BERT model to complete both tasks of predicting the column types and the relationships. 
Additionally, \texttt{Doduo} incorporates the entire table as input to capture the table context. 
Despite recent efforts, existing methods rely heavily on large-scale and high-quality labeled data. 
Miao and Wang~\cite{miao_watchog_2023} propose \texttt{Watchog}, a lightweight framework for column annotation.
\texttt{Watchog} employs contrastive learning techniques to learn robust representations for tables while maintaining minimal overhead, leveraging a large-scale unlabeled table pool.

\remarkbox{
Real-world tables often have incomplete and incorrect metadata, necessitating table annotation to recover table semantics. 
Ontology-based table annotation can be limited by the ontology's coverage, particularly for domain-specific data, while supervised-learning table annotation depend heavily on large-scale, high-quality labeled data. 
Ontology-based methods typically offer high efficiency, whereas learning methods (including those based on supervised learning and those based on PLMs) achieve higher accuracy.
One potential direction for improvement involves integrating these methods and exploring the efficient use of PLMs for table annotation. Additionally, integrating LLMs and Retrieval-Augmented Generation (RAG)~\cite{zhang_raft_2024} with ontologies can lead to promising developments, particularly in domain-specific scenarios where leveraging external business knowledge is crucial.
}

\subsection{Table Simplification}
\label{section:preaug-table-condensation}

Table simplification involves streamlining a table down to its essential elements, which can be addressed from both the content and semantic perspectives.
From the table content perspective, this procedure, known as \emph{table sampling}, involves selecting portions of the table to retain as much information as possible. This is particularly useful for fitting data into limited token lengths for language models.
From the table semantics perspective, the procedure, referred to as table summarization, entails identifying the main topic or theme of the table to better understand its meaning.
Accurate summarization helps in comparing tables for similarity and ensures that any added rows and columns remain consistent with the table's original theme.
These two different perspectives are introduced as follows.

\textbf{Table sampling} selectively choose table content to preserve the original information as much as possible.
An early work~\cite{bharadwaj_discovering_2021} directly selects the top-$K$ samples for each column as input. \texttt{AutoFeature}~\cite{liu_feature_2022} and \texttt{ARDA}~\cite{chepurko_arda_2020} adopts the stratified sampling method that divides the samples into several subsets and randomly selects samples from each subset.
\texttt{Deepjoin}\cite{dong_deepjoin_2023} adopts a frequency-based approach that samples the most frequent cell values for each column.
Meanwhile, \texttt{Starmie}~\cite{fan_semantics-aware_2023} and \texttt{AUTOTUS}~\cite{hu_automatic_2023} sample rows based on an importance score derived from the cell-level TF-IDF score for each row.

\textbf{Table summarization} aims to extract semantic themes or topics within tables. 
For example, Zhang et al.~\cite{zhang_sato_2020} incorporate a topic-aware prediction module into their framework \texttt{SATO}, responsible for summarizing table semantics. 
In particular, this module produces a topic vector from the values across the entire table, representing the global context of that table. 
\texttt{SATO}'s ablation experiments show that considering a table's global context improves table understanding and mitigates ambiguities.
For web tables, there are often pre-existing table metadata, such as table captions that summarize the table's contents~\cite{zhang_web_2020}. 
In this case, various works~\cite{yakout_infogather_2012,zhang_auto-completion_2019,deng_turl_2022,trabelsi_strubert_2022,miao_watchog_2023,glass_retrieval-based_2023} directly leverage these metadata by converting them into vectors and concatenating them with the table's representation.

\remarkbox{
Table simplification aims to extract the core information of a table either in content or semantic level. 
At the content level, table sampling selectively samples table content for constraints like limited token length for LLMs.
Table sampling can also improve efficiency and scalability when facing large-scale tables.
At present, most table sampling methods are statistical-based which are fast and resource-efficient, yet they fall short in terms of precision. Sampling based solely on the value distribution (such as sampling the most frequent values), can lead to an incomplete representation of the original table. Therefore, there is a need to investigate more efficient and accurate sampling algorithms, potentially leveraging learning-based approaches.
At the semantic level, table summarization serves both to extract the key semantic essence of the table and verify the coherence of any table augmentations. At present, as a preprocessing step, there are relatively few techniques that directly extract the primary topics or themes of tables. This is largely due to the challenge of achieving lightweight, efficient table summaries. Emerging LLMs may offer a promising solution in this regard. 
}

\subsection{Table Representation}
\label{section:preaug-table-representation}

Table representation involves converting the table elements such as rows, columns, and cells into a latent vector space.
This transformation prepares the table for robust use in subsequent TDA model. 
The past decade has witnessed the effectiveness of employing deep neural networks for table representation learning~\cite{nargesian_table_2018,bogatu_dataset_2020,zhang_sato_2020,dong_efficient_2021,deng_turl_2022,suhara_annotating_2022,khatiwada_integrating_2022,miao_watchog_2023,fan_semantics-aware_2023,hu_automatic_2023,dong_deepjoin_2023}, thereby enhancing TDA. 
The basic idea of table representation is to create vector representations of tables that preserve their syntax and semantics.
These vectors can then be compared using methods like cosine similarity to assess their relatedness. 
The effectiveness of these table representations depends on the information source they contain, which can be summarized as table content, table context, and metadata.
We briefly introduce each type of approach as follows.

\textbf{Content-based table representation}.
Typically, several works~\cite{yakout_infogather_2012,nargesian_table_2018,bogatu_dataset_2020,koutras_omnimatch_2024} derive features from table content, such as column distribution, column type, to create table representations.
Also, many studies~\cite{park_data_2018,yoon_gain_2018,chen_faketables_2019,xu_modeling_2019,zheng_diffusion_2023,liu_controllable_2024} directly transform table content into vectors. For example, discrete columns can be represented as one-hot vectors~\cite{xu_modeling_2019,zheng_diffusion_2023}, analog bits vectors~\cite{zheng_diffusion_2023,liu_controllable_2024}, and embedding vectors~\cite{zheng_diffusion_2023}.
More recently, there are increasing studies~\cite{deng_turl_2022,trabelsi_strubert_2022,fan_semantics-aware_2023,dong_deepjoin_2023} leveraging language models to encode tables. They first sequentialize table content, and then feed the sequential text into language models for encoding.

\textbf{Context-based table representation}. 
Recent research~\cite{zhang_sato_2020,cappuzzo_creating_2020,suhara_annotating_2022,khatiwada_santos_2023,hu_automatic_2023,fan_semantics-aware_2023,miao_watchog_2023,glass_retrieval-based_2023,chen_hytrel_2023} underscores the importance of contextual information, such as column relationships~\cite{suhara_annotating_2022,khatiwada_santos_2023,hu_automatic_2023}, alongside table content.
For instance, Khatiwada et al.~\cite{khatiwada_santos_2023} enhance the \texttt{SANTOS} model by incorporating semantic relationships between column pairs, refining its comprehension of table contexts and filtering out tables with similar columns but different contexts during the TDA process.
Hu et al.~\cite{hu_automatic_2023} utilize PLMs (e.g., BERT) to capture the contextual relationships between columns, aiding in the identification of relevant tables in the table pool for TDA.
\texttt{SATO}~\cite{zhang_sato_2020} employs a hybrid model to utilize signals from both the global context (values from the entire table) and the local context (predicted types of neighboring columns). 
Similarly, several studies~\cite{suhara_annotating_2022,fan_semantics-aware_2023,miao_watchog_2023} capture the global context of the table by feeding the entire table into column encoders, resulting in the encoded column vectors that contain distilled global table context.
Several works~\cite{cappuzzo_creating_2020,chen_hytrel_2023,liu_goggle_2023} leverage graphs to represent tables, with nodes representing cell values and edges indicating relationships between nodes (e.g., cell belonging to one specific column), aiming to better capture the table structure.
These approaches have demonstrated high prediction accuracy, highlighting the effectiveness of integrating table context into table representations.

\textbf{Metadata-based table representation}. 
In addition to using data values organized in rows and columns, several studies~\cite{das_sarma_finding_2012,yakout_infogather_2012,zhang_auto-completion_2019,deng_turl_2022,trabelsi_strubert_2022,miao_watchog_2023,glass_retrieval-based_2023} also utilize metadata. 
Tables are often accompanied by secondary information such as the table captions and the containing webpage's title, forming part of the textual information~\cite{trabelsi_strubert_2022}. 
Yakout et al.~\cite{yakout_infogather_2012} consider element-wise and cross-element similarities, where elements encompass both the table content (e.g., tuples and column values) and metadata (e.g., table captions and URLs). 
\texttt{Watchog}~\cite{miao_watchog_2023} improves upon \texttt{Starmie}~\cite{fan_semantics-aware_2023} by incorporating metadata such as headers, captions, and topics.

\remarkbox{
Table representation forms the basis of tabular data processing and is not limited solely to TDA tasks. 
Consequently, numerous studies have concentrated on this domain. 
Early approaches that relied on table content were found inadequate. 
Current research typically leverages both the content and contextual information of tables to create richer table representations, yielding improved outcomes. 
The use of LLMs has further improved the extraction of semantic information from tables. 
However, a notable challenge remains in accurately capturing the structural information of tables, such as row/column permutations invariance.
}

\subsection{Table Indexing}
\label{section:preaug-table-indexing}

Table indexing involves assigning a unique identifier (index value) to table elements, allowing for quick and efficient lookup and retrieval based on their index values. 
Many retrieval-based TDA methods use indexes to enhance efficiency and scalability, particularly when dealing with large-scale table pools with millions of tables. 
Researchers have utilized various types of indexes, such as the inverted index~\cite{yakout_infogather_2012,ahmadov_towards_2015,zhang_entitables_2017,dong_efficient_2021,esmailoghli_mate_2022,zhu_josie_2019,khatiwada_santos_2023}, Locality Sensitive Hashing (LSH) index~\cite{nargesian_table_2018,fan_semantics-aware_2023,bogatu_dataset_2020,zhu_lsh_2016,castro_fernandez_aurum_2018}, and graph index such as Hierarchical Navigable Small World (HNSW)~\cite{malkov_efficient_2020,dong_deepjoin_2023,fan_semantics-aware_2023}. 

\begin{figure}[!htbp]
  \centering
  \includegraphics[width=\linewidth]{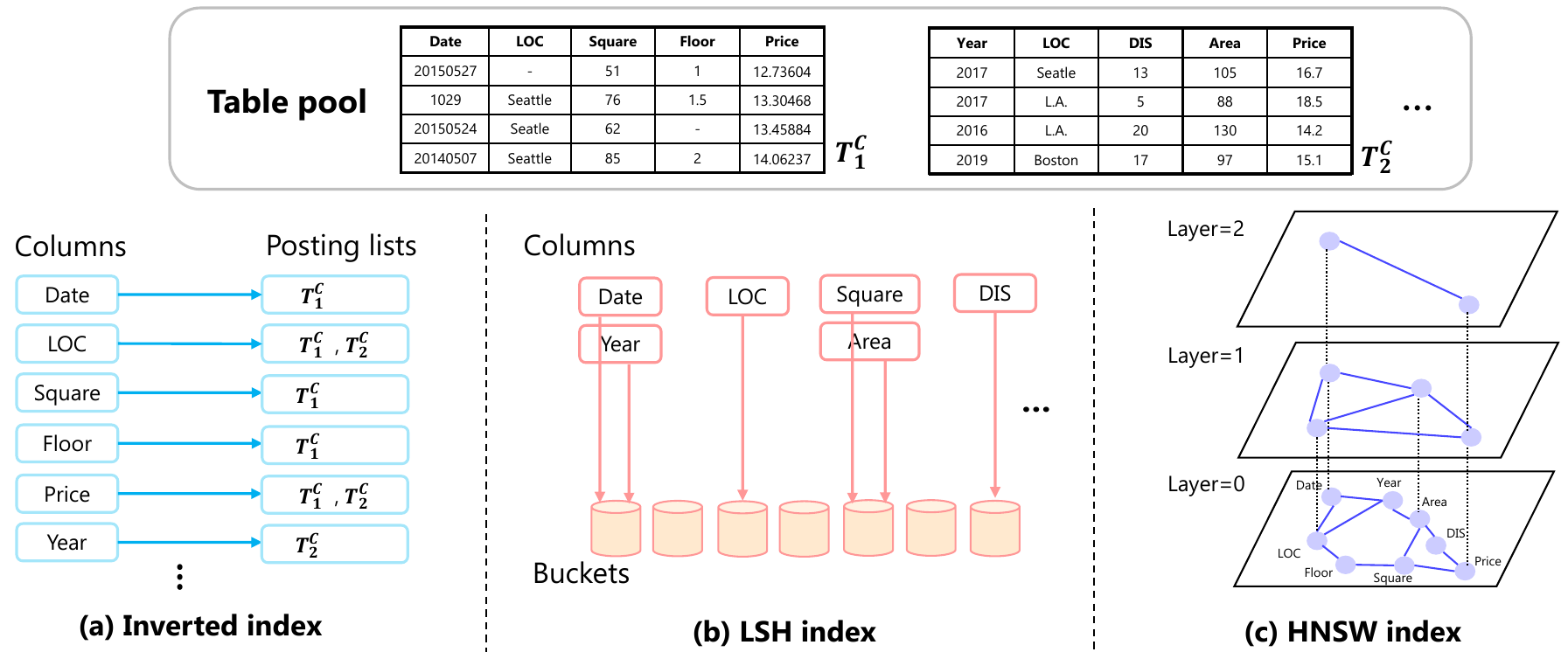}
  \caption{The illustration of table indices, including (a) inverted index, (b) LSH index, and (c) HNSW index.}
  \label{f5_preaug_table_indexing}
  \Description{}
\end{figure}

\textbf{Inverted index} is a structure of posting lists that maps each distinct value to its containing structures, such as tables (see Fig.~\ref{f5_preaug_table_indexing} (a)), rows, columns, or cells. 
When searching for candidates, the inverted index requires reading all or a large subset of these posting lists, which can lead to memory management issues and long read times in large-scale table pools. 
To address this, \texttt{JOSIE}~\cite{zhu_josie_2019} incorporates a dictionary for quick access to the posting lists, avoiding the need to search the entire inverted index. 
\texttt{Mate}~\cite{esmailoghli_mate_2022} extends the single-attribute inverted index by adding a super key, a fixed-size hash value that aggregates all possible key value combinations into a single entry. This approach enables efficient multi-attribute join discovery without the prohibitive storage requirements of a full multi-attribute index.

\textbf{LSH index} has been widely adopted in approximate nearest neighbor search in high-dimensional spaces~\cite{nargesian_table_2018,zhu_lsh_2016}.
The basic idea is that similar vectors are more likely to be hashed into the same bucket~\cite{fan_semantics-aware_2023} (see Fig.~\ref{f5_preaug_table_indexing} (b)), which significantly improves query time with minimal accuracy loss.
For instance, \texttt{D$^3$L}~\cite{bogatu_dataset_2020} uses an extension of the LSH index to ensure that search time remains constant regardless of repository size. 
\texttt{Starmie}~\cite{fan_semantics-aware_2023} adjusts the LSH index using the simHash function to better estimate the similarity between column embedding vectors.

\textbf{HNSW index} is a graph-based approximate nearest neighbor search technique designed for high-dimensional data. 
HNSW involves a hierarchy of graphs, where each layer's graph is built on top of the previous layer, allowing for efficient search: a query can start in any layer and traverse the graph to find the nearest neighbors, as illustrated in Fig.~\ref{f5_preaug_table_indexing} (c). 
The HNSW index has recently been applied to accelerating the table retrieval process for TDA, reportedly achieving a 400$\times$ performance gain over the LSH index~\cite{fan_semantics-aware_2023}. 

\remarkbox{
Table indexing substantially improves the efficiency and scalability of table retrieval for TDA. 
While inverted indexes provide high accuracy, they can encounter memory management issues and prohibitively long read times in large-scale table pools.
To address these limitations, recent works~\cite{nargesian_table_2018,bogatu_dataset_2020,fan_semantics-aware_2023} have explored techniques like LSH and HNSW for table indexing, which can significantly improve efficiency and maintain accuracy. 
Furthermore, some study~\cite{fan_semantics-aware_2023} has adopted a hybrid approach, combining both LSH and HNSW indexing techniques to utilize the strengths of each method. 
However, most existing approaches are designed for static table pools. In today's information age, table pools are continuously evolving. Adapting table indexing methods to handle dynamic, ever-changing data remains a major challenge that has yet to be fully addressed.
}

\subsection{Table Navigation}
\label{section:preaug-table-navigation}

Table navigation involves establishing a navigational framework over a table pool. 
Essentially, it refers to organizing the table pool in a way that highlights connections between similar tables, such as through edges in a graph or by clustering them together.
With table navigation, the subsequent TDA can retrieve relevant data for augmentation more easily and efficiently.
Existing works typically employ cluster structures~\cite{chai_selective_2022,khatiwada_integrating_2022}, hierarchical structures~\cite{ouellette_ronin_2021}, or linkage graphs~\cite{nargesian_organizing_2020,nargesian_data_2022,castro_fernandez_aurum_2018} to manage tables in table pools.

\begin{figure}[!htbp]
  \centering
  \includegraphics[width=\linewidth]{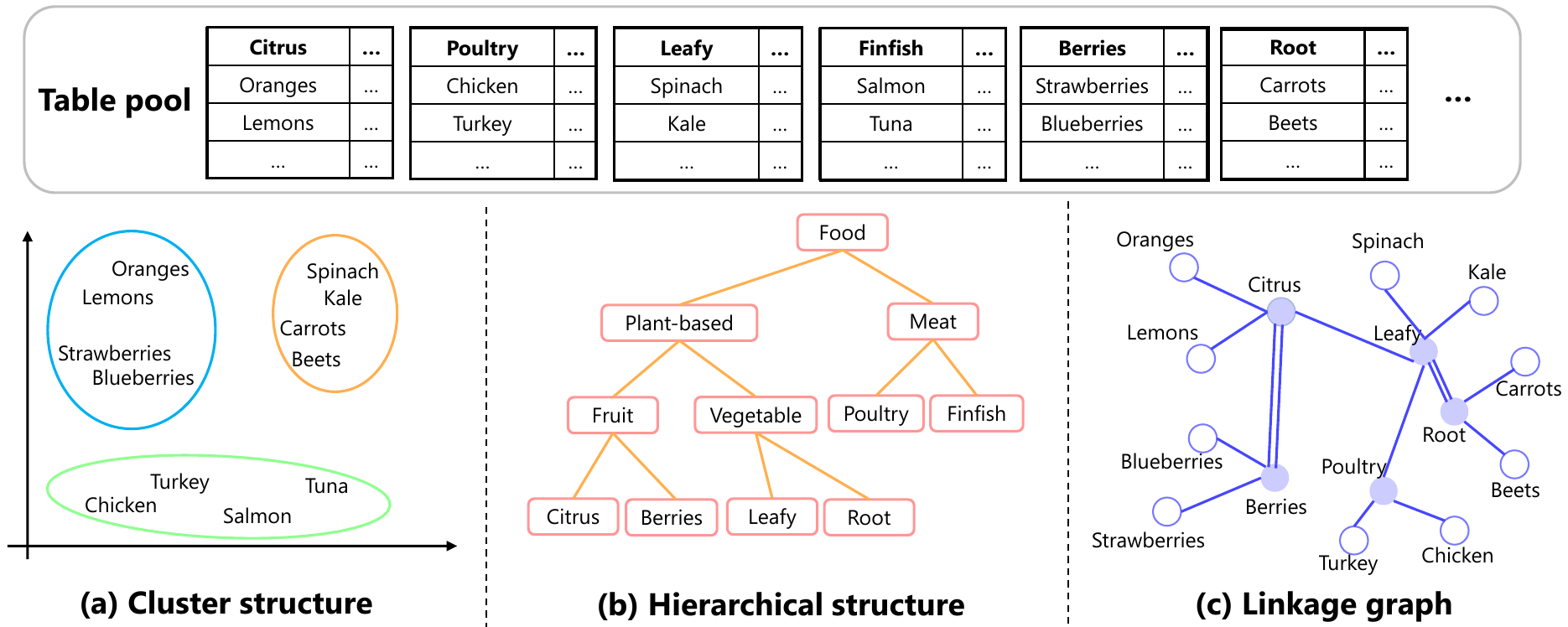}
  \caption{The illustration of table navigation techniques, including (a) cluster structure, (b) hierarchical structure, and (c) linkage graph.}
  \label{f6_preaug_table_navigation}
  \Description{}
\end{figure}

\textbf{Cluster structure} groups related tables in the table pool.
For example, Chai et al.~\cite{chai_selective_2022} employ Multivariate Gaussian Mixture Model to cluster similar data points (rows), as illustrated in Fig.~\ref{f6_preaug_table_navigation} (a), aiding in the downstream data point selection and augmentation tasks.
To facilitate the subsequent column integration, \texttt{ALITE}~\cite{khatiwada_integrating_2022} clusters similar columns using hierarchical clustering by iteratively merging the closest clusters based on Euclidean distances between column embeddings. This method can also be considered as a hierarchical structure to some extent.

\textbf{Hierarchical structure} allows navigation from broader concepts to more specific ones, as demonstrated in Fig.~\ref{f6_preaug_table_navigation} (b).
For instance, Ouellette et al.~\cite{ouellette_ronin_2021} propose a hierarchical structure called \texttt{RONIN} for organizing the tables in the pool. This model allows users to narrow down to potential datasets by navigating between groups of attributes sets connected by edges indicating subset relationships in a hierarchical manner.
\texttt{RONIN} seamlessly integrates table search and table navigation, making the acquisition of related table more effective. 
 
\textbf{Linkage graph} uses edges to indicate the relevance between tables in the table pool, as shown in Fig.~\ref{f6_preaug_table_navigation} (c). 
For instance, Nargesian et al.~\cite{nargesian_organizing_2020,nargesian_data_2022} make use of Directed Acyclic Graphs (DAG) to structure the table pool.
In the DAG, nodes represent attribute sets, with the labels of a non-leaf node summarizing the content of the attribute sets in their corresponding subgraph.
Another approach \texttt{Aurum}~\cite{castro_fernandez_aurum_2018} constructs an enterprise knowledge graph, with nodes representing attributes within the table pool, edges indicating relationships between two nodes. 
Specifically, \texttt{Aurum} also contains hyperedges connecting any number of nodes hierarchically related, such as attributes of the same table, or tables of the same table pool, thereby facilitating the exploration of relevant data items. 
More recently, \texttt{OmniMatch}~\cite{koutras_omnimatch_2024} constructs a similarity graph where columns are represented as nodes. These nodes are connected by different types of edges corresponding to various features, such as embedding similarity and distribution similarity.
GNNs are then used to propagate these similarity signals to facilitate join discovery.

\remarkbox{
Table navigation restructures the table pool, allowing retrieval-based TDA to locate and access similar tables more effectively.
This concept is fairly new and continues to evolve.
Clusters offer a relatively simple method for organizing table pools, indicating whether tables have similarity or belong to the same class; however, they lack the capability to convey more complex information, such as hierarchical relationships.
For both cluster and hierarchical structure, they may not fully capture the relationships within table structures, such as the ``entity-property'' relationship between columns (e.g., entity ``person'' has the property ``gender'').
Both hierarchical structure and linkage graph face efficiency issues. 
Developing scalable and robust methods for navigating such vast table repositories remains a significant challenge in this field.
}

\subsection{Schema Matching}
\label{section:preaug-schema-matching}

Schema matching involves evaluating the relatedness between two table columns. 
In this context, the set of column headers is typically referred to as the table schema~\cite{zhang_web_2020}.
Schema matching methods are frequently adopted in retrieval-based TDA for identifying and fetching those related columns and tables to expand the features in ML models. 
Due to the varied data types within tables, schema matching methods are categorized into textual matching, numerical matching, and metadata matching.

\textbf{Textual matching} is the most commonly used schema matching technique because textual columns usually contain more information than numerical ones. 
Below is a concise overview of some common textual matching methods.
\begin{itemize}[leftmargin=*]
\item \emph{Value overlap}~\cite{yakout_infogather_2012,venetis_recovering_2011,nargesian_table_2018,bogatu_dataset_2020,fan_semantics-aware_2023,koutras_omnimatch_2024}: If there is a significant overlap in the value sets of two columns, then the columns are considered related.
\item \emph{Semantic overlap}~\cite{das_sarma_finding_2012,nargesian_table_2018,khatiwada_santos_2023,koutras_omnimatch_2024,liang_featnavigator_2024}: When leveraging table annotations (see Section~\ref{section:preaug-table-annotation}) to derive labels describing column semantics, two columns are considered a match if there is a substantial overlap between their respective labels.
\item \emph{Embedding similarity}~\cite{nargesian_table_2018,bogatu_dataset_2020,dong_efficient_2021,khatiwada_santos_2023,dong_deepjoin_2023,koutras_omnimatch_2024,liang_featnavigator_2024}: Related columns are identified by computing the similarity of their corresponding embeddings in vector space.
\end{itemize}

\textbf{Numerical matching} is concerned with numerical columns. 
These methods typically evaluate the value distribution~\cite{santos_sketch-based_2022,bogatu_dataset_2020} to derive insight from numerical data. 
For instance, \texttt{D$^3$L}~\cite{bogatu_dataset_2020} utilizes the Kolmogorov-Smirnov statistic to decide whether two numerical columns come from the same domain distribution and thus can be matched. 
Santos et al.~\cite{santos_sketch-based_2022} propose the Quadrant Count Ratio (QCR) hashing scheme. This method divides the numerical values of two columns into four quadrants based on their sign (i.e., positive or negative). Only the points in the same quadrant would be assigned with the same hash value. Columns with a certain number of matching values are likely to be correlated.

\textbf{Metadata matching} is also commonly used, as tables are usually accompanied by secondary information such as column headers. For instance, several studies~\cite{das_sarma_finding_2012,yakout_infogather_2012,bogatu_dataset_2020,miao_watchog_2023} consider two columns to be related when their column headers have a semantic overlap above a given threshold.

\remarkbox{
Schema matching is a crucial step in retrieval-based TDA, used to determine the similarity between table attributes (columns) and infer the overall similarity between tables. 
This field has been extensively studied, and different types of schema matching methods are often adopted simultaneously, yielding promising results.
However, there remain a wide range of research opportunities, particularly in the domain of numerical schema matching, such as handling different numerical display formats. 
Even current powerful LLMs cannot handle numbers well.
Moreover, the vast majority of existing work in this area has focused solely on the similarity between single columns. The similarity between combined or composite columns is rarely addressed.
}

\subsection{Entity Matching}
\label{section:preaug-entity-matching}

Entity matching involves identifying connections between entities in different tables, facilitating the localization of relevant entities and tables for retrieval-based TDA. 
These methods are particularly relevant in the context of horizontal tabular tables, where entities are typically represented as rows and their attributes are in columns. 
Based on the data source to which the entities are matched, the methods are categorized into KB-referenced entity matching and DB-referenced entity matching.

\textbf{KB-referenced entity matching} maps table entities to their referenced entity in a knowledge base (KB)~\cite{arenas_tabel_2015}, aiming to enhance the semantic understanding of tables. 
Essentially, if two table entities are related to the same KB entity, the likelihood of these two table entities being related is high.
For example, Sarma et al.~\cite{das_sarma_finding_2012} link table entities to KB entities to acquire weighted label sets for representing table entities. 
However, not all table elements can be mapped to predefined types and relationships in the referenced KB.
To address this, \texttt{TabEL}~\cite{arenas_tabel_2015} proposes an alternative way to weaken the strict mapping, using soft constraints based on graphical model to encode a higher preference for sets of related entities. 

\textbf{DB-referenced entity matching} determines whether entities and their corresponding properties from table pools (or databases) refer to the same real-world object as the entities in the original table.
Christophides et al.~\cite{christophides_entity_2015} outline a general framework for this task, comprising two main components: (1) similarity metrics, which compare entity descriptions and (2) blocking techniques, which group tables in the table pool that are approximately similar for enhanced efficiency of this process.
More recent efforts have begun to use iterative approaches, where previously discovered matching entities serve as input for computing similarities between further tables in the table pool~\cite{chapman_dataset_2020}.

\textbf{Hybrid entity matching} combines both KB- and DB-referenced approaches to obtain a wider range of information sources. 
For example, \texttt{Entitables}~\cite{zhang_entitables_2017} identifies candidate entities from both sources. Entities in KB that share categories or types with the original table entities are considered good candidates. Similarly, entities in DB tables that contain the original table entities or have related captions to the original table are also considered candidates. \texttt{Table2Vec}~\cite{zhang_table2vec_2019} enhances \texttt{Entitables} by incorporating word embeddings for table entities. 
\texttt{CellAutoComplete}~\cite{zhang_auto-completion_2019} further improves \texttt{Entitables} by carefully designing features that combine evidence from multiple sources (e.g., the similarity between table headings can be calculated using edit distance from DB or mapping probabilities from KB and DB).

\remarkbox{
Entity matching uncovers relationships between entities across different tables, facilitating the retrieval of relevant entities and tables.
For KB-referenced matching, knowledge bases may have limited coverage when applied to real-world table pools.
To address this, several works~\cite{zhang_entitables_2017,zhang_table2vec_2019,zhang_auto-completion_2019} have integrated KB- and DB-referenced approaches to broaden the range of information sources.
However, a common limitation in both KB- and DB-referenced entity matching is the assumption made in previous works that the leftmost column in a table contains the entity ID or name, which is \emph{not} always the case. 
Indeed, entity matching has seen decreased use in the past few years compared to schema matching. 
An emerging field might be the combination of LLMs and RAG to replace traditional KB-based methods for entity matching. LLMs can better capture entity semantics, reducing the reliance on the leftmost column. Meanwhile, RAG techniques can effectively leverage and update the KB, counteracting the issue of limited scope.
}

\subsection{Scenarios for Pre-augmentation Techniques}
\label{section:preaug-discussion}

\begin{figure}[!htbp]
  \centering
  \includegraphics[width=\linewidth]{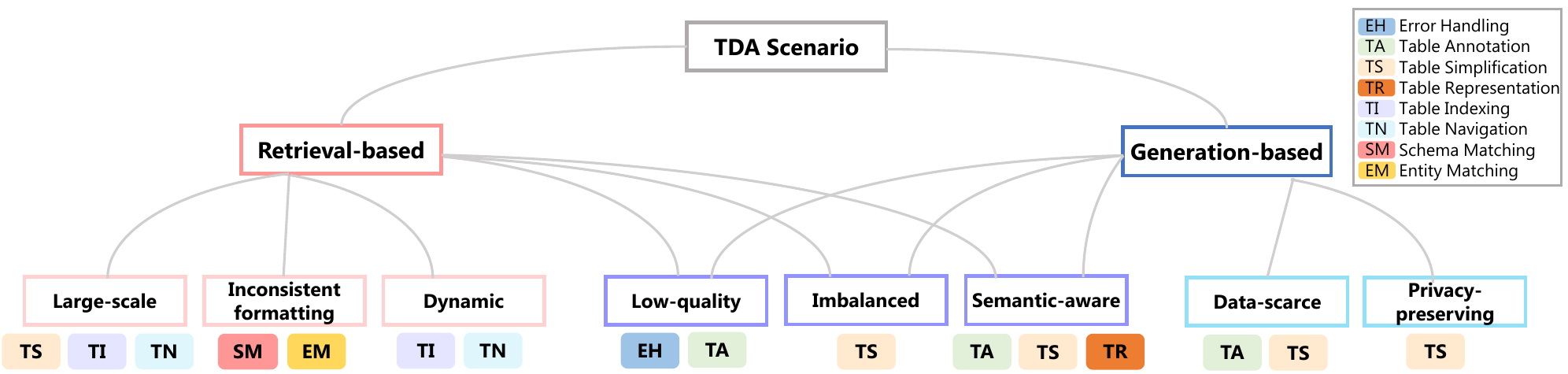}
  \caption{The illustration of different TDA scenarios and their suited pre-augmentation tasks.}
  \label{f7_preaug_discussion}
  \Description{}
\end{figure}

This section empirically summarizes the specific pre-augmentation tasks used in different TDA scenarios, as depicted in Fig.~\ref{f7_preaug_discussion}. 
The four single-table-setting pre-augmentation tasks apply to both retrieval and generation-based TDA, while the multi-table-setting tasks are only suitable for retrieval-based methods, which require an external table pool to handle multi-table relationships.

Retrieval-based TDA operates on a table pool often possessing \textbf{large-scale} data, thus leading to issues such as \textbf{inconsistent formatting} and \textbf{dynamic} data. 
Common pre-augmentation methods in these scenarios include: 
(1) table simplification for reducing table information; (2) table indexing and table navigation for fast retrieval; (3) schema matching and entity matching for addressing inconsistent formatting; and (4) table indexing and table navigation for managing dynamic and large-scale table pools.

Generation-based TDA using a single original table may suffer from issues in \textbf{data-scarce} scenarios. In such scenarios, pre-augmentation techniques like table annotation (providing additional information or labels) and table summarization (extracting key information) are necessary.
Furthermore, in \textbf{privacy-preserving} scenarios, generating synthetic data often benefits from table sampling, which involves using only partial data.
Note that both table summarization and table sampling are part of table simplification, as discussed in Section~\ref{section:preaug-table-condensation}.

Meanwhile, both retrieval- and generation-based TDA approaches face some common challenges, such as low-quality and imbalanced tables. 
In \textbf{low-quality} scenarios (e.g., tables with null or erroneous values), common preaugmentation techniques include error handling and table annotation (to annotate column when column names are missing). 
In \textbf{imbalanced} scenarios, table sampling within table simplification is a common scheme. 
Additionally, TDA often requires semantic awareness; for these \textbf{semantic-aware} scenarios, and commonly used techniques include table annotation, table summarization within table simplification, and table representation.

\section{Techniques in Tabular Data Augmentation}
\label{section:aug} 

This section will delve into the current state-of-the-art techniques for tabular data augmentation (TDA). 
As outlined in Table~\ref{preliminaries-pipeline-aug}, we first classify TDA tasks into two primary categories: retrieval-based TDA (see Section~\ref{section:retrieval_based_TDA}) and generation-based TDA (see Section~\ref{section:generation_based_TDA}). Within these two categories, the approaches can be further divided into different levels: adding rows~\cite{zhang_entitables_2017,park_data_2018} or columns~\cite{dong_deepjoin_2023,lin_smartfeat_2024}, augmenting individual cells~\cite{zhang_auto-completion_2019}, and extending the original table with both rows and columns~\cite{khatiwada_integrating_2022}. Thus, for each category, we will discuss the corresponding table augmentation work at the \texttt{row}, \texttt{column}, \texttt{cell}, or \texttt{table} level. Fig.~\ref{fig5:aug_main_table} provides a concise overview of the TDA works discussed in this section, along with their detailed taxonomy.

\tikzset{
  FARROW/.style={arrows={-{Latex[length=1.25mm, width=1.mm]}}, }, %
  U/.style = {circle, draw=melon!400, fill=melon, minimum width=1.4em, align=center, inner sep=0, outer sep=0},
  I/.style = {circle, draw=tea_green!400, fill=tea_green, minimum width=1.4em, align=center, inner sep=0, outer sep=0},
  cate/.style = {rectangle, draw, minimum width=8em, minimum height=2em, align=center, rounded corners=3}, %
  cate2/.style = {rectangle, draw, minimum width=33em, minimum height=2em, text width=33em, align=left, rounded corners=3,anchor=west},%
  cate3/.style = {rectangle, draw, minimum width=43em, text width=43em, minimum height=2em, align=left, rounded corners=3},%
  encoder/.style = {rectangle, fill=Madang!82, minimum width=10em, minimum height=3em, align=center, rounded corners=3},
}

\begin{figure}
    \centering
    \resizebox{0.9\linewidth}{!}{
    \begin{tikzpicture}
    
    \node [cate, rotate=90, align=left] (n1) at (0, -4) {\textbf{Tabular Data Augmentation Techniques}};

    \node [cate,fill=orange!20, anchor=west] (n20) at (0.7,3.2) {Retrieval-based};
    \draw[] (n1.south) -- (0.5,-4) -- (0.5,3.2) -- (n20.west);
    
    \node [cate, fill=blue!20, anchor=west] (n201) at (4.05,10.4)  {Entity\\Augmentation\\(Row-level)};
    \draw[] (n20.east) -- (3.8,3.2) -- (3.8,10.4) -- (n201.west);

    \node [cate, fill=blue!20, anchor=west] (n2001) at (7.55,12.6)  {Statistical};
    \draw[] (n201.east) -- (7.15, 10.4)-- (7.15,12.6)--  (n2001.west);
    \node [cate2, fill=blue!20, anchor=west] (n20001) at (10.7,12.6)  {\texttt{Infogather}~\cite{yakout_infogather_2012} uses TF-IDF score; \texttt{TUS}~\cite{nargesian_table_2018} and \texttt{D$^3$L}~\cite{bogatu_dataset_2020} utilize statistical models to calculate features.};
    \draw[] (n2001.east) --  (n20001.west);

    \node [cate, fill=blue!20, anchor=west] (n2002) at (7.55,11)  {KB-based};
    \draw[] (n201.east) -- (7.15, 10.4)-- (7.15, 11)--  (n2002.west);
    \node [cate2, fill=blue!20, anchor=west] (n20002) at (10.7,11)  {Das Sarma et al.~\cite{das_sarma_finding_2012} label table entities with KB; \texttt{EntiTables}~\cite{zhang_entitables_2017} and \texttt{Table2Vec}~\cite{zhang_table2vec_2019} identify candidate entities sharing the same types from DBpedia; \texttt{SANTOS}~\cite{khatiwada_santos_2023} uses KB to annotate columns and the relationships between columns.};
    \draw[] (n2002.east) --  (n20002.west);
    
    \node [cate, fill=blue!20, anchor=west] (n2003) at (7.5,9.4)  {Graph-based};
    \draw[] (n201.east) -- (7.15, 10.4)-- (7.15, 9.4)--  (n2003.west);
    \node [cate2, fill=blue!20, anchor=west] (n20003) at (10.7,9.4)  {\texttt{InfoGather}~\cite{yakout_infogather_2012} constructs a weighted graph with tables as nodes; \texttt{EmbDI}~\cite{cappuzzo_creating_2020} and \texttt{HYTREL}~\cite{chen_hytrel_2023} model a table as a hypergraph.};
    \draw[] (n2003.east) --  (n20003.west);

    \node [cate, fill=blue!20, anchor=west] (n2003) at (7.5,8.2)  {PLM-based};
    \draw[] (n201.east) -- (7.15, 10.4)-- (7.15, 8.2)--  (n2003.west);
    \node [cate2, fill=blue!20, anchor=west] (n20003) at (10.7,8.2)  {\texttt{Starmie}~\cite{fan_semantics-aware_2023} leverages RoBERTa to encode columns; \texttt{AUTOTUS}~\cite{hu_automatic_2023} employs BERT to encode column pair relationships.};
    \draw[] (n2003.east) --  (n20003.west);

    \node [cate, fill=cyan!20, anchor=west] (n202) at (4.05,5.4)  {Schema\\Augmentation\\(Column-level)};
    \draw[] (n20.east) -- (3.8,3.2) -- (3.8,5.4) -- (n202.west);

    \node [cate, fill=cyan!20, anchor=west] (n2021) at (7.5,7)  {Value-based\\join};
    \draw[] (n202.east) -- (7.15, 5.4)-- (7.15,7)--  (n2021.west);
    \node [cate2, fill=cyan!20, anchor=west] (n20211) at (10.7,7)  {\texttt{LSH Ensemble}~\cite{zhu_lsh_2016} and \texttt{JOSIE}~\cite{zhu_josie_2019} formulate the join problem as an overlap set similarity search.};
    \draw[] (n2021.east) -- (n20211.west);

    \node [cate, fill=cyan!20, anchor=west] (n2022) at (7.5,5.4)  {Semantic-based\\joins};
    \draw[] (n202.east) -- (7.15, 5.4)-- (7.15, 5.4)--  (n2022.west);
    \node [cate2, fill=cyan!20, anchor=west] (n20212) at (10.7,5.4)  {\texttt{ARDA}~\cite{chepurko_arda_2020} joins on soft keys; \texttt{AutoFeature}~\cite{liu_feature_2022} and \texttt{FeatNavigator}~\cite{liang_featnavigator_2024} consider multi-hop semantic join; \texttt{PEXESO}~\cite{dong_efficient_2021} and \texttt{DeepJoin}~\cite{dong_deepjoin_2023} encode columns into high-dimensional vectors and join them based on similarity predicates; \texttt{OmniMatch}~\cite{koutras_omnimatch_2024} incorporates GNN for various similarity signals propagation.};
    \draw[] (n2022.east) --  (n20212.west);

    \node [cate, fill=cyan!20, anchor=west] (n2023) at (7.5,3.8)  {Structure-base\\joins};
    \draw[] (n202.east) -- (7.15, 5.4)-- (7.15, 3.8)--  (n2023.west);
    \node [cate2, fill=cyan!20, anchor=west] (n20213) at (10.7,3.8)  {\texttt{EmbDI}~\cite{cappuzzo_creating_2020} constructs a hypergraph capturing the table structure; \texttt{Leva}~\cite{zhao_leva_2022} constructs a database-level graph with unique table pool values as nodes.};
    \draw[] (n2023.east) --  (n20213.west);

    \node [cate, fill=green!20, anchor=west] (n203) at (4.05, 0.7)  {Cell Completion\\(Cell-level)};
    \draw[] (n20.east) -- (3.8,3.2) -- (3.8,0.7) -- (n203.west);

    \node [cate, fill=green!20, anchor=west] (n2031) at (7.5,2.25)  {Attribute name};
    \draw[] (n203.east) -- (7.15, 0.7)-- (7.15, 2.25)--  (n2031.west);
    \node [cate2, fill=green!20, anchor=west] (n20311) at (10.7,2.25)  {\texttt{Infogather}~\cite{yakout_infogather_2012} searches for similar tables and then match column labels; \texttt{EntiTables}~\cite{zhang_entitables_2017} ranks attribute labels based on KB and DB; \texttt{RATA}~\cite{glass_retrieval-based_2023} uses a retrieval-augmented self-trained transformer model to retrieve and select attribute names.};
    \draw[] (n2031.east) --  (n20311.west);

    \node [cate, fill=green!20, anchor=west] (n2032) at (7.5,0.7)  {Entity ID/name};
    \draw[] (n203.east) -- (7.15, 0.7)-- (7.15, 0.7)--  (n2032.west);
    \node [cate2, fill=green!20, anchor=west] (n20312) at (10.7,0.7)  {\texttt{EntiTables}~\cite{zhang_entitables_2017} takes a select-then-rank strategy; \texttt{RATA}~\cite{glass_retrieval-based_2023} uses a retrieval-augmented strategy.};
    \draw[] (n2032.east) --  (n20312.west);

    \node [cate, fill=green!20, anchor=west] (n2033) at (7.5,-0.9)  {Cell value};
    \draw[] (n203.east) -- (7.15, 0.7)-- (7.15,-0.9)--  (n2033.west);
    \node [cate2, fill=green!20, anchor=west] (n20313) at (10.7,-0.9)  {\texttt{Infogather}~\cite{yakout_infogather_2012}, \texttt{EntiTables}~\cite{zhang_entitables_2017} and \texttt{CellAutoComplete}~\cite{zhang_auto-completion_2019} retrieve related tables and then extract values; Ahmadov et al.~\cite{ahmadov_towards_2015} combine a retrieval-based method with a value prediction model; \texttt{TURL}~\cite{deng_turl_2022} and \texttt{RATA}~\cite{glass_retrieval-based_2023} leverage transformer.};
    \draw[] (n2033.east) --  (n20313.west);
    
    \node [cate, fill=pink!20, anchor=west] (n204) at (4.05, -3.4)  {Table Integration\\(Table-level)};
    \draw[] (n20.east) -- (3.8,3.2) -- (3.8,-3.4) -- (n204.west);

    \node [cate, fill=pink!20, anchor=west] (n2041) at (7.5,-2.7)  {Compositional};
    \draw[] (n204.east) -- (7.15, -3.4)-- (7.15, -2.7)--  (n2041.west);
    \node [cate2, fill=pink!20, anchor=west] (n20411) at (10.7,-2.7)  {\texttt{InfoGather}~\cite{yakout_infogather_2012} and \texttt{Entitables}~\cite{zhang_entitables_2017} can perform column and row augmentation simultaneously; \texttt{RATA}~\cite{glass_retrieval-based_2023} integrates three TDA tasks: row, column and cell population.};
    \draw[] (n2041.east) --  (n20411.west);

    \node [cate, fill=pink!20, anchor=west] (n2042) at (7.5,-4.1)  {Direct};
    \draw[] (n204.east) -- (7.15, -3.4)-- (7.15, -4.1)--  (n2042.west);
    \node [cate2, fill=pink!20, anchor=west] (n20412) at (10.7,-4.1)  {\texttt{ALITE}~\cite{khatiwada_integrating_2022} integrates two tables using Full Disjunction; \texttt{Leva}~\cite{zhao_leva_2022} constructs and embeds a table-pool-level graph to featurize the original table.};
    \draw[] (n2042.east) --  (n20412.west);

    \node [cate, fill=orange!20, anchor=west] (n21) at (0.7,-10.55){Generation-based};
    \draw[] (n1.south) -- (0.5,-4) -- (0.5,-10.55) -- (n21.west);

    \node [cate, fill=blue!20, anchor=west] (n210) at (4.05,-6.8)  {Record\\Generation\\(Row-level)};
    \draw[] (n21.east) -- (3.8,-10.55) -- (3.8,-6.8) -- (n210.west);
    
    \node [cate, fill=blue!20, anchor=west] (n2101) at (7.5,-5.9)  {Distribution\\-preserving};
    \draw[] (n210.east) -- (7.15, -6.8)-- (7.15, -5.9)--  (n2101.west);
    \node [cate2, fill=blue!20, anchor=west] (n21011) at (10.7,-5.9)  { Barak et al.~\cite{barak_privacy_2007} and Zhang et al.~\cite{zhang_privbayes_2014} use statistical approaches; \texttt{PATE-GAN}~\cite{jordon_pate-gan_2019}, \texttt{table-GAN}~\cite{park_data_2018}, \texttt{ITS-GAN}~\cite{chen_faketables_2019} and \texttt{GANBLR}~\cite{zhang_interpretable_2023} utilize GANs; \texttt{STaSy}~\cite{kim_stasy_2023} adopts a score-based generative model; \texttt{GOGGLE}~\cite{liu_goggle_2023} combines generative modeling with a graph that capturing the table structure; \texttt{CoDi}~\cite{lee_codi_2023} and \texttt{RelDDPM}~\cite{liu_controllable_2024} use diffusion models; \texttt{DP-LLMTGen}~\cite{tran_differentially_2024} utilizes LLMs.};
    \draw[] (n2101.east) --  (n21011.west);

    \node [cate, fill=blue!20, anchor=west] (n2102) at (7.5,-7.7)  {Class-imbalance\\-aware};
    \draw[] (n210.east) -- (7.15, -6.8)-- (7.15, -7.7)--  (n2102.west);
    \node [cate2, fill=blue!20, anchor=west] (n21021) at (10.7,-7.7)  {\texttt{CTGAN}~\cite{xu_modeling_2019} uses a contitional GAN; \texttt{cWGAN}~\cite{engelmann_conditional_2021} leverages Wasserstein GAN; \texttt{SIGRNN}~\cite{al-bahrani_sigrnn_2021} employs RNN; \texttt{SOS}~\cite{kim_sos_2022} utilizes a score-based generative model.};
    \draw[] (n2102.east) --  (n21021.west);

    \node [cate, fill=cyan!20, anchor=west] (n211) at (4.05,-9.65)  {Feature\\Construction\\(Column-level)};
    \draw[] (n21.east) -- (3.8,-10.55) -- (3.8,-9.65) -- (n211.west);
    
    \node [cate, fill=cyan!20, anchor=west] (n2111) at (7.5,-9.0)  {Explicit};
    \draw[] (n211.east) -- (7.15, -9.65)-- (7.15, -9.0)--  (n2111.west);
    \node [cate2, fill=cyan!20, anchor=west] (n21111) at (10.7,-9.0)  {Kanter and Veeramachaneni~\cite{kanter_deep_2015} use mathematical transformations; \texttt{ExploreKit}~\cite{katz_explorekit_2016} employs predefined operators;  \texttt{SMARTFEAT}~\cite{lin_smartfeat_2024} leverages FMs.};
    \draw[] (n2111.east) --  (n21111.west);

    \node [cate, fill=cyan!20, anchor=west] (n2112) at (7.5,-10.35)  {Implicit};
    \draw[] (n211.east) -- (7.15, -9.65)-- (7.15, -10.35)--  (n2112.west);
    \node [cate2, fill=cyan!20, anchor=west] (n21121) at (10.7,-10.35)  {\texttt{GAINS}~\cite{xiao_beyond_2023} encodes and optimizes feature vectors; Wang et al.~\cite{wang_reinforcement-enhanced_2023} encode and optimize feature transformation operation sequences.};
    \draw[] (n2112.east) --  (n21121.west);

     \node [cate, fill=green!20, anchor=west] (n212) at (4.05,-12.1)  {Cell Imputation\\(Cell-level)};
    \draw[] (n21.east) -- (3.8,-10.55) -- (3.8,-12.1) -- (n212.west);
    
    \node [cate, fill=green!20, anchor=west] (n2121) at (7.5,-11.5)  {Statistical};
    \draw[] (n212.east) -- (7.15, -12.1)-- (7.15, -11.5)--  (n2121.west);
    \node [cate2, fill=green!20, anchor=west] (n21211) at (10.7,-11.5)  {\texttt{MICE}~\cite{buuren_mice_2011} creates a statistical model for each column; \texttt{MissForest}~\cite{stekhoven_missforestnon-parametric_2012} trains a random forest on the observed parts.};
    \draw[] (n2121.east) --  (n21211.west);

    \node [cate, fill=green!20, anchor=west] (n2122) at (7.5,-12.7)  {Deep-learning};
    \draw[] (n212.east) -- (7.15, -12.1)-- (7.15, -12.7)--  (n2122.west);
    \node [cate2, fill=green!20, anchor=west] (n21221) at (10.7,-12.7)  {\texttt{MIWAE}~\cite{mattei_miwae_2019} and \texttt{HI-VAE}~\cite{nazabal_handling_2020} use VAE; \texttt{GAIN}~\cite{yoon_gain_2018} and \texttt{MIGAN}~\cite{dai_multiple_2021} employ GAN; \texttt{TabCSDI}~\cite{zheng_diffusion_2023} leverages the diffusion model.};
    \draw[] (n2122.east) --  (n21221.west);

    \node [cate, fill=pink!20, anchor=west] (n213) at (4.05,-14.6)  {Table Synthesis\\(Table-level)};
    \draw[] (n21.east) -- (3.8,-10.55) -- (3.8,-14.6) -- (n213.west);
    
    \node [cate, fill=pink!20, anchor=west] (n2131) at (7.5,-13.9)  {Compositional};
    \draw[] (n213.east) -- (7.15, -14.6)-- (7.15, -13.9)--  (n2131.west);
    \node [cate2, fill=pink!20, anchor=west] (n21311) at (10.7,-13.9)  {Compositional approach combines previous works that generate rows or columns separately.};
    \draw[] (n2131.east) --  (n21311.west);

    \node [cate, fill=pink!20, anchor=west] (n2132) at (7.5,-15.3)  {Direct};
    \draw[] (n213.east) -- (7.15, -14.6)-- (7.15, -15.3)--  (n2132.west);
    \node [cate2, fill=pink!20, anchor=west] (n21321) at (10.7,-15.3)  {\texttt{TransTab}~\cite{wang_transtab_2022} encodes and retains knowledge from the table pool and then synthesizes a new encoded table; leverage LLMs to perform table synthesis, such as prompt engineering. };
    \draw[] (n2132.east) --  (n21321.west);

    \end{tikzpicture}}
    \caption{The categorization of the TDA approaches, from both task-oriented and table-level perspectives. We also provide a concise introduction to the key TDA techniques within each category.}
    \label{fig5:aug_main_table}
    \Description[Overview of Augmentation Approaches]{Categorization of Augmentation work}
\end{figure}
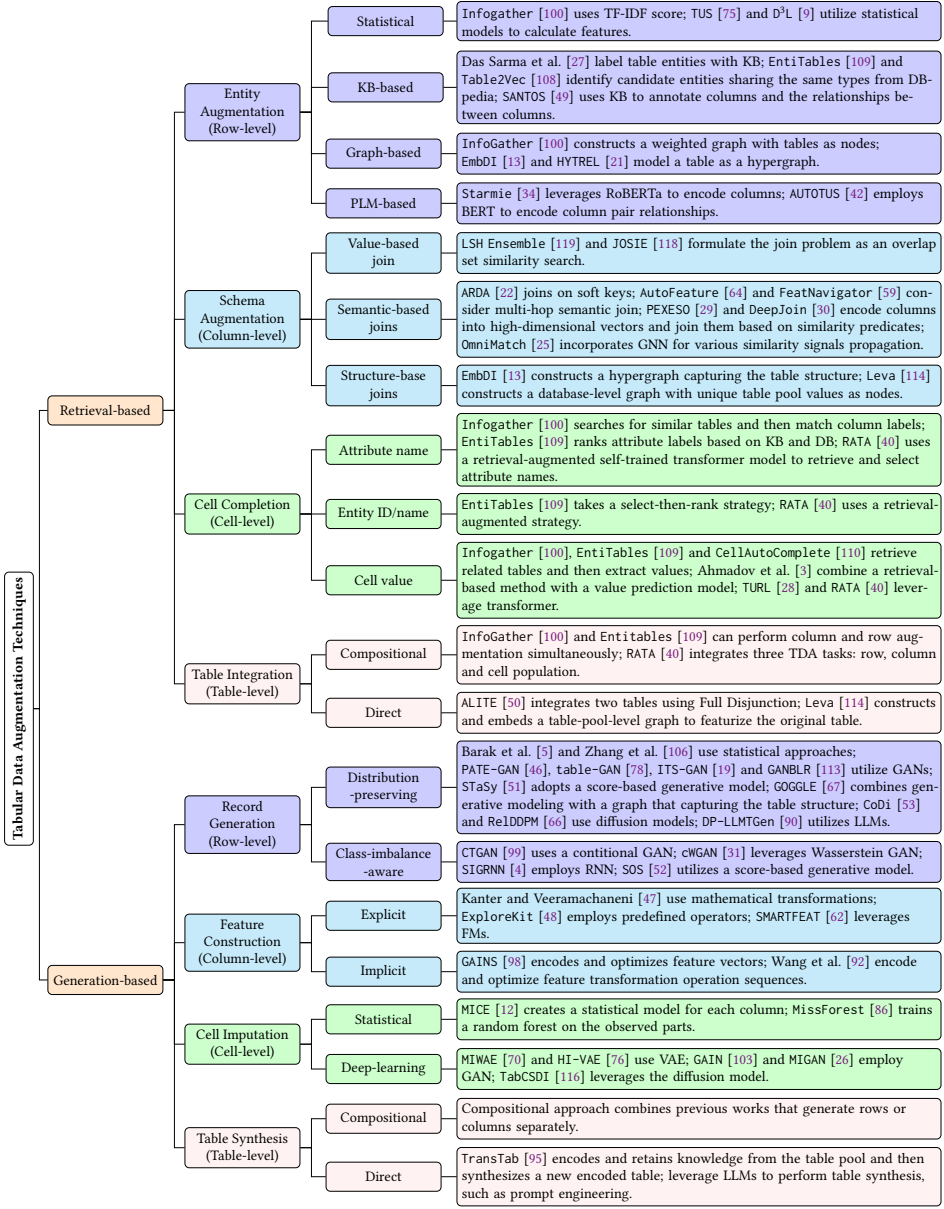

\subsection{Retrieval-based TDA}
\label{section:retrieval_based_TDA}

By retrieval-based, we refer to the process of enhancing the original table (query table $T^O$) with realistic data sourced from table pools $\mathbb{T} = \{T_i\}$.
One important difference between augmenting tabular data and augmenting other data modalities lies in the availability of existing data resources (such as databases and data warehouses), which provide opportunities for discovering fresh, realistic data.
In contrast, other data modalities (such as images) primarily rely on transforming the original data to generate new data that has not been seen before~\cite{chai_data_2022}.
Retrieval-based TDA tasks are further divided into Entity Augmentation (\REA{}) at the \texttt{row} level, Scheme Augmentation (\RSA{}) at the \texttt{column} level, Cell Completion (\RCC{}) at the \texttt{cell} level, and Table Integration (\RTI{}) at the \texttt{table} level. 
They will be introduced as follows.

\subsubsection{Entity Augmentation (\REA{})}

Previous works~\cite{yakout_infogather_2012,zhang_web_2020} define retrieval-based TDA at the row-level as entity augmentation, as rows in tabular data generally correspond to specific entities. 
Entity augmentation extends a given table with more rows or row elements retrieved from table pools. 
Directly retrieving table rows without considering the context of the entire table is not feasible. Therefore, existing solutions typically first search for tables that are \emph{unionable} with the input table and then select entities from these unionable tables to augment the original table. 
As we shall see, table search~\cite{das_sarma_finding_2012,castro_fernandez_aurum_2018,trabelsi_strubert_2022} is inherently involved in this process, serving as an intermediate step that feeds into tabular data augmentation~\cite{zhang_web_2020}. 

To measure the unionability of two tables, existing solutions typically start by using table representation techniques (see Section~\ref{section:preaug-table-representation}) to convert tabular data into latent-space vectors. 
These vectors, at the higher level, are then used to compute the relatedness between the source and target tables, with the resulting scores used to rank the target tables.
For instance, Sarma et al.~\cite{das_sarma_finding_2012} search for the top-$k$ related tables that are entity complements to the input table and then use these related tables to populate the input table. 
Most related works follow a similar approach as~\cite{das_sarma_finding_2012}, with distinctions mainly in the methodology employed to identify the top-$k$ unionable tables. These methodologies can be further categorized into four domains: statistical, KB-based, graph-based, and PLM-based.

\textbf{Statistical entity augmentation methods} use statistical models to estimate the unionability between two tables. 
For example, \texttt{Infogather}~\cite{yakout_infogather_2012} measures the context-to-context and table-to-context similarity by calculating the cosine similarity of their TF-IDF vectors.
\texttt{TUS}~\cite{nargesian_table_2018} utilizes three statistical models to statistically test the value overlap, semantic overlap, and embedding overlap between attributes (columns), and then aggregate the results to derive the unionability between tables. 
\texttt{D$^3$L}~\cite{bogatu_dataset_2020} extends \texttt{TUS} by incorporating additional statistical measures for numerical value distribution, which is based on Kolmogorov-Smirnov (KS) statistic.

\textbf{KB-based entity augmentation methods} consider a knowledge base (KB) for identifying potential unionable tables, instead of solely relying on tables within a table corpus or a table pool. 
For example, Das Sarma et al.~\cite{das_sarma_finding_2012} represent table entities as weighted label sets from a knowledge base (WebIsA~\cite{limaye_annotating_2010} or Freebase\footnote{\url{http://www.freebase.com}}) or from a table corpus, and take their dot product to compute the unionability between the input and candidate tables. 
\texttt{EntiTables}~\cite{zhang_entitables_2017} incorporates the DBpedia\footnote{\url{https://www.dbpedia.org}} knowledge base to identify candidate entities. In their approach, they not only collect entities from similar tables, but also collect entities sharing the same types or categories from DBpedia with the input entities. They have proved that using related tables and using a KB are complementary when searching for candidate entities. 
Subsequently, \texttt{Table2Vec}~\cite{zhang_table2vec_2019} further improves \texttt{EntiTables} for entity augmentation by incorporating Word2Vec to train table embeddings for entities.
More recently, \texttt{SANTOS}~\cite{khatiwada_santos_2023} leverages external knowledge bases, YAGO 4\footnote{\url{http://yago-knowledge.org}}, to annotate both columns and the binary relationships between columns. To address the limited coverage of KB, \texttt{SANTOS} proposes a self-curated knowledge base on top of the table pool.
To be more specific, \texttt{SANTOS} hypothesizes that columns with common semantics have overlap values. Thus, it first assigns a unique synthesized label to each column in the table pool with a confidence score 1; if two columns have overlap values, their synthesized labels can be assigned to each other with a confidence score based on the relatedness that are computed using the two columns' values.

\textbf{Graph-based entity augmentation methods} convert tables to a graph to compare the unionability between tables. 
An illustrative example is \texttt{InfoGather}~\cite{yakout_infogather_2012}, which computes topic-sensitive pagerank (TSP) over a weighted graph, where nodes represent tables and edges indicate direct pairwise match between tables. 
\texttt{EmbDI}~\cite{cappuzzo_creating_2020} determines entity similarity from a table representation using compact tripartite graphs. Initially, \texttt{EmbDI} forms a heterogeneous graph based on the table, where cells, entities, and columns act as nodes, and the relationships derived from the table schema serve as edges. \texttt{EmbDI} then uses random walks to traverse the graph to construct sentences that can describe the table. The sentences are fed into word embedding algorithms like word2vec for entity representation and unionability computation.
More recently, \texttt{HYTREL}~\cite{chen_hytrel_2023} models a table as a hypergraph and then encodes the hypergraph for table similarity prediction. In the hypergraph, nodes indicate table cells and three different types of hyperedges represent row, column, and the entire table, respectively.

\textbf{PLM-based entity augmentation methods} have been adapted for tabular data more recently, where TDA is one of the cases.
PLMs often serve as table encoders, enhancing the capture of table semantics and structure for subsequent similarity calculations.
Recently, \texttt{Starmie}~\cite{fan_semantics-aware_2023} leverages PLMs, i.e., RoBERTa, to encode columns.
It then derives a unionability score between two tables using column aggregation algorithms.
Recently, Hu et al.~\cite{hu_automatic_2023} employ PLMs (BERT and RoBERTa) to obtain contextualized representation of column pair relationships, capturing nuanced table contexts to identify unionable tables in the pools.

\remarkbox{
Entity augmentation can expand and diversify the samples for ML, making it a long-standing research focus.
Statistical entity augmentation methods are typically based on fixed distribution assumptions and hard to refine. Meanwhile, it may encounter issues of computational efficiency and scalability when dealing with large-scale datasets.
KB-based entity augmentation, on the other hand, suffers from limited KB coverage, often leading to low recall in practical scenarios. However, with the development of LLMs, the incorporation of RAG may be a future direction. RAG combines the dynamically retrieved information with the ability of generative models, reducing information loss and improving information relevance, thereby increasing recall rates.
Graph-based entity augmentation, which converts tables to graphs, can better grasp the structure relationship of tables, but it also requires an additional encoding mechanism for the constructed graph (e.g., \texttt{HYTREL}~\cite{chen_hytrel_2023} uses a transformer module to encode the hypergraphs). The entire process would be resource-intensive and time-consuming.
For PLM-based entity augmentation, most methods directly sequentialize tabular data ignoring the table structure. 
These limitations highlight the need for more advanced, structure-aware methods that can effectively leverage the strengths of language models while addressing the unique challenges posed by tabular data.
Furthermore, a significant portion of the existing research has focused primarily on table union search, without addressing the potential differences between the retrieved tables and the original data. 
Even when unionable tables are found, discrepancies may still need resolution before the tables can be combined.
}

\subsubsection{Schema Augmentation (\RSA{})}
We refer to retrieval-based TDA at the column level as schema augmentation, the process of extending the schema of the original table with additional columns from table pools. 
This process is somewhat similar to the \emph{join operation} commonly executed in databases. 
To be specific, given a table pool $\mathbb{T}$ containing $m$ tables $\{T_i\}^m_{i=1}$, the task of schema augmentation \RSA{} is to search $\mathbb{T}$ and find the columns $T_i[:, j]$ joinable to the query table $T^O$. 
As a result, the result table $T^R =$ \RSA{} $(T^O, \texttt{column}, \mathbb{T})$ can boost the downstream ML task.
To determine the joinability between two columns, the column values, column semantics, and table structure are the most considered three perspectives.
The corresponding subcategories are introduced below.

\begin{figure}[!htbp]
  \centering
  \includegraphics[width=\linewidth]{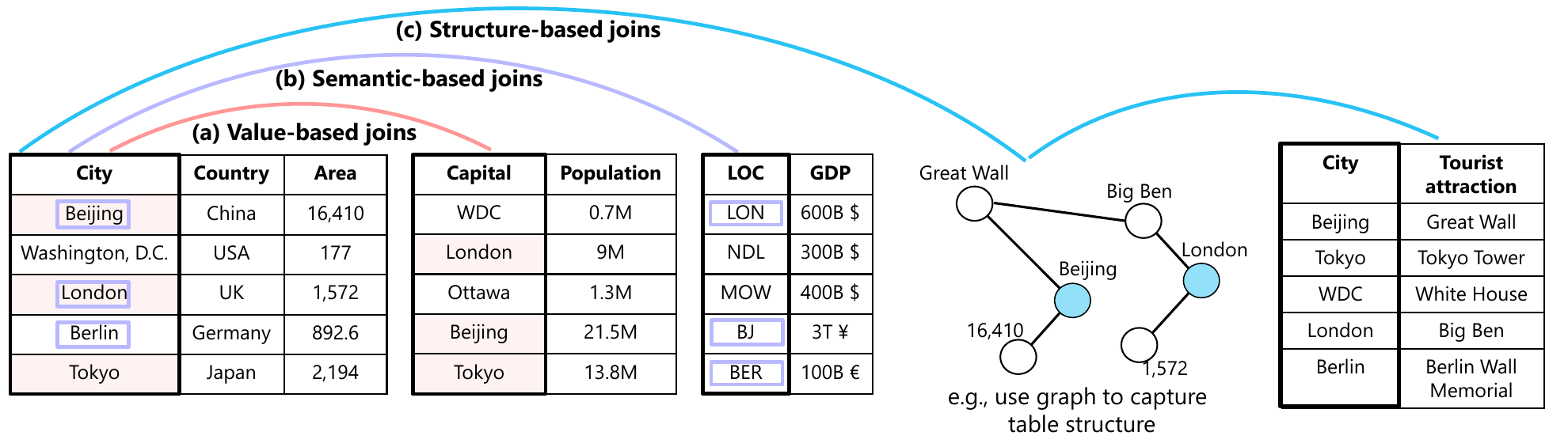}
  \caption{The illustration of schema augmentation, including (a) value-based joins, (b) semantic-based joins, and (c) structure-based joins.}
  \label{f8_aug_schema_augmentation}
  \Description{}
\end{figure}

\textbf{Value-based joins}. 
Early approaches~\cite{zhu_lsh_2016,zhu_josie_2019} for schema augmentation primarily focus on value-based join, where only exactly matching value can be joined~\cite{dong_deepjoin_2023} (see Fig.~\ref{f8_aug_schema_augmentation} (a)). 
For example, \texttt{LSH Ensemble}~\cite{zhu_lsh_2016} formulates the join problem as an overlap set similarity search by treating columns as sets and matching values as intersections between sets.
\texttt{JOSIE}~\cite{zhu_josie_2019} further improves \texttt{LSH Ensemble} by using intersection estimation to reduce the cost of set reads and the top-$k$ set search. \texttt{JOSIE} starts by calculating the cost of reading posting lists and sets using statistical approximation techniques. It then uses an adaptive algorithm that switches between reading posting lists to gather candidates and reading sets to compute exact intersection sizes.

\textbf{Semantic-based joins}. 
Recent studies~\cite{chepurko_arda_2020,dong_efficient_2021,dong_deepjoin_2023,koutras_omnimatch_2024} have explored the concept of semantic joins, which join those columns with similar meanings (see Fig.~\ref{f8_aug_schema_augmentation} (b)) instead of those ones having identical values. 
This approach is capable of handling misspellings and formatting variations, resulting in larger set of join results.
For instance, \texttt{ARDA}~\cite{chepurko_arda_2020} performs joins on soft keys without requiring exact match. However, \texttt{ARDA} only consider one-hop semantic join (e.g., customer $\Join$ order, customer $\Join$ product), while \texttt{AutoFeature}~\cite{liu_feature_2022} and \texttt{FeatNavigator}~\cite{liang_featnavigator_2024} consider multi-hop semantic join (e.g., customer $\Join$ order $\Join$ product).
\texttt{PEXESO}~\cite{dong_efficient_2021} targets the case that columns are embedded into high-dimensional vectors and they are joined based on similarity predicates. 
\texttt{DeepJoin}~\cite{dong_deepjoin_2023} enhances the capabilities of \texttt{PEXESO}~\cite{dong_efficient_2021} by employing PLMs (e.g., BERT) as the column vector encoder. Meanwhile, \texttt{DeepJoin} can perform both value- and semantic-based joins for textual columns, breaking the drawback of previous methods~\cite{zhu_lsh_2016,zhu_josie_2019,dong_efficient_2021} that could only handle one type of join and surpassing these works in performance. However, \texttt{DeepJoin} can only handle textual columns with a relatively small cardinality.
\texttt{OmniMatch}~\cite{koutras_omnimatch_2024} also detects both value- and semantic-based join between columns. \texttt{OmniMatch} incorporates Graph Neural Networks for similarity signals propagation to better capture column-pair similarity. Meanwhile, \texttt{OmniMatch} adopts a self-supervised learning approach by generating positive and negative join examples from the table pool, eliminating the need for large amounts of labeled data.

\textbf{Structure-based joins}. Most recent works tend to incorporate the table structure information in the column embeddings, such as using graphs to capture the table structure as illustrated in Fig.~\ref{f8_aug_schema_augmentation} (c).
For example, \texttt{EmbDI}~\cite{cappuzzo_creating_2020} first constructs a hypergraph capturing the table structure. The hypergraph contains one type of nodes representing columns. \texttt{EmbDI} then derives column embeddings from the column nodes containing the table structure information. 
\texttt{Leva}~\cite{zhao_leva_2022} also uses a graph to capture the table structure, but the graph contains the information from the entire database with unique values in the table pool as nodes. 
Similarly, Bharadwaj et al.~\cite{bharadwaj_discovering_2021} also construct a database-level graph, with columns as nodes and the relationships between columns as edges.

\remarkbox{
Schema augmentation extends the original table with more features, thereby improving the downstream ML tasks. 
This area has evolved from traditional value-based joins, which depend on exact value matches, to more sophisticated semantic- and structure-based joins.
Semantic-based joins have seen a trend of leveraging generative AI (e.g., PLMs) as encoders to capture the inherent semantics and contextual relationships within and across table columns. 
However, due to token length limitations, these methods are often restricted to handling table columns with relatively small cardinalities. 
When dealing with columns that have high cardinalities, additional sampling or chunking steps are required, which can lead to potential information loss and reduced join performance.
In structure-based joins, most methods utilize graphs to capture the table structure. It has been observed that more complex graph structures can better represent the nuanced aspects of the table schema and relationships.
However, this increased complexity often requires more resources. Therefore, it is important to explore a balanced approach that finds a careful equilibrium between the level of structural detail captured and the computational resources needed. 
}

\begin{figure}[!htbp]
  \centering
  \includegraphics[width=\linewidth]{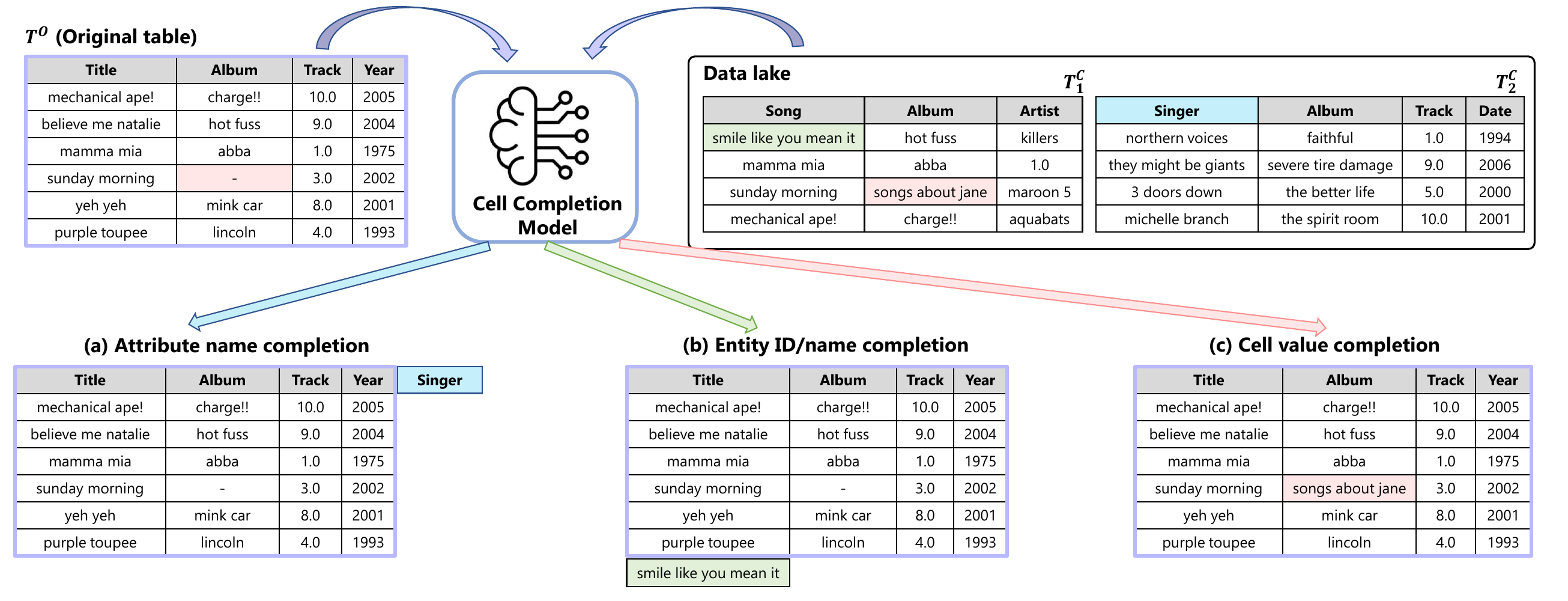}
  \caption{The illustration of three cell completion tasks: (a) attribute name completion, (b) entity ID/name completion, and (c) cell value completion.}
  \label{f9_aug_cell_completion}
  \Description{}
\end{figure}

\subsubsection{Cell Completion (\RCC{})}

As with previous work~\cite{zhang_auto-completion_2019}, we refer to retrieval-based TDA at the cell level as cell completion.
This task involves filling in empty cells within the input table by leveraging information extracted from table pools. 
Cell completion can be divided into several subtasks based on cell types: populating attribute names, adding additional entity IDs/names, and filling values for table cells~\cite{zhang_auto-completion_2019}. 
The approaches per category are introduced as follows.

\textbf{Attribute name completion}. 
This task is to populate the input table with additional possible column headers or labels. As shown in Fig.~\ref{f9_aug_cell_completion} (a), the original song-related table $T^O$ is augmented with a new column header ``Singer'' retrieved from a related table $T^C_2$ in the table pool, involving new features for ML.
As a typical study, Yakout et al.~\cite{yakout_infogather_2012} search for similar tables and then match column labels. Their proposed approach for matching column labels utilize additional information, including similarities that are based on context, attribute names, and column values, respectively. 
\texttt{EntiTables}~\cite{zhang_entitables_2017} ranks a list of labels to be added as headings of new columns by using probabilistic models. The ranking is based on information obtained from a knowledge base and similar tables retrieved from a table corpus. 
A recent approach, \texttt{RATA}~\cite{glass_retrieval-based_2023}, addresses this problem using a retrieval-augmented self-trained transformer model. Specifically, \texttt{RATA} first indexes and searches tables from the table pool using a bi-encoder retrieval model, and then identifies augmentations from retrieved tables using a reader transformer.

\textbf{Entity ID/name completion}. 
Entity ID/name completion refers to populate the key column (i.e., the entity's ID or name) for a row (an entity). For example in Fig.~\ref{f9_aug_cell_completion} (b), this procedure augments $T^O$ with a new entity name ``smile like you mean it'' (a song) retrieved from the table pool table $T^C_1$, thereby enriching the potential samples.
Zhang and Balog~\cite{zhang_entitables_2017} populate rows with additional entities using a two-step approach: (1) candidate entity selection: search for table pool tables that contain the same or similar entities in the original table or search for KB entities with similar KB labels as the original table entities; and (2) entity ranking: leverage a probabilistic formulation based on Bayes' theorem to calculate the relatedness between entities. 
\texttt{RATA}~\cite{glass_retrieval-based_2023} tackles the entity ID/name completion in a similar manner to its attribute name population, using a retrieval-augmented strategy. This strategy pretrains the retrieval-based model by randomly removing entities from the corpus and then reconstructing the removed entities, aiming to better capture semantics and structure of tables.

\textbf{Cell value completion}. 
In this task of finding values for empty data cells, models are designed to estimate a specific value to fill in based on the information from the table pool. Referring to Fig.~\ref{f9_aug_cell_completion} (c), this procedure fills in the empty cell in $T^O$ with a new value ``songs about jane'' retrieved from the table pool table $T^C_1$ to avoid null values in learning tasks.
A common approach is to retrieve tables from the table corpus and then extract values from those tables~\cite{yakout_infogather_2012,zhang_entitables_2017}. 
Zhang and Balog~\cite{zhang_entitables_2017,zhang_auto-completion_2019} build upon this approach such that they consider integrating supplementary information from a knowledge base. 
Ahmadov et al.~\cite{ahmadov_towards_2015} further improve these approaches~\cite{yakout_infogather_2012,zhang_entitables_2017} by introducing a hybrid method that combines a retrieval-based method with a
generation-based method (a value prediction model). The prediction model is implemented using a black box approach that can automatically choose the best ML model (e.g., $k$NN) and the corresponding parameters.
More recently, Deng et al.~\cite{deng_turl_2022} propose the pre-trained \texttt{TURL} model, with cell value completion as one of its downstream tasks. 
Specifically, they pretrain a structure-aware transformer encoder with table pool tables to model the row-column structure by using a new Masked Entity Recovery (MER) objective. 
Likewise, \texttt{RATA}~\cite{glass_retrieval-based_2023} addresses the cell value completion problem with a transformer model. 
However, it incorporates a reader-or-selection component, that reads the retrieved table and selects the most related ones. This component is based on an extractive approach, ensuring that the model's predictions are always based on existing data rather than speculative assumptions.

\remarkbox{
Missing values are common and typically have a negative impact in practice, making cell completion crucial. 
This field has been actively studied for decades, and multiple works~\cite{zhang_entitables_2017,glass_retrieval-based_2023} consider addressing various cell types simultaneously. 
For attribute name and entity ID/name completion, an internal problem is that additional cell completion is required after these two operations to obtain a complete table, which may lead to problem propagation. 
Another issue is that retrieval-based methods may sometimes be outperformed by simple statistical techniques like averaging, especially in domain-specific tasks.
Consequently, given the successful application of RAG in the NLP field, there is an opportunity to combine retrieval-based and generation-based methods~\cite{wang_transtab_2022,tran_differentially_2024}, as generative approaches can better capture the inherent structure and patterns within tables. 
Future advancements could involve integrating generative AI models, such as language models, into this hybrid approach. 
}

\subsubsection{Table Integration (\RTI{})}

We define retrieval-based TDA at the table level as table integration, the procedure of extending the original table with both rows and columns from related tables retrieved from the table pool via table search algorithms. 
There are two main methods for implementing table integration: one is compositional table integration, which combines retrieval-based TDA results at various levels (rows, columns, and/or cells), while the other is direct table integration, which directly enriches the original table with the content from the retrieved related tables.

\textbf{Compositional table integration}. 
An early work \texttt{InfoGather}~\cite{yakout_infogather_2012} presents a holistic augmentation framework that can perform column and/or row augmentation simultaneously. 
To be more specific, they introduce three core operations: entity augmentation by attribute name, entity augmentation by example, and attribute discovery. 
\texttt{Entitables}~\cite{zhang_entitables_2017} targets two tasks: augmenting rows with additional entities and augmenting columns with new headers. 
More recently, \texttt{RATA}~\cite{glass_retrieval-based_2023} integrates three TDA tasks: row population, column population, and cell completion. \texttt{RATA} functions as an end-to-end model that initially retrieves related tables and then extracts various table elements from these tables to perform different levels of TDA tasks. 

\textbf{Direct table integration}. 
More recently, the concept of direct table integration was proposed by~\cite{miller_open_2018}, which aims to find the right operators (e.g., join, nest, group, link, and twist) to integrate tabular data into a desired form. 
A recent work~\cite{khatiwada_integrating_2022} has implemented this approach, proposing a model that extends the original table with both rows and columns simultaneously. They integrate two tables using Full Disjunction, which first connects two tables through an outer-join and then eliminates redundant rows. 
\texttt{Leva}~\cite{zhao_leva_2022} takes a different approach by integrating tables in the latent space. 
To be specific, they construct a data-lake-level graph, and the embedding of this graph is used to featurize the original table.

\remarkbox{
Table integration is a relatively new concept compared to other levels of retrieval-based TDA tasks, thus it still has a vast range of unexplored opportunities for further research and development. 
For compositional approaches that concatenate combining different levels of TDA, there is a risk of error propagation. 
The compounding of potential errors or biases from the individual TDA tasks can diminish the overall effectiveness and reliability of the integrated table.
For direct approaches, there are relatively few works so far, possibly due to the lack of integration benchmarks. 
Most existing works rely on strong assumptions and are only applicable to small table pools. 
Researchers and practitioners may need to develop more robust and scalable integration techniques, as well as establish comprehensive benchmarking platforms.
}

\subsection{Generation-based TDA}
\label{section:generation_based_TDA}

Generation-based TDA refers to the augmentation of tabular data through the generation of synthetic data. Unlike retrieval-based methods, generation-based methods do not require external data sources and are often built upon generative models. Generation-based TDA tasks can be further categorized into the following sub-tasks: Record Generation (\GRG{}) at the \texttt{row} level, Feature Construction (\GFC{}) at the \texttt{column} level, Cell Imputation (\GCI{}) at the \texttt{cell} level, and Table Synthesis (\GTS{}) at the \texttt{table} level. These generation-based TDA tasks will be introduced and discussed in more detail in the following sections.

\subsubsection{Record Generation (\GRG{})}

Record generation aims at generating additional records (table rows) based on the original table and its associated information. 
In many practical scenarios, publicly released tables often contain only a small subset of the total available records due to various concerns such as legal constraints or privacy issues. 
They are referred to as \emph{sub-tables} in this context.
As a result, ML models trained on these limited sub-tables may suffer from suboptimal performance. 
Therefore, it is crucial to generate more synthetic records from the released sub-table. 
Based on the distribution statuses of the original and result tables, the goals of record generation approaches differ: one approach focuses on preserving the original distribution, while the other aims to address imbalanced tabular data through oversampling.

\textbf{Distribution-preserving record generation} ensures that the generated records maintain the same distribution as the original table. 
For example, early works use statistical approaches (e.g., Bayesian networks~\cite{zhang_privbayes_2014} and Fourier decomposition~\cite{barak_privacy_2007}) to model the distribution of the original table and then generate synthetic records by sampling from the distribution.
Recently, \texttt{table-GAN}~\cite{park_data_2018} leverages generative adversarial networks (GAN) to generate synthetic records that are statistically similar to distribution of the original table. Specifically, \texttt{table-GAN}'s framework comprises three neural networks: a generator, a discriminator, and a classifier, which collectively enhance the semantic coherence of the synthetic records.
Similar to \texttt{table-GAN}, several other works such as \texttt{PATE-GAN}~\cite{jordon_pate-gan_2019}, \texttt{ITS-GAN}~\cite{chen_faketables_2019} and \texttt{GANBLR}~\cite{zhang_interpretable_2023} also employ GAN to generate distribution-preserving records. \texttt{PATE-GAN} introduces differential privacy guarantees for privacy concern; \texttt{ITS-GAN}~\cite{chen_faketables_2019} further maintains functional dependencies to capture the relationships between attributes; while \texttt{GANBLR} addresses the interpretation limitation of previous GAN-based methods and further consider explicit feature interactions.
Diffsion models have also been applied for record generation.
\texttt{STaSy}~\cite{kim_stasy_2023} adopts a score-based generative model that uses the reverse diffusion process to generate records based on the score function aiming at enhancing sampling quality and diversity.
\texttt{CoDi}~\cite{lee_codi_2023} leverages two diffusion models to process continuous and discrete columns separately and then co-evolve the two diffusion models by transforming conditions from and to each other during training.
\texttt{RelDDPM}~\cite{liu_controllable_2024} use diffusion models to generate tuples that not only cater to the original distribution but also meet specific conditions, such as satisfying a particular criterion for a given attribute. In particular, \texttt{RelDDPM} first trains an unconditional generative model (diffuser module) to capture the overall distribution of the original tabular data. Then, it uses controllers to measure how well the synthetic data matches the condition and guide the diffuser module to generate data that better satisfies the condition.
\texttt{GOGGLE}~\cite{liu_goggle_2023} further considers the relational structure of the original table when generating records. 
It first learns an approximate relational structure through the construction of a graph, which serves as the foundation of the generative modeling. Subsequently, it uses a variational autoencoders (VAE) architecture to gradually generate synthetic records similar to the original table through message passing over the constructed graph.
Most recently, \texttt{DP-LLMTGen}~\cite{tran_differentially_2024} explores LLMs for record generation. The LLM undergoes a two-stage fine-tuning process, one for capturing table formats, and the other for learning the feature distributions and dependencies.

\textbf{Class-imbalance-aware record generation} particularly focuses on oversampling imbalanced samples/instances (rows) within the tabular data, since class imbalance can significantly hinder the predictive performance of classification models.
For example, \texttt{CTGAN}~\cite{xu_modeling_2019} deals with the imbalanced tabular data by using a conditional generator that can condition on one of the discrete attributes.
It also uses a training-by-sampling strategy, which ensures even sampling from the discrete attributes.
Similarly, \texttt{cWGAN}~\cite{engelmann_conditional_2021} first estimates the underlying distribution of the original table using Wasserstein GAN, a derivate of GAN optimizing GAN's training process, and then uses the trained generator to synthesize additional samples of the minority class.
\texttt{SIGRNN}~\cite{al-bahrani_sigrnn_2021} employs a sequence-to-sequence recurrent neural network (RNN) to generate minority class instances. The RNN is trained on the minority class instances to learn their data distribution, and then used to generate synthetic instances to augment the original dataset and balance the minority class.
\texttt{SOS}~\cite{kim_sos_2022} proposes a transfer-based oversampling technique. \texttt{SOS} leverages a score-based generative model to transform majority class records to those ``fake'' minority class records.

\remarkbox{
Record generation has a broad range of applications in ML, such as increasing sample size, oversampling imbalanced classes, and preserving privacy. This field has evolved from using statistical models to incorporating deep neural network models.
Despite the progress made, a key challenge remains: how to effectively capture the complex attribute correlations present in the original tabular data. The synthetic records generated should not only reflect the overall statistical distribution of the source table but also accurately preserve the intricate relationships between different columns or features. 
Since the database community has extensively studied attribute correlations, integrating database techniques with newly proposed neural network models could be a promising direction. 
Additionally, tabular data often comes with a limited number of samples, making it challenging to train neural network models effectively and leading to a risk of overfitting.
Techniques such as employing regularization methods (e.g., \texttt{GOGGLE}~\cite{liu_goggle_2023}) and adopting self-supervised learning should be further explored to address this issue.
}

\subsubsection{Feature Construction (\GFC{})}

Feature construction refers to the task of generating additional columns, also known as features, to enhance the original table. 
Adequate features are essential for training ML models to achieve optimal performance, though good features alone are not always sufficient~\cite{liu_feature_2022}. 
Consequently, numerous studies have focused on feature construction. Based on whether features are transformed into vectors and manipulated in hidden layers, feature construction methods can be categorized into two main groups: explicit and implicit.

\begin{figure}[!htbp]
  \centering
  \includegraphics[width=\linewidth]{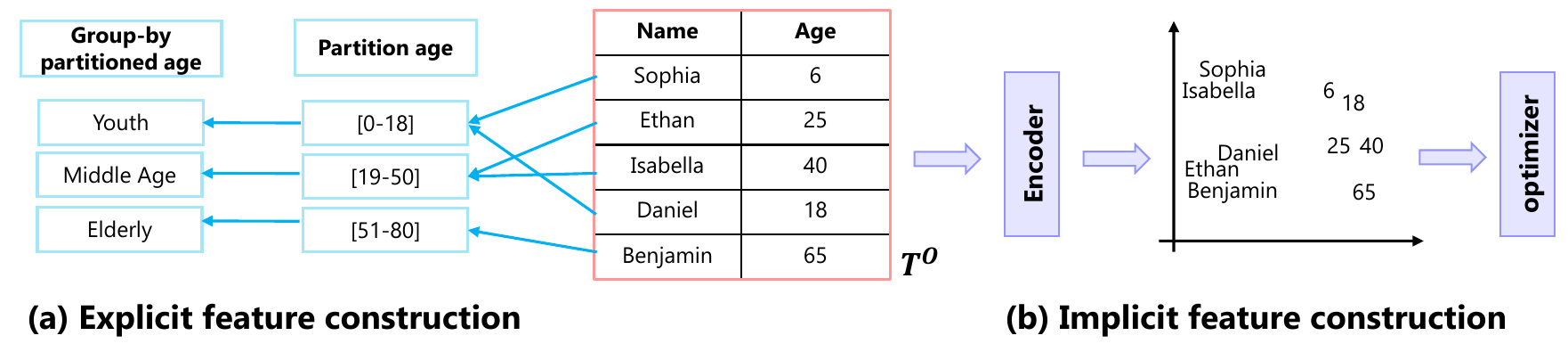}
  \caption{The illustration of feature construction: (a) explicit feature construction and (b) implicit feature construction.}
  \label{f10_aug_feature_construction}
  \Description{}
\end{figure}

\textbf{Explicit feature construction} generates new features by directly manipulating existing ones. For example, it can partition continuous columns into segments, as shown in Fig.~\ref{f10_aug_feature_construction} (a), where the ``Age'' column is divided into [0, 18], [19, 50], and [51, 80].
For instance, Kanter and Veeramachaneni~\cite{kanter_deep_2015} use existing raw features and apply a variety of mathematical transformations (e.g., AVG, MIN, and MAX) to create a hierarchy of synthesized features that express more complex relationships in the data. 
Similarly, \texttt{ExploreKit}~\cite{katz_explorekit_2016} generates new features by applying predefined operators, including unary, binary and higher-order operators, to existing features. 
More recently, \texttt{SMARTFEAT}~\cite{lin_smartfeat_2024} uses Foundation Models (FMs) to construct new features. It uses a prompt strategy to instruct the FMs to generate transformation functions for generating diverse and meaningful features. 

\textbf{Implicit feature construction} generates new features by indirectly manipulating existing feature vectors in hidden layers, as illustrated in Fig.~\ref{f10_aug_feature_construction} (b). 
For example, \texttt{GAINS}~\cite{xiao_beyond_2023} starts by training an encoder to map features into vectors, and then uses an evaluator to optimize the feature vector along the gradient direction. 
Finally, a decoder generates optimal feature subsets.
Wang et al.~\cite{wang_reinforcement-enhanced_2023} combine the explicit and implicit methods by encoding feature transformation operation sequences into embedding vectors. 
They modify \texttt{GAINS} by replacing the original feature vectors with these operation sequence vectors to disclose a new feature space with discriminative patterns.

\remarkbox{
ML is heavily based on high-quality features, which makes feature construction of great importance. 
For explicit methods, previous work typically relies on predefined operators that are hard to refine and often generate meaningless features. 
Recently, \texttt{SMARTFEAT}~\cite{lin_smartfeat_2024} has begun to use foundation models for feature construction, indicating a potential future direction.
For implicit methods, manipulating existing feature vectors in hidden layers lacks interpretability. 
Integrating explicit methods, such as \texttt{GAINS}~\cite{wang_reinforcement-enhanced_2023} may be a potential solution.
Meanwhile, since implicit methods involve encoding features into vectors, using LLMs for feature encoding to better capture feature semantics is also a promising direction.
}

\subsubsection{Cell Imputation (\GCI{})}

Cell imputation is the procedure of generating assumed values to replace the unknown or missing values within the tabular data. 
Estimating such unseen values within the dataset is particularly challenging due to the high heterogeneity in data types and the large size of datasets, which often comprise millions of rows. 
Cell imputation methods can be divided into two main categories: statistical cell imputation and deep-learning-based cell imputation.

\textbf{Statistical cell imputation} identifies and computes missing cell values based on the statistical characteristics of the original table. 
For example, \texttt{MICE}~\cite{buuren_mice_2011} creates a statistical model for each variable (i.e., a column in the tabular data setting) with missing values, filling in those missing values iteratively until convergence is reached.
The statistical model can be a simple mean or median, or it can be a more complex statistical model like a regressor. 
\texttt{MissForest}~\cite{stekhoven_missforestnon-parametric_2012} trains a random forest on the observed parts of the dataset and then use this trained random forest to predict the missing values. 

\textbf{Deep-learning-based cell imputation} has recently utilized a deep generative network to model tabular data for cell imputation, aiming to better capture the distribution and correlations between attributes in the original table. These methods can be further categorized on the basis of the type of generative model they use:
\begin{itemize}[leftmargin=*]
\item \emph{VAE-based methods} typically employ an encoder to capture the underlying structure of table and a decoder to generate the imputed value.
For example, \texttt{MIWAE}~\cite{mattei_miwae_2019} handles missing data using the encoder to approximate the posterior of the latent variables given the observed data and the decoder to reconstruct the complete data. 
\texttt{HI-VAE}~\cite{nazabal_handling_2020} modifies the VAE architecture to handle missing data by marginalizing out the missing data. To be specific, \texttt{HI-VAE} uses input dropout to make the encoder rely only on the observed data. Meanwhile, \texttt{HI-VAE} modifies the VAE decoder to factorize the likelihood into separate components for observed and missing data.

\item \emph{GAN-based methods} generally employ the generator to impute missing data and the discriminator to distinguish between real and imputed data. 
For instance, \texttt{GAIN}~\cite{yoon_gain_2018} and \texttt{MisGAN}~\cite{li_misgan_2019} use the generator to impute missing data by analyzing real data vectors, while the discriminator, aided by a hint vector that provides additional information about the missing data, differentiates between observed and imputed components.
However, \texttt{GAIN} and \texttt{MisGAN} are theoretically only supported under the \emph{Missing Completely at Random} (MCAR) mechanism, where the probability of missing data is uniform across all cells regardless of their values. In contrast, the \emph{Missing at Random} (MAR) mechanism, where the likelihood of missing data depends on the observed values, presents a greater challenge in practice.
\texttt{MIGAN}~\cite{dai_multiple_2021} addresses this by modeling the conditional distribution of missing values based on the observed data for each pattern of missing data.

\item \emph{Diffusion-based methods} use a reverse denoising process to learn the distribution of the data and perform the imputation. 
\texttt{TabCSDI}~\cite{zheng_diffusion_2023} adapts the diffusion model to model the distribution of the missing parts given the observed parts by employing the reverse denoising process.

\end{itemize}

\remarkbox{
Cell imputation addresses missing values in tabular data, a common issue in real-world datasets, and has been a longstanding concern for the database community.
Statistical methods rely on analytically formulated techniques that often require human intervention and may not capture complex relationships between attributes. 
Recently, there has been a noticeable shift from these methods to deep-learning-based methods.
However, deep-learning-based methods, while powerful, come with the challenges of more complex optimization and often necessitate fully observed datasets for training.
A promising future direction is the adoption of self-supervised methods, which can leverage partially observed data without requiring complete datasets. Furthermore, some recent work, such as \texttt{Hyperimpute}~\cite{jarrett_hyperimpute_2022}, utilizes an AutoML framework, presenting another promising direction for the field.
}

\subsubsection{Table Synthesis (\GTS{})}

We define generation-based TDA at the table level as table synthesis, which involves generating both rows and columns based on the original table.
At present, most research focuses on generating only rows, columns, or cells individually.
However, table synthesis as a whole could be a promising direction for future research.
Tables with small size of samples and limited features are common, such as web tables. In this case, a simple and feasible approach is to combine previous works that generate rows or columns separately, but additional steps bring increased resource consumption and error propagation.

Therefore, table synthesis that generates rows and columns simultaneously in one pass is necessary and worth exploring.
Currently, a transformer-based method \texttt{TransTab}~\cite{wang_transtab_2022},
which represents a combination of retrieval- and generation-based TDA.
\texttt{TransTab} encodes tables into tokens to retain knowledge across the table pool, redistributes attention on these tokens to highlight important features, and ultimately synthesizes a new encoded table.
Prompt engineering is a potential direct table synthesis strategy, where crafting appropriate prompts could guide LLMs in generating tables.

\remarkbox{
Table synthesis aims to enhance the original table by generating both rows and columns, particularly useful when the training dataset has low-quality features and limited samples. 
This emerging field requires further exploration. Directly combining row and column generation methods can lead to error propagation. 
Therefore, more targeted solutions are necessary, such as leveraging large language models.
}

\subsection{Retrieval vs. Generation in TDA} 
\label{section:aug-discussion}

In this section, we summarize and analyze the aforementioned TDA techniques from the perspective of comparing the retrieval-based approaches and generation-based approaches. 
First, using the proposed level-based taxonomy, we will distinguish from the task objectives and methodologies of these two approaches at different levels of granularity in Section~\ref{section:aug-discussion-table-level}. Then, we will provide a general overview of the pros and cons of retrieval-based approaches and generation-based approaches in Section~\ref{section:aug-discussion-pros-and-cons}, enabling researchers to choose the approach that best matches their tasks and requirements.

\subsubsection{Comparison at Different Levels}
\label{section:aug-discussion-table-level}

We analyze the differences between retrieval-based and generation-based methods at each level:
\begin{itemize}[leftmargin=*]
\item {Row-level} (\REA{} vs \GRG{}): Retrieval-based methods (\REA{}) focus more on the similarity between the original table and candidate tables in the table pool. This method can introduce new and more diverse samples, but often overlooks the impact of retrieved records on the distribution of the original table. In contrast, generation-based methods (\GRG{}) pay attention to the original table (e.g., statistical distribution and the relationships between columns) to ensure that the generated data maintain the logic of the original table. However, these synthetic samples tend to be similar to the original ones and cannot introduce new, previously unseen samples.
\item {Column-level} (\RSA{} vs \GFC{}): Retrieval-based methods (\RSA{}) are akin to the database join operations, requiring the specification of the target column in the original table, which can introduce completely new related features. Generation-based methods (\GFC{}), similar to feature engineering, optimize existing features to generate new ones and then select the best among them.
\item {Cell-level} (\RCC{} vs \GCI{}): Retrieval-based methods (\RCC{}) target regular empty cells and special cells such as column headers. They identify candidate values by searching for similar tables in the table pool or KB entities and then rank these candidates. This process is repeated for each null value and may not be friendly to often-empty regular cells.
In contrast, generation-based methods (\GCI{}) focus on regular empty cells, synthesizing new values by analyzing the statistical distribution and structural information of the original table. This method can handle multiple null values in one pass and can handle special cells with simple modifications.
\item {Table-level} (\RTI{} vs \GTS{}): Retrieval-based methods (\RTI{}) that integrate both rows and columns from external tables can easily introduce null values. While generation-based methods (\GTS{}) do not produce null values, their credibility is uncertain due to a lack of interpretability and potential model hallucination.
\end{itemize}

\subsubsection{Summary of Pros and Cons}
\label{section:aug-discussion-pros-and-cons}
We summarize the pros and cons of retrieval-based and generation-based TDA, highlighting their common challenges, as illustrated in Table~\ref{aug-discussion}.

Retrieval-based TDA first retrieves related tables from table pools and then uses the retrieved tables for augmentation. The row-and-column structure and the data abundance make this method uniquely suited for tabular data.
This approach enriches the original table with real external data, improving interpretability and introducing new, related information.
Although studied for over a decade, recent advancements in deep learning, such as PLMs~\cite{fan_semantics-aware_2023,hu_automatic_2023}, have revitalized the field.
However, challenges remain. The retrieval process requires preprocessing and indexing potentially millions of tables, entailing improvements in efficiency and scalability. 
Additionally, the lack of labeled data in large-scale table pools suggests self-supervised approaches as a future direction. Furthermore, retrieval-based methods often struggle with generalization.

Generation-based TDA generates synthetic data for TDA.
This approach does not require external data sources, eliminating the need for preprocessing and indexing numerous tables, thereby saving time and resources.
Synthetic data generation also offers privacy protection. However, this field is nascent and faces challenges. Current deep generative models are often over-parameterized, leading to overfitting, especially with small tables. Moreover, the process lacks interpretability and may lead to models' hallucination.

In addition to the challenges of the two methods mentioned above, TDA faces some common issues due to the nature of tabular data:
First, capturing the table semantics, such as differentiating the context of ``apple'' in a technology table versus a fruit table. 
Second, capturing the structural information, including row (column) invariance and relationships between columns (columns only depend on a subset of other columns). 
Additionally, incorporating generative AI models in a more effective and interpretable manner is a future direction for both retrieval- and generation-based TDA approaches. Given the respective pros and cons of retrieval- and generation-based methods, exploring their combination is also worthwhile.

\begin{table}[!htbp]
\caption{The comparison and analysis between retrieval- and generation-based TDA.}%
\label{aug-discussion}
\centering
\resizebox{\linewidth}{!}{
\begin{tabular}{@{}l||l|l||l@{}}
\toprule
                                                                                                                 & Pros                                                                                                            & Cons                                                                                                                   & Common Challenge and Opportunities                                                                                          \\ \midrule
\multirow{3}{*}{\begin{tabular}[c]{@{}l@{}}\textbf{Retrieval}\\ (studied for long\\ and still has vitality)\end{tabular}} & \multirow{3}{*}{\begin{tabular}[c]{@{}l@{}} interpretability\\ introduce brand new information\end{tabular}}           & \multirow{3}{*}{\begin{tabular}[c]{@{}l@{}}face efficiency and scalability issues\\ lack of labeled data \\ lack generalization ability\end{tabular}} & \multirow{5}{*}{\begin{tabular}[c]{@{}l@{}}capture the table semantics\\ capture the table structure \\ the incorporation of generative AI models \\ the combination of retrieval- and generation-based methods\end{tabular}} \\
                                                                                                                 &                                                                                                                 &                                                                                                                        &                                                                                                            \\
                                                                                                                 &                                                                                                                 &                                                                                                                        &                                                                                                            \\
                                                                                                                 \cmidrule{1-3}
\multirow{2}{*}{\begin{tabular}[c]{@{}l@{}}\textbf{Generation}\\ (a growing field)\end{tabular}}                          & \multirow{2}{*}{\begin{tabular}[c]{@{}l@{}}privacy protection\\ not require external data sources\end{tabular}} & \multirow{2}{*}{\begin{tabular}[c]{@{}l@{}}over-parameterized\\ lack interpretability and may cause model hallucination \end{tabular}}                    &                                                                                                            \\
                                                                                                                 &                                                                                                                 &                                                                                                                        &                                                                                                           
\\ \bottomrule
\end{tabular}
}
\end{table}

\section{Techniques in Post-Augmentation}
\label{section:postaug}
In this section, we mainly focus on three crucial aspects of post-augmentation: publicly available datasets used for TDA and its assessment (see Section~\ref{section:postaug-dataset}), policies for evaluating the performance of the augmentation methods (see Section~\ref{section:postaug-evaluation}), and strategies for further optimizing the augmentation module (Section~\ref{section:postaug-optimization}).

\subsection{TDA Datasets}
\label{section:postaug-dataset}

This section delves into the classic datasets commonly used in TDA work, aiming to assist newcomers to the field. We only consider datasets that have been utilized in multiple TDA works, as these are likely to serve as strong benchmarks for the community. 
Key characteristics of each dataset are summarized in Table~\ref{postaug-dataset-table}. 
These datasets can be broadly categorized into two main groups: retrieval-based TDA datasets and generation-based TDA datasets. The main difference between these categories lies in the input data requirements for the respective TDA approaches. 
Retrieval-based TDA necessitates the collection of additional external tables along with the original table, whereas generation-based TDA only requires the original table as input.
Nevertheless, training a generative model for TDA might require a substantial quantity of tabular data. Yet, from a post-augmentation viewpoint, the extensive data used for training is not taken into account here.

\begin{table}[!htbp]
\caption{Representative datasets used in TDA studies, including their basic properties and the specific TDA tasks they are suitable for. Some works provide either the number of columns (rows) or the average number of cols (rows). We manually calculate the missing values from the other and mark them in italics.}%
\label{postaug-dataset-table}
\resizebox{\linewidth}{!}{
\begin{threeparttable}
\begin{tabular}{@{}llccccccll@{}}
\toprule
No. & Dataset                                & Tables & Cols & AVG.\,Cols & Rows    & AVG.\,Rows  & Scenario        & Suitable tasks & Used-by                                                                                                             \\ \midrule
1   & Web\_Manual~$^\clubsuit$                             & 371    & $\smallsetminus$       & $\smallsetminus$              & \textit{18921}       & 51           & Web       & \REA{}, \RSA{}         & Limaye et al.~\cite{limaye_annotating_2010}, \texttt{TabEL}~\cite{arenas_tabel_2015}                                                                                                       \\
2   & Wiki\_Link~$^\clubsuit$                             & 6085   & $\smallsetminus$       & $\smallsetminus$              & \textit{121700}       & 20           & Web       & \REA{}         & Limaye et al.~\cite{limaye_annotating_2010}, \texttt{TabEL}~\cite{arenas_tabel_2015}                                    \\
3   & WDC Web table corpus~$^\diamondsuit$                  & 50M      & 250M       & \textit{5}              & \textit{700M}       & 14            & Web       & \RSA{}             & \texttt{JOSIE}~\cite{zhu_josie_2019}, \texttt{PEXESO}~\cite{dong_efficient_2021}, \texttt{DeepJoin}~\cite{dong_deepjoin_2023}, \texttt{Starmie}~\cite{fan_semantics-aware_2023}                                                                                    \\
4   & WikiTables corpus~$^\heartsuit$                      & 1.6M   & \textit{30.4M}         & 19             & $\smallsetminus$       & $\smallsetminus$            & Web       & \REA{}, \RSA{}, \RCC{}     & \begin{tabular}[c]{@{}l@{}}\texttt{TabEL}~\cite{arenas_tabel_2015}, \texttt{Entitables}~\cite{zhang_entitables_2017}, \texttt{Table2Vec}~\cite{zhang_table2vec_2019},\\ \texttt{CellAutoComplete}~\cite{zhang_auto-completion_2019}, \texttt{TURL}~\cite{deng_turl_2022}, \texttt{DeepJoin}~\cite{dong_deepjoin_2023}, \texttt{RATA}~\cite{glass_retrieval-based_2023}\end{tabular}      \\
5   & \texttt{TUS Small}~$^\spadesuit$                            & 1,530  & 14,810  & \textit{10}              & 6.8M   & 4,466       & Relational & \REA{}             & \texttt{TUS}~\cite{nargesian_table_2018}, \texttt{SANTOS}~\cite{khatiwada_santos_2023}, \texttt{Starmie}~\cite{fan_semantics-aware_2023}, AutoTUS~\cite{hu_automatic_2023}                                                                                       \\
6   & TUS Large~$^\spadesuit$                              & 5,043  & 54,923  & \textit{11}              & \textit{9.7M}       & 1,915        & Relational & \REA{}             & \texttt{TUS}~\cite{nargesian_table_2018}, \texttt{D$^3$L}~\cite{bogatu_dataset_2020}, \texttt{SANTOS}~\cite{khatiwada_santos_2023}, \texttt{Starmie}~\cite{fan_semantics-aware_2023}, \texttt{AutoTUS}~\cite{hu_automatic_2023}                                                                                  \\
7   & SANTOS Small~$^\varclubsuit$                           & 550    & 6,322   & \textit{11}             & 3.8M   & 6,921       & Relational & \REA{}             & \texttt{SANTOS}~\cite{khatiwada_santos_2023}, \texttt{Starmie}~\cite{fan_semantics-aware_2023}, \texttt{AutoTUS}~\cite{hu_automatic_2023}                                                                                            \\
8   & SANTOS Large~$^\varclubsuit$                          & 11,090 & 123,477 & \textit{11}               & 70M    & 7,675        & Relational & \REA{}             & \texttt{SANTOS}~\cite{khatiwada_santos_2023}, \texttt{Starmie}~\cite{fan_semantics-aware_2023}, \texttt{AutoTUS}~\cite{hu_automatic_2023}                                                                                            \\ \midrule
9   & BTS~$^\vardiamondsuit$                                    & 1      & 30      & $\smallsetminus$              & 1M & $\smallsetminus$           & Relational & \GRG{}             & \texttt{table-GAN}~\cite{park_data_2018}, \texttt{ITS-GAN}~\cite{chen_faketables_2019}\\
10  & \begin{tabular}[c]{@{}l@{}}UCI datasets~$^\varheartsuit$ \\ (e.g., Adult, Covertype)\end{tabular}    & $\smallsetminus$      & $\smallsetminus$       & $\smallsetminus$              & $\smallsetminus$       & $\smallsetminus$           & Relational & \GRG{}, \GFC{}, \GCI{}     & \begin{tabular}[c]{@{}l@{}}\texttt{CTGAN}~\cite{xu_modeling_2019}, \texttt{SIGRNN}~\cite{al-bahrani_sigrnn_2021}, \texttt{GOGGLE}~\cite{liu_goggle_2023}, \texttt{RelDDPM}~\cite{liu_controllable_2024}, \texttt{GAINS}~\cite{xiao_beyond_2023},\\ \texttt{GAIN}~\cite{yoon_gain_2018}, \texttt{HI-VAE}~\cite{nazabal_handling_2020}, \texttt{TabCSDI}~\cite{zheng_diffusion_2023}, \texttt{HyperImpute}~\cite{jarrett_hyperimpute_2022}\end{tabular} \\
11  & \begin{tabular}[c]{@{}l@{}}Kaggle~$^\varspadesuit$  \\ (e.g., Diabetes, Bank)\end{tabular}          & $\smallsetminus$      & $\smallsetminus$       & $\smallsetminus$              & $\smallsetminus$       & $\smallsetminus$           & Relational & \GFC{}             & \texttt{GAINS}~\cite{xiao_beyond_2023}, \texttt{SMARTFEAT}~\cite{lin_smartfeat_2024}                                                                                                    \\
12  & \begin{tabular}[c]{@{}l@{}}OpenML repository~$^\bigstar$ \\ (e.g., Heart, Horce)\end{tabular} & $\smallsetminus$      & $\smallsetminus$       & $\smallsetminus$              & $\smallsetminus$       & $\smallsetminus$           & Relational & \GRG{}             & \texttt{ExploreKit}~\cite{katz_explorekit_2016}, \texttt{GAINS}~\cite{xiao_beyond_2023}, \texttt{RelDDPM}~\cite{liu_controllable_2024}                                                                                          \\ \bottomrule
\end{tabular}
\begin{tablenotes}
    \item[$\clubsuit$] \url{http://websail-fe.cs.northwestern.edu/TabEL/\#Web\_Manual}, publication year: 2010.
    \item[$\diamondsuit$] \url{https://webdatacommons.org/webtables/\#results-2015}, publication year: 2015.
    \item[$\heartsuit$] \url{http://websail-fe.cs.northwestern.edu/TabEL/\#WikiTables}, publication year: 2015.
    \item[$\spadesuit$] \url{https://github.com/RJMillerLab/table-union-search-benchmark}, publication year: 2018.
    \item[$\varclubsuit$] \url{https://github.com/northeastern-datalab/santos}, publication year: 2023.
    \item[$\vardiamondsuit$] \url{https://www.transtats.bts.gov/DataIndex.asp} 
    \item[$\varheartsuit$] \url{https://archive.ics.uci.edu}
    \item[$\varspadesuit$] \url{https://www.kaggle.com}
    \item[$\bigstar$] \url{https://www.openml.org}
\end{tablenotes}
\end{threeparttable}
}
\end{table}

\textbf{Retrieval-based TDA datasets} usually consist of a table pool with hundreds to thousands of tables.
The earliest examples, Web\_Manual and Wiki\_Link, originate from the same study~\cite{limaye_annotating_2010}. 
In the Web\_Manual dataset, the researchers use Wikipedia tables as their queries and retrieve 371 Web tables to serve as the target corpus. 
These Web tables are then manually annotated with entities, types, and inter-column relationships. 
In contrast, the Wiki\_Link dataset is designed for larger-scale use without extensive human annotation. 
It is created by selecting Wikipedia tables where at least 90\% of the cell values were internally linked to entities in Wikipedia. 
While this automated approach leads to a larger dataset, the annotations are limited to only entity information, without the more detailed and accurate annotations found in Web\_Manual.
Because of the trade-offs between dataset size and annotation quality, both Web\_Manual and Wiki\_Link are less frequently used in recent TDA research.
Nargesian et al.~\cite{nargesian_table_2018} focus on table union search, an immediate step before entity augmentation, and propose two synthesized datasets TUS Small and TUS Large. 
They identify high-quality base tables (with abundant rows and at least 5 textual columns) from Canadian and UK Open Data and then partition the base tables horizontally and vertically to obtain non-overlapping unionable tables to the base ones.
Similarly, Khatiwada et al.~\cite{khatiwada_santos_2023} use the same dataset synthesis technique as \texttt{TUS} to create SANTOS Small and SANTOS Large. 
These four datasets are commonly used for recent entity augmentation tasks.
Additionally, open data repositories like the WDC Web table corpus and WikiTables corpus are frequently used. They contain a vast number of tables from various domains, adaptable for different types of retrieval-based TDA tasks (e.g., \REA{}, \RSA{}, and \RCC{}).

\textbf{Generation-based TDA datasets} typically do not require a table pool or the collection and annotation of multiple similar tables, as only a single original table is necessary for generation-based TDA. 
Consequently, there are no datasets specifically designed for generation-based TDA tasks. Most generation-based TDA methods utilize multiple original tables from various fields, targeting different tasks such as binary classification, multi-class classification, and regression.
For example, \texttt{table-GAN} and \texttt{ITS-GAN} adopt the BTS dataset, which is a single table containing 1 million records of domestic air ticket sales. 
These two methods take a portion of the BTS dataset as input and generate augmented tables. 
The effectiveness of these augmented tables is then assessed through regression tests on the "ticket price" attribute in the original BTS dataset.
Common sources for generation-based TDA datasets are primarily from three public platforms: UCI, Kaggle, and OpenML. These platforms are popular partly due to their large scale and diverse range of tables.

\remarkbox{
It is natural to observe that the same dataset, when processed differently, can be applied to different TDA tasks.
For example, \texttt{EntiTables}~\cite{zhang_entitables_2017} adapts the WikiTables corpus for entity augmentation, while \texttt{DeepJoin}~\cite{dong_deepjoin_2023} uses WikiTables corpus for schema augmentation.
\texttt{EntiTables} filters out those tables that focus on entities, specifically those where the leftmost column contains unique entities, to create a dataset for the \REA{} task. Entities in related tables, such as those containing entities from the original table or having similar captions, are considered candidate entities.
On the other hand, \texttt{DeepJoin} preprocesses datasets by stipulating the key columns as the ones with the most unique values in each table for subsequent joins.
Another observation is that the current generation-based TDA methods do not have standardized benchmarks. These methods often choose their own datasets, which makes it challenging to evaluate and compare their performance on a shared dataset, thereby hindering the assessment of their effectiveness and differences.

With the development of generative AI, an increasing number of TDA works are incorporating the use of pre-training. This has led to a growing need for the construction of robust pre-training tabular datasets.
An ideal pre-training dataset should possess three main attributes: high quality, large scale, and wide coverage.
Previous datasets often compromise between quality and scale, but combining human expertise with generative models might address this issue.
Wide coverage means curating diverse datasets across multiple domains, such as finance and healthcare, and various data types like CSV, JSON, and Markdown. 
This diversity is essential for pre-training large-scale generative models to handle the variety of tabular data found in real-world scenarios. However, privacy concerns might pose challenges in developing high-quality TDA datasets.
}

\subsection{Evaluation Polices}
\label{section:postaug-evaluation}

We summarize the common evaluation policies for TDA works. These polices can be categorized into two main groups based on the involvement of models: original-table-based evaluation and model-based evaluation. 
Many studies~\cite{park_data_2018,chen_faketables_2019} have utilized both evaluation methods simultaneously.

\textbf{Original-table-based evaluation} refers to evaluating the augmented table by comparing it to the groundtruth, which is typically the original table or its derivatives.
For example, several works~\cite{zhang_entitables_2017,zhang_table2vec_2019} start from a base table and derive a subtable from it to serve as the original table $T^O$, while the entities or columns outside this subtable are used as the ground truth. This method can efficiently construct the ground truth, but the range of truth values is rather limited, and there may be data related to the base table in the table pool that are not present in the base table (e.g., the base table about IT companies in one country with a table in the table pool about another country).
Several works~\cite{park_data_2018,chen_faketables_2019} calculate the cumulative distribution functions (CDFs) of the augmented and original tables to compare their statistic similarity. 
While this method is simple and effective, it only evaluates statistical distribution information and cannot capture more complex details, such as relationships between columns (e.g., the connection between "position" and "salary").

\textbf{Model-based evaluation} refers to feeding the augmented table alongside other \emph{baseline tables} to a specific ML model and then evaluating the model's performance. 
These baseline tables generally include three primary datasets:
(1) \textbf{None}~\cite{park_data_2018,chen_faketables_2019,liu_feature_2022,chai_selective_2022,zhao_leva_2022,galhotra_metam_2023,fan_semantics-aware_2023,liang_featnavigator_2024} refers to the original dataset without any augmentation. 
For example, \texttt{ITS-GAN}~\cite{park_data_2018} feeds the augmented and original table to a specific classification model that performs a grid search over RandomForest, AdaBoost, and GradientBoosting, then comparing the corresponding classification results.
(2) \textbf{Random}~\cite{liu_feature_2022,chai_selective_2022} refers to the original dataset augmented with randomly selected candidates. For instance, a schema augmentation work \texttt{AutoFeature} treats features in a table pool as independent entities and randomly selects a predefined number of features to augment the original table as the Random baseline.
(3) \textbf{All}~\cite{liu_feature_2022,chai_selective_2022,zhao_leva_2022} refers to the original dataset augmented with all possible candidates. For example, \texttt{Leva} involves a ALL baseline that joins the original table with as many tables as possible.
In general, None baseline is suitable for both retrieval- and generation-based TDA while the other two are suitable only for retrieval-based TDA.

\remarkbox{
The two evaluation methods, original-table-based and model-based, each come with their own set of pros and cons.
The original-table-based evaluation is straightforward and efficient but may lack precision and be limited in scope, especially for specific augmentation needs like enhancing the minority class.
The model-based evaluation offers greater accuracy but requires more time and resources.
Additionally, using models introduces extra layers of uncertainty, such as model characteristics and hyperparameters, making it difficult to control variables and ensure fairness.
This also complicates determining whether issues arise from the model itself or the data augmentation method.
To address these tradeoffs, several studies~\cite{park_data_2018,chen_faketables_2019} adopt a hybrid approach, using both evaluation methods. This raises an interesting question: do these two evaluation methods yield conflicting results?
It may be worth exploring the development of alternative evaluation methods, such as rule-based TDA evaluation policies, which could potentially strike a balance between efficiency and accuracy better than current methods.
}

\subsection{Optimization Strategies}
\label{section:postaug-optimization}

Optimization strategies aim to further refine the augmented results based on the performance of specific downstream ML models. These techniques can be categorized into two main types: iteration-based and reinforcement-learning-based.

\textbf{Iteration-based optimization} involves a simple and direct method of using feedback from the downstream ML model to determine whether adding a candidate augmentation enhances the performance of the task.
For instance, Chepurko et al.~\cite{chepurko_arda_2020} devise the random injection feature selection (RIFS) algorithm, which compares model performance using candidate features against deliberately constructed random features as a baseline.
The objective is to identify and eliminate irrelevant features, finding a subset of features that contain signals relevant to the downstream ML task.
Their subsequent work \texttt{ARDA}~\cite{chepurko_arda_2020} is an automated system that searches and joins data with the input table end-to-end. 
Similarly, \texttt{Leva}~\cite{zhao_leva_2022} leverages the supervision signal from the downstream ML task to filter out unnecessary information. \texttt{Leva} uses a graph to capture information from the entire database, including both useful and potentially spurious relationships. 
During training, the downstream ML model will automatically focus on using the valuable information while ignoring or downweighting the non-useful parts. 
{More recently, \texttt{FeatNavigator}~\cite{liang_featnavigator_2024} assesses the actual utility gain of candidate features by running ML model on the original and augmented tables; it then iteratively selects features based on their  utility gain and the feasibility of the join path.}

\textbf{Reinforcement-learning-based optimization} employs reinforcement learning (RL) to explore the features that improve the performance of the ML model. 
For example, Liu et al.~\cite{liu_automatic_2021} propose an RL-based automatic data search system that retrieves fresh training data from table pools and interacts with the downstream ML model. 
In this RL-based framework, each training data point and its corresponding influence score, calculated by the ``Environment'', serve as the ``State''. Given the ``State'', the ``Agent'' composed of a Search-Policy selects the optimal ``Action'' (a set of training data points retrieved from the table pool), and feeds them back into the ``Environment''.
\texttt{AutoFeature}~\cite{liu_feature_2022} makes further improvement by not only exploring features that boost performance but also utilizing rarely selected ones to avoid local optima.
Chai et al.~\cite{chai_selective_2022} further extend \texttt{AutoFeature} by broadening the scope of table pools, such as enterprise data warehouses, online repositories, and data markets.

\remarkbox{
Optimization strategies have proven effective in improving downstream ML task performance, with a noticeable trend towards the use of RL.
However, they often come with computational overhead, which can be a constraint in time-sensitive or resource-limited situations and can impact scalability.
As such, a worthwhile research direction is to improve the iteration efficiency of these optimization strategies while maintaining their effectiveness. This could involve reducing the number of required iterations or exploring the use of lighter surrogate models in place of the original, specific ML models.
}

\section{Trends and Opportunities}
\label{section:future}

This section examines the current landscape and future prospects of TDA techniques. 
We first explore emerging trends shaping the field (Section~\ref{subsection:future-trends}), followed by a discussion of promising opportunities for further research and application (Section~\ref{subsection:future-opportunities}). By highlighting these aspects, we aim to inspire continued interest in this vital area of study.

\subsection{Major Trends in TDA Development}
\label{subsection:future-trends}

Based on our review of numerous research articles and our three-step pipeline, we have identified three significant trends in current TDA work:
\begin{enumerate}
\item[T1.] \emph{Enhanced Table Representation}. In the pre-augmentation phase, researchers are increasingly employing more advanced table representations to better capture both structural and semantic details within tables;
\item[T2.] \emph{Confluence of Retrieval and Generation}. In the augmentation phase, retrieval- and generation-based methods offer distinct advantages and disadvantages. A fruitful path forward may lie in the integration of these two methodologies;
\item[T3.] \emph{Automated TDA}. Looking at the entire TDA pipeline, a promising future direction is the creation of end-to-end automated TDA systems that can efficiently manage the entire TDA process.
We proceed to elaborate on each of these trends.
\end{enumerate}

\smallskip
\noindent\textbf{T1. Enhanced Table Representation}.
The representation of tables is crucial for effective TDA. 
By accurately capturing content, semantic, and structural information in table representation, downstream augmentation processes can achieve better outcomes.
Accordingly, there is a clear trend in TDA towards more complex and sophisticated table representations.
Recent TDA research~\cite{deng_turl_2022,miao_watchog_2023} includes diverse information such as table context and metadata, unlike earlier work~\cite{nargesian_table_2018,bogatu_dataset_2020} that focused solely on table content. 
Additionally, newer TDA approaches~\cite{zhao_leva_2022,liu_goggle_2023} utilize graph structures to capture relational information within tables more effectively, encoding these structures for improved table representation.
Notably, many of the latest TDA techniques, both retrieval-based methods~\cite{fan_semantics-aware_2023,hu_automatic_2023} and generation-based methods, utilize~\cite{tran_differentially_2024} generative AI (e.g., PLMs) to generate robust table representations, leveraging their semantic understanding and generalization capabilities.
While generative AI models have shown promise in tabular data representation, there remains ample opportunity for further development, especially compared to advancements in NLP and CV.
In conclusion, the move towards more sophisticated and comprehensive table representations is a key trend driving progress in TDA research.

\smallskip
\noindent\textbf{T2. Confluence of Retrieval and Generation}. 
As discussed in Section~\ref{section:aug-discussion}, retrieval-based and generation-based TDA have their own pros and cons. 
Retrieval-based methods incorporate external data sources for better interpretability but struggle with efficiency and scalability as data increases. Generation-based methods do not use external data, thus lacking interpretability and potentially leading to model hallucination. 
Given the trade-offs between these two approaches, combining their strengths while mitigating their weaknesses should be the future direction for TDA research. This aligns with the prevalent use of RAG models in the field of NLP, where the benefits of both retrieval- and generation-based techniques are harnessed. By blending these complementary methodologies, we can work towards more robust, efficient, and interpretable TDA solutions that can meet the evolving needs of diverse applications and domains.

\smallskip
\noindent\textbf{T3. Automated TDA}.
A notable trend in TDA is the move towards automating the entire process, creating end-to-end platforms that integrate various operators.
For example, Chepurko et al.~\cite{chepurko_arda_2020} propose \texttt{ARDA}, an end-to-end system that integrates multiple operators (e.g., imputation, hyperparameters optimization, feature selection, etc.) for automatic schema augmentation.
Similarly, \texttt{Hyperimpute}~\cite{jarrett_hyperimpute_2022} automates cell imputation by proposing an AutoML framework that utilize search algorithms to automatically select candidate ML models.
Despite these advancements, current solutions often focus on single subtasks of TDA, leaving room for further exploration and development.
An area for improvement is to consider the TDA pipeline as a whole, integrating various pre-augmentation and augmentation operators, and intelligently selecting suitable operators based on user requirements. Additionally, refining the automated TDA pipeline could involve an iterative step to identify and eliminate unnecessary or even harmful steps, based on augmentation evaluation and data characteristics.
By advancing towards more holistic, end-to-end TDA automation, the field can unlock greater efficiency, scalability, and customization, ultimately enhancing the practical value and applicability of TDA techniques.

\subsection{Emerging Opportunities for TDA}
\label{subsection:future-opportunities}

As we navigate the era of big data and the trends towards generative AI and autoML, we observe that obtaining high-quality data from massive amounts of data to facilitate generate AI is imperative. TDA is an important sub task of data quality and there is still a long way to go.
Firstly, from a data perspective, the quantity and complexity of tabular data have significantly increased. Tables now typically have multiple patterns~\cite{zheng_multimodal_2024} and contain millions of samples~\cite{fan_table_2023}, posing new challenges to TDA work.
Secondly, from a model perspective, ML models, especially generative AI, have always faced interpretability issues and potential of privacy leakage.
Below, we will elaborate on these key opportunities for further advancements in this field.

\smallskip
\noindent\textbf{O1. Multimodal TDA}.
Current TDA works often assume that tables contain only textual and numerical values. However, the reality is that modern tables often encompass a broader range of modalities, such as images~\cite{zheng_multimodal_2024}. 
Handling these multimodal tables presents unique challenges that current table processing techniques may not adequately address. 
The representation and indexing of such heterogeneous tables, for instance, may require fundamentally different approaches compared to traditional text-based and numerical tables.
Additionally, user needs might be expressed in modalities other than the tabular data itself~\cite{lin_li-rage_2023,yu_unified_2023}.
For instance, a user might use natural language to request the oversampling of a minor class. 
Addressing these multimodal user inputs and aligning them with TDA operations is another key challenge to tackle.

\smallskip
\noindent\textbf{O2. Efficiency and Scalability}.
Another key challenge for TDA is the issue of efficiency and scalability. 
Retrieval-based TDA approaches often involve similarity comparisons at the table pool level, which can encompass millions of tables or more. 
Despite that, most existing methods~\cite{zhu_josie_2019,dong_efficient_2021,khatiwada_santos_2023} are exact algorithms with a worst-case time complexity that is linear in relation to the product of the query column size and the table repository size, raising concerns about their scalability.
This makes it a crucial research direction to explore.
Meanwhile, the table itself may contain tens of thousands of records\footnote{Large-scale tables do not necessarily indicate high quality or no need for augmentation (e.g., many tables often have many records but few features for ML model training). Similarly, candidate tables in the table pool may also be large-scale.}.
Moreover, both retrieval- and generation-based TDA methods are embracing the large-scale generative AI models such as PLMs and LLMs. 
The computational and memory requirements of training, fine-tuning, and deploying these models can hinder the efficiency and scalability of TDA techniques in practice. 
Addressing the scalability challenges associated with large-scale generative AI models in TDA is another crucial area of research. 
Potential solutions may include the development of more efficient model architectures the exploration of model compression~\cite{zhang_draft_2024} and distillation techniques~\cite{wang_solo_2023}.

\smallskip
\noindent\textbf{O3. Domain-specific Tasks}.
Another promising direction for TDA research is the exploration of domain-specific applications. 
Domain-specific data often exhibit strong professionalism and have some unique characteristics~\cite{shi_domain-relevance_2023}.
For instance, medical data typically possesses specialized terminology and intricate data distributions that differ from more general tabular data. 
Developing TDA techniques tailored to these domain-specific characteristics could result in more effective and domain-friendly TDA strategies.
This might involve incorporating domain knowledge, such as expertise from professionals~\cite{li_chatdoctor_nodate}, knowledge base~\cite{agarwal_financial_2023} and knowledge graphs~\cite{abu-salih_healthcare_2023}, and utilizing specialized table representations~\cite{li_universal_2024}.
Combining retrieval-based and generation-based TDA holds promise for domain-specific tasks, allowing for extensive use of limited domain knowledge.
An interesting question arises as to whether the knowledge and techniques developed for TDA in one domain can be efficiently transferred to other domains. 
For example, can the insights and models derived from financial data augmentation be effectively applied to medical data, or vice versa?
The ability to successfully transfer TDA across domains would be highly valuable, as it could accelerate research and development efforts, enable the reuse of existing resources, and promote the exchange of ideas.

\smallskip
\noindent\textbf{O4. Interpretability}.
Existing TDA works face a notable challenge with respect to interpretability, particularly in generation-based TDA approaches.
Retrieval-based TDA, on the other hand, has demonstrated better interpretability, as it can be interpreted as leveraging the retrieved relevant table information to augment the original data.
One promising future direction for TDA research would be to explore the combination of retrieval- and generation-based TDA approaches.
This hybrid method could improve the interpretability of generation-based TDA while also harness the respective advantages of both approaches.
At the same time, the recent trend towards the incorporation of deep neural networks~\cite{fan_semantics-aware_2023,hu_automatic_2023,dong_deepjoin_2023}, such as language models, into TDA workflows has introduced a new challenge regarding interpretability. 
The complex, black-box nature of these models can make it difficult to understand and explain the underlying rationale behind the augmented data.
Addressing this interpretability challenge is crucial, as it impacts the trust, transparency, and practical adoption of TDA methods, particularly in domains where explainability is of paramount importance, such as high-stakes decision-making~\cite{sahoh_role_2023} or regulated industries~\cite{lisboa_coming_2023}.

\smallskip
\noindent\textbf{O5. Privacy and Security}.
Last but not least, privacy and security represent critical opportunities for TDA. 
For retrieval-based TDA approaches, they carry out similarity comparisons between the original table and the tables in the pool, which may lead to information leakage of the user's original table. In contrast, generation-based TDA has inherent advantages in privacy protection, and several works~\cite{jordon_pate-gan_2019,chen_faketables_2019,fan_relational_2020} have utilized record genration for privacy-preserving data publishing.
However, the recent trend towards the incorporation of LLMs into TDA workflows introduces a new set of privacy and security challenges. While LLMs have powerful generative capabilities, they also carry the risk of information leakage~\cite{sallou_breaking_2024,yao_survey_2024}, potentially exposing or memorizing sensitive data from their training corpora.
Addressing the offensive and defensive aspects of LLMs behaviors, including the development of robust privacy-preserving training techniques~\cite{yan_protecting_2024,peris_privacy_2023}, remains an active area of investigation.

\section{Conclusion}
\label{section:conclusion}

This survey presents a thorough investigation of tabular data augmentation (TDA) for ML, with a particular emphasis on the recent advancements in leveraging prevalent generative AI models.
Our work meticulously outlines the essential steps involved in TDA by constructing an end-to-end pipeline encompassing three critical procedures: (1) pre-augmentation, where we summarize and analyze the commonly used preparation techniques for TDA; (2) augmentation, where we systematically compare current TDA techniques, including both retrieval-based and generation-based approaches; and (3) post-augmentation, where we delve into the evaluation and optimization processes following TDA. Additionally, we provide a comprehensive analysis of the pros and cons of current methodologies and outline future trends and opportunities for TDA.

The era of generative AI heralds a transformational phase for TDA. ML on tabular data is ubiquitous and demands a substantial amount of high-quality data --- a requirement that generative AI can significantly enhance. Despite the distinct characteristics of tabular data, generative AI models have predominantly been applied to fields like computer vision and natural language processing, with their application to tabular data still in its nascent stages. This leaves a vast landscape for further innovation and advancements in TDA through the use of generative AI.
Our comprehensive survey aims to bridge this gap by providing a detailed, systematic overview of the current state of TDA and its potential future directions. We believe that this work can contribute to the community and serve as a valuable resource for researchers and practitioners alike.
Our endeavor will be presented as a continuously updated literature repository, maintained online at \repoUrl.

\section*{Acknowledgments}
The authors would like to thank Yifan Wu, Jianfeng Zhang, Ankai Hao, Zhaoyi Yuan, Jiahui Long, and Hongwei Yuan for their invaluable feedback, which significantly improved the manuscript.

\bibliographystyle{ACM-Reference-Format}
\bibliography{main}

\end{document}